\documentclass[10pt,journal,compsoc]{IEEEtran}
\usepackage{amsmath,amsfonts,amssymb,amsthm}
\usepackage{array}
\usepackage{graphicx,subfigure}
\usepackage{booktabs}
\usepackage{multirow}
\usepackage[nocompress]{cite}
\usepackage{mathrsfs,enumerate,color,bm}
\usepackage{algorithm,algpseudocode}
\usepackage{color}
\usepackage{lscape}
\newtheorem{theorem}{Theorem}[section]
\newtheorem{lemma}[theorem]{Lemma}

\newcommand{\R}{\mathbb{R}}

\newcommand{\rank}{\operatorname{rank}}

\graphicspath{{fig/}}

\begin{document}

\title{Tangent Space Based Alternating Projections for Nonnegative Low Rank Matrix Approximation}

\author{Guangjing~Song,
Michael~K.~Ng,
and~Tai-Xiang~Jiang,
        \IEEEcompsocitemizethanks{\IEEEcompsocthanksitem Guangjing~Song is with School of Mathematics and Information Sciences, Weifang University,
Weifang 261061, P.R. China. (email: sgjshu@163.com).
        \IEEEcompsocthanksitem M. K. Ng is with Department of Mathematics, The University of Hong Kong, Pokfulam, Hong Kong (e-mail: mng@maths.hku.hk). M. Ng's research supported in part by the HKRGC GRF 12306616, 12200317, 12300218, 12300519 and 17201020.
        \IEEEcompsocthanksitem T.-X. Jiang is with FinTech Innovation Center, School of Economic Information Engineering, Southwestern University of Finance and Economics, Chengdu, Sichuan, P.R.China (e-mail: taixiangjiang@gmail.com, jiangtx@swufe.edu.cn).  T.-X. Jiang's research is supported in part by the Fundamental Research Funds for the Central Universities (JBK2001011, JBK2001035). The Corresponding Author.}}


\IEEEtitleabstractindextext{%
\begin{abstract}
In this paper, we develop a new alternating projection method to compute
nonnegative low rank matrix approximation for nonnegative matrices.
In the nonnegative low rank matrix approximation method,
the projection onto the manifold of fixed rank matrices can be expensive as the singular value
decomposition is required. We propose to use the
tangent space of the point in the manifold
to approximate the projection onto the manifold in order to reduce the computational cost.
We show that the sequence generated by the alternating projections onto the tangent spaces of the fixed rank matrices manifold and the nonnegative matrix manifold,
converge linearly to a point in the intersection of the two manifolds where the convergent point
is sufficiently close to optimal solutions. This convergence result based inexact projection onto the manifold
is new and is not studied in the literature.
Numerical examples in data clustering, pattern recognition and hyperspectral data
analysis are given to demonstrate that
the performance of the proposed method is better than that of nonnegative matrix factorization methods
in terms of computational time and accuracy.
\end{abstract}

}

\maketitle
\IEEEdisplaynontitleabstractindextext
\IEEEpeerreviewmaketitle

%
\IEEEraisesectionheading{
\section{Introduction}\label{sec:introduction}}
\IEEEPARstart{N}{onnegative} data matrices appear in many data analysis applications.
For instance, in image analysis, image pixel values are nonnegative and the associated nonnegative image data matrices can be
formed for clustering and recognition \cite{chen2005matrix,chen2013non,ding2005equivalence,ding2006orthogonal,guillamet2002non,guillamet2003introducing,jing2012semi,lee1999learning,liu2012novel,liu2015predicting,wang2005non,zhang2005two}.
In text mining, the frequencies of terms in documents are nonnegative and the resulted nonnegative
term-to-document data matrices can be constructed for clustering \cite{berry2010text,li2006relationships,pauca2004text,xu2003document}.
In bioinformatics, nonnegative gene expression values are studied and nonnegative gene expression data matrices
are generated for diseases and genes classification \cite{cichocki2009nonnegative,kim2003subsystem,kim2007sparse,pascual2006nonsmooth,wang2006ls}.
Low rank matrix approximation for nonnegative matrices
plays a key role in all these applications. Its main purpose is to identify a latent feature
space for objects representation. The classification, clustering or recognition analysis can be
done by using these latent features.

Nonnegative Matrix Factorization (NMF) has emerged in 1994 by Paatero and Tapper \cite{paatero1994positive}
for performing environmental data analysis.
The purpose of NMF is to decompose an input $m$-by-$n$ nonnegative matrix ${\bf A} \in \mathbb{R}_{+}^{m\times n}$
into $m$-by-$r$ nonnegative matrix ${\bf B} \in \mathbb{R}_{+}^{m\times r}$ and
$r$-by-$n$ nonnegative matrix ${\bf C} \in \mathbb{R}_{+}^{r\times n}$:
${\bf A} \approx {\bf B} {\bf C}$,
and more precisely
\begin{equation} \label{nmfa}
\min_{ {\bf B}, {\bf C} \geq 0} \ \| {\bf A} - {\bf B} {\bf C} \|^2_F,
\end{equation}
where ${\bf B}, {\bf C} \geq 0$ means that each entry of ${\bf B}$ and ${\bf C}$ is nonnegative,
$\| \cdot \|_F$ is the Frobenius norm of a matrix,
and $r$ (the low rank value) is smaller than $m$ and $n$.
Several researchers have proposed and developed
algorithms for determining such nonnegative matrix factorization in the literature
\cite{cichocki2007hierarchical,choi2008algorithms,ding2006orthogonal,gillis2012accelerated, gillis2013fast,kuang2012symmetric,kuang2015symnmf,lee1999learning,lin2007projected,lee2001algorithms,pan2019generalized,yuan2005projective}.
Lee and Seung \cite{lee1999learning} proposed and developed NMF algorithms, and demonstrated
that NMF has part-based representation which can be used for
intuitive perception interpretation. For the development of NMF, we refer to the recently edited book \cite{naik2016non}.

In \cite{sm2019}, Song and Ng proposed a new algorithm for computing Nonnegative Low Rank Matrix (NLRM) approximation for nonnegative matrices. The
approach is completely different from NMF which has been studied for more than twenty five years. The new approach aims to find a nonnegative low rank matrix $X$ such that their difference is as small as possible.
Mathematically, it can be formulated as the following optimization problem
\begin{equation}\label{pmain}
\min_{\rank({\bf X})=r,{\bf X}\geq 0} \ \| {\bf A}- {\bf X}\|_{\textrm{F}}^{2}.
\end{equation}
The convergence of the proposed algorithm is also proved and  experimental results are shown that the minimized distance by the
NLRM method can be smaller than that by the NMF method. Moreover, according to the
ordering of singular values, the proposed method can identify important singular
basis vectors, while this information may not be obtained in the classical NMF.

\subsection{The Contribution}

The algorithm proposed in \cite{sm2019} for computing the nonnegative low rank matrix
approximation is based on using the alternating projections on the fixed-rank matrices manifold and the nonnegative matrices manifold. Note that the computational cost of the above alternating projection method is dominant by the calculation of the singular value decomposition
at each iteration. The computational workload of the singular value decomposition can be large
especially when the size of the matrix is large.
In this paper, we propose to use the
tangent space of the point in the manifold
to approximate the projection onto the manifold in order to reduce the computational cost.
We show that the sequence generated by the alternating projections onto the tangent spaces of the fixed rank matrices manifold and the nonnegative matrix manifold,
converge linearly to a point in the intersection of the two manifolds where the convergent point
is sufficiently close to optimal solutions.
Numerical examples will be presented to show that the computational time
of the proposed tangent space based method is less than that
of the original alternating projection method.
Moreover, experimental results in data clustering, pattern recognition and
hyperspectral data analysis, are given to demonstrate that
the performance of the proposed method is better than that of other nonnegative
matrix factorization methods in terms of computational time and accuracy.

The rest of this paper is organized as follows.
In Section \ref{sec:non}, we propose tangent space based alternating projection method.
In Section \ref{sec:conv}, we show the convergence of the proposed method.
In Section \ref{sec:experiment}, numerical examples are given to show the advantages of the proposed method.
Finally, some concluding remarks are given in Section \ref{sec:clu}.

\section{Nonnegative Low Rank Matrix Approximation}\label{sec:non}
In this paper, we are interested in the $m\times n$ fixed-rank matrices manifold
\begin{align}\label{v1}
\mathcal{M}_{r}:=\left\{ {\bf X} \in \mathbb{R}^{m\times n},~ \rank({\bf X})= r\right\},
\end{align}
the $m\times n$ non-negativity matrices manifold
\begin{align}
\mathcal{M}_{n}:=\left\{ {\bf X} \in \mathbb{R}^{m\times n},{\bf X}_{ij}\geq 0,~ i=1,\cdots,m,\ j=1,\cdots,n\right\},\label{v2}
\end{align}
and the $m\times n$ nonnegative fixed rank matrices manifold
\begin{align}\label{v3}
\mathcal{M}_{rn} =\mathcal{M}_{r} \cap \mathcal{M}_{n}
 =&\left\{ {\bf X} \in \mathbb{R}^{m\times n}, ~ \rank({\mathbf X})=r,~{\bf X}_{ij}\geq 0,\right.\nonumber\\
  &\left. ~i=1,...,m,~j=1,...,n\right\}.
\end{align}
The proof of $\mathcal{M}_{rn}$ is a manifold can be found in \cite{sm2019}.
Let ${\bf X}\in \R^{m\times n}$ be an arbitrary matrix in the manifold $\mathcal{M}_{r}$.
We set the singular value decomposition of ${\bf X}$ as follows:
${\bf X} ={\bf U} {\bf \Sigma} {\bf V}^{T}$ where
${\bf U} \in \R^{m \times r}$,
${\bf \Sigma} \in \R^{r \times r}$, and
${\bf V} \in  \R^{n \times r}$.
It follows from Proposition 2.1 in \cite{Vandereyckenl2013}
that the tangent space of $\mathcal{M}_{r}$ at ${\bf X}$ can be expressed as
\begin{equation}\label{tang1}
T_{\mathcal{M}_{r}}({\bf X})=\{ {\bf U} {\bf W}^{T} + {\bf Z} {\bf V}^{T}\},
\end{equation}
where ${\bf W} \in \mathbb{R}^{n\times r},{\bf Z} \in \mathbb{R}^{m\times r}$ are arbitrary.
Here $\cdot^T$ denotes the transpose of a matrix.
For a given $m$-by-$n$ matrix ${\bf Y}$, the orthogonal projection of ${\bf Y}$
onto the subspace $T_{\mathcal{M}_{r}}(X)$ can be written as
\begin{align}\label{pj2}
P_{T_{\mathcal{M}_{r}}({\bf X})}({\bf Y})= {\bf U} {\bf U}^{T} {\bf Y} + {\bf Y} {\bf V} {\bf V}^{T} -
{\bf U} {\bf U}^{T} {\bf Y} {\bf V} {\bf V}^{T}.
\end{align}
The alternating projection method studied in \cite{sm2019} is based on projecting the given nonnegative matrix onto the $m\times n$ fixed-rank matrices manifold $\mathcal{M}_{r}$ and the non-negativity matrices manifold $\mathcal{M}_{n}$ iteratively.
The projection onto the fixed rank matrix set $\mathcal{M}_{r}$ is derived by the Eckart-Young-Mirsky theorem \cite{golub2012matrix} which  can be expressed as follows:
\begin{align}\label{p1}
\pi_{1}({\bf X})=\sum_{i=1}^{r}\sigma_{i}({\bf X}) u_{i}({\bf X}) {v}_{i}^{T}({\bf X}),
\end{align}
where $\sigma_{i}({\bf X})$ are first $r$ singular values of ${\bf X}$, and $u_{i}({\bf X})$,
$v_{i}({\bf X})$ are their corresponding singular vectors.
The projection onto the nonnegative matrix set $\mathcal{M}_{n}$ is expressed as
\begin{align}\label{p2}
\pi_{2}({\bf X})=\left\{\begin{array}{c}
                     X_{ij}, ~~~ {\rm if} ~~ X_{ij}\geq 0, \\
                     0,     ~~~~~ {\rm if} ~~   X_{ij} < 0.
                   \end{array}
\right.
\end{align}
Moreover, $\mathcal{M}_{rn}$ refers to the nonnegative fixed rank matrices manifold given as \eqref{v3}, and $\pi({\bf X})$ refers to the closest matrix to the given nonnegative matrix ${\bf X}$, i.e.,
\begin{equation}\label{pj1}
\pi({\bf X}) = \underset{{\bf Y} \in {\cal M}_{rn}}{\operatorname{argmax}} \| {\bf X}-{\bf Y}  \|_F^2.
 \end{equation}

\subsection{Projections Based on Tangent Spaces}

\begin{figure}
\centering
\includegraphics[width=0.75\linewidth]{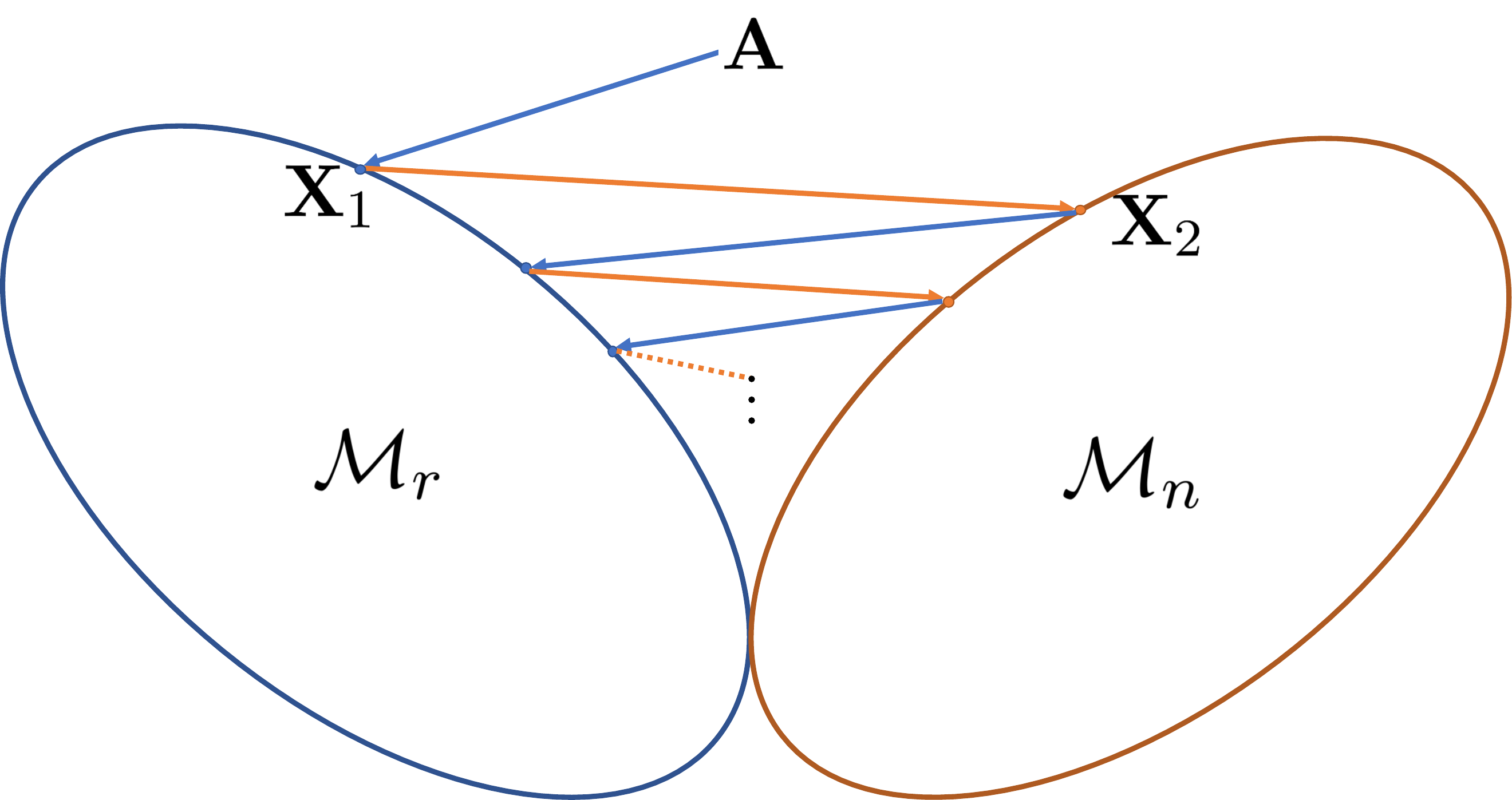}

\centerline{(a)}

\includegraphics[width=0.75\linewidth]{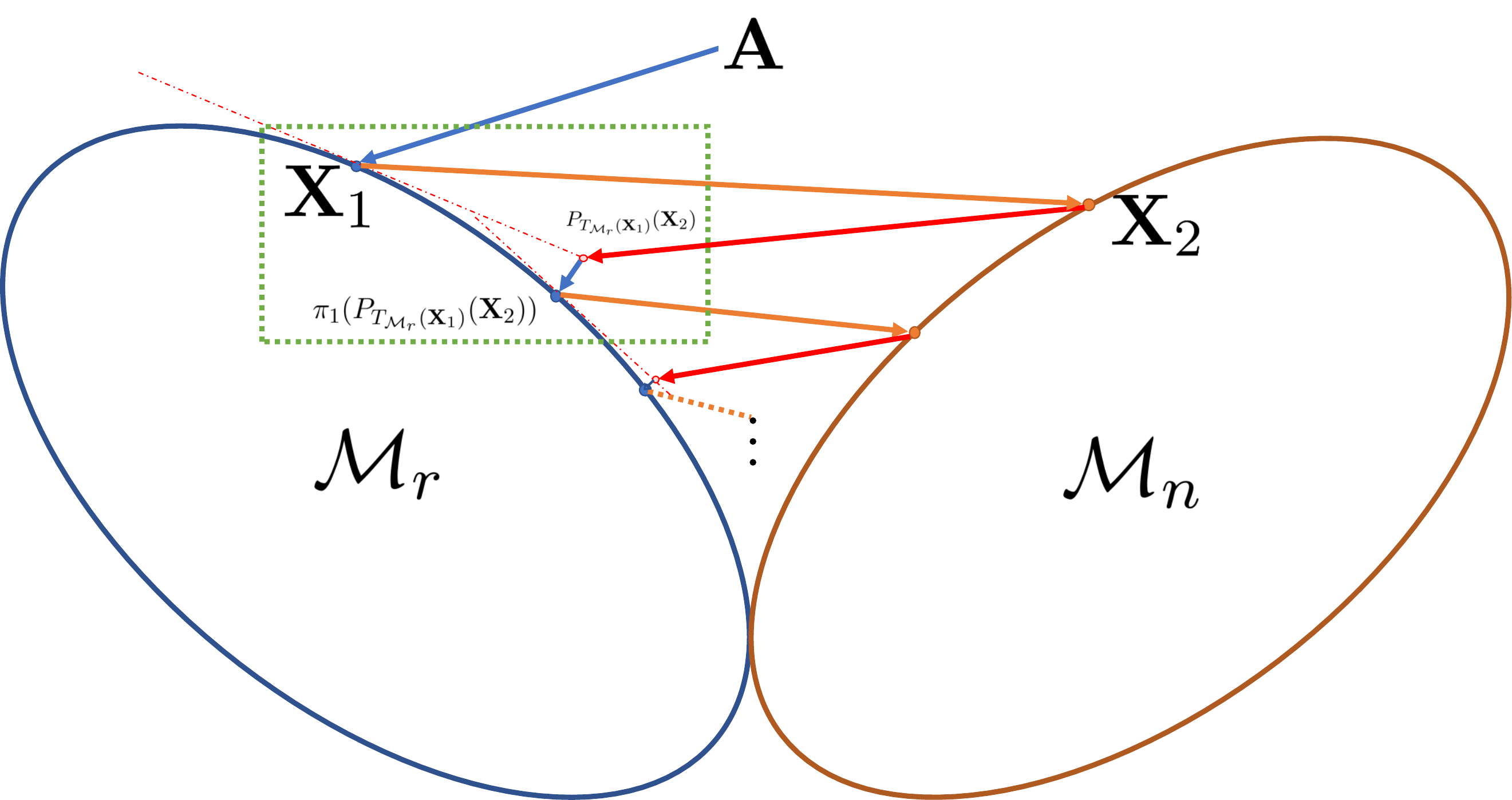}

\centerline{(b)}

\caption{The comparison between (a) the original alternating projection
method and (b) the proposed TAP method.} \label{figure1}
\end{figure}

\begin{figure}
\centering
\includegraphics[width=0.55\linewidth]{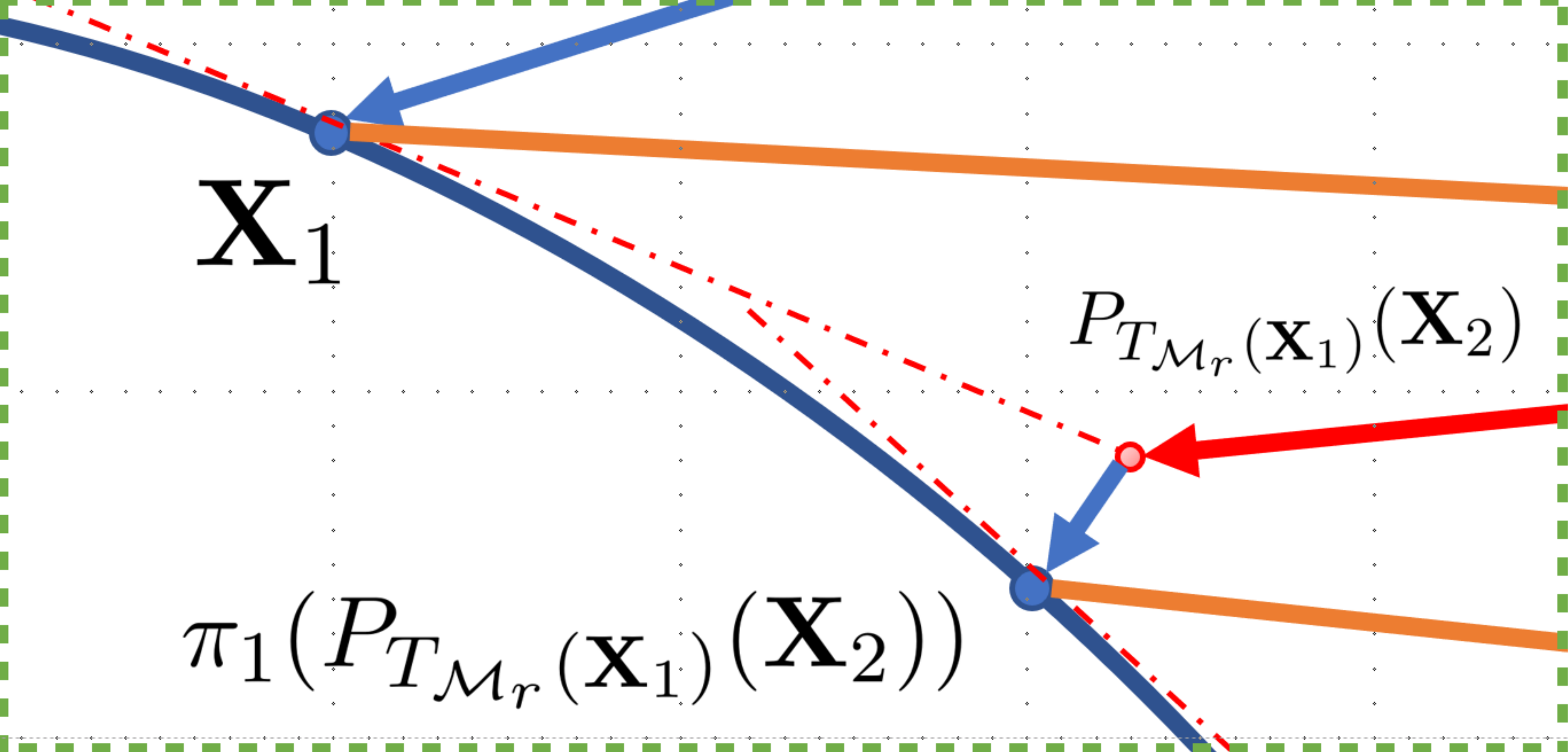}
\caption{The zoomed region in Figure 1(b).}
\label{figure1a}
\end{figure}

Note that it can be expensive to project a matrix onto the fixed rank manifold by using
the singular value decomposition.
In this paper, we make use of tangent spaces and construct {\tt T}angent space based {\tt A}lternating
{\tt P}rojection (TAP) method to find nonnnegative low rank matrix approximation such that
the computational cost can be reduced compared with the original alternating projection
method in \cite{sm2019}.
In Figure \ref{figure1} and Figure \ref{figure1a}, we demonstrate the proposed TAP method.
In the method,
the given nonnegative matrix ${\bf X_{0}}={\bf A}$ was first projected onto the manifold $\mathcal{M}_{r}$ to get a
point ${\bf X}_{1}$ by $\pi_{1}$, and then ${\bf X}_{2}$ is derived by projecting
${\bf X}_{1}$ onto the manifold $\mathcal{M}_{n}$ by $\pi_{2}$.
The first two steps are same as the original alternating projection
method \cite{sm2019}.
According to the third step, the point ${\bf X}_{2}$ is first projected onto the tangent space at
${\bf X}_{1}$ of the manifold $\mathcal{M}_{r}$ by the orthogonal projection $P_{T_{\mathcal{M}_{r}}({\bf X}_{1})}$,  and then the derived point is projected from the tangent space to the manifold ${\cal M}_r$
to get
${\bf X}_{3}$.
Thus the sequence can be derived as follows:
\begin{align*}
&{\bf X}_{0}={\bf A},~{\bf X}_{1}=\pi_{1}({\bf X}_{0}),~{\bf X}_{2}=\pi_{2}({\bf X}_{1}),\\
&{\bf X}_{3}=\pi_{1}(P_{T_{\mathcal{M}_{r}}({\bf X}_{1})} ({\bf X}_{2})),~{\bf X}_{4}=\pi_{2}({\bf X}_{3}),~\cdots,\\
& {\bf X}_{2k+1}=\pi_{1}(P_{T_{\mathcal{M}_{r}}({\bf X}_{2k-1)})} ({\bf X}_{2k})),~{\bf X}_{2k+2}=\pi_{2}({\bf X}_{2k+1}),~\cdots
\end{align*}
where $P_{T_{\mathcal{M}_{r}}({\bf X}_{2k-1)})} ({\bf X}_{2k})$ denotes the orthogonal projections of ${\bf X}_{2k}$ onto the tangent space of $\mathcal{M}_{r}$ at ${\bf X}_{2k-1}$.
The algorithm is summarized in Algorithm 1.

\begin{algorithm}[h]
\caption{Tangent spaces based Alternating Projection (TAP) Method} \label{ag1}
\textbf{Input:}
Given a nonnegative matrix ${\bf A}\in \mathbb{R}^{m\times n}$ this algorithm computes nearest rank-$r$ nonnegative matrix. \\
~~1:  Initialize ${\bf X}_{0}={\bf A}$;\\
~~2: ${\bf X}_{1}=\pi_{1}({\bf X}_{0})$ and ${\bf Y}_{1}=\pi_{2}({\bf X}_{1})$\\
~~3: for k=1,2,...,\\
~~4:  \quad ${\bf X}_{k+1}=\pi_{1}(P_{T_{\mathcal{M}_{r}}({\bf X}_{k})} ({\bf Y}_{k}));$\\
~~5:  \quad ${\bf Y}_{k+1}=\pi_{2}({\bf X}_{k+1});$\\
~~6: \textbf{end}\\
\textbf{Output:} ${\bf X}_k$ when the stopping criterion is satisfied.
\end{algorithm}

Let us analyze the computational cost of TAP method in each iteration.
Let
${\bf X}_{k}= {\bf U}_k {\bf \Sigma}_k {\bf V}_k^{T}$ be the skinny SVD decomposition of ${\bf X}_{k}$.  By \eqref{tang1}, the tangent space of $\mathcal{M}_{r}$ at ${\bf X}_{k}$ can be expressed as
\begin{align*}
T_{\mathcal{M}_{r}}&({\mathbf X}_{k})=\{ \mathbf U_k \mathbf W^{T} + {\bf Z} {\bf V}_k^{T} \},
\end{align*}
where ${\bf W} \in \mathbb{R}^{n\times r}, {\bf Z} \in \mathbb{R}^{m\times r} $are arbitrary.
According to (\ref{pj1}),
the orthogonal projection of ${\bf Y}_{k}$ onto the subspace $T_{\mathcal{M}_{r}({\bf X}_{k})}$
can be written as follows:
\begin{align*}
P_{T_{\mathcal{M}_{r}({\bf X}_{k})}}({\bf Y}_{k})= {\bf U}_k {\bf U}_k^{T}
{\bf Y}_{k} \hspace{-.3mm}+\hspace{-.3mm} {\bf Y}_{k} {\bf V}_k {\bf V}_k^{T} \hspace{-.3mm}-\hspace{-.3mm}
{\bf U}_k {\bf U}_k^{T} {\bf Y}_{k} {\bf V}_k {\bf V}_k^{T}.
\end{align*}
Now it is required to compute the SVD of the projected matrix
$P_{T_{\mathcal{M}_{r}({\bf X}_{k})}}({\bf Y}_{k})$ with smaller size.
Suppose that the QR decompositions of $({\bf I}-{\bf U}_k {\bf U}_k^{T}) {\bf Y}_{k} {\bf V}_k$
and $({\bf I}-{\bf V}_k {\bf V}_k^{T}) {\bf Y}_{k} {\bf U}_k$ are given as follows:
\begin{align*}
({\bf I}-{\bf U}_k {\bf U}_k^{T}) {\bf Y}_{k} {\bf V}_k &= {\bf Q}_k {\bf R}_k
\end{align*}
and
\begin{align*}
({\bf I}-{\bf V}_k {\bf V}_k^{T}) {\bf Y}^{T}_{k} {\bf U}_k = \hat{\bf Q}_k \hat{\bf R}_k,
\end{align*}
respectively.
Recall that ${\bf U}_k^{T} {\bf Q}_k= {\bf V}_k^{T} \hat{\bf Q}_k = {\bf 0}$
and then by a direct computation, we have
\begin{align*}
&~P_{T_{\mathcal{M}_{r}({\bf X}_{k})}}({\bf Y}_{k})\\
=&~
{\bf U}_k {\bf U}_k^{T} {\bf Y}_{k} ({\bf I}-{\bf V}_k {\bf V}_k^{T})
+({\bf I}-{\bf U}_k {\bf U}_k^{T}) {\bf Y}_{k} {\bf V}_k {\bf V}_k^T\\
 &~+
{\bf U}_k {\bf U}_k^{T} {\bf Y}_{k} {\bf V}_k {\bf V}_k^{T} \\
=&~{\bf U}_k \hat{\bf R}_k^{T} \hat{\bf Q}_k^{T} + {\bf Q}_k {\bf R}_k
{\bf V}_k^T + {\bf U}_k {\bf U}_k^T {\bf Y}_k {\bf V}_k {\bf V}_k^T \\
=&~\left(
     \begin{array}{cc}
       {\bf U}_k & {\bf Q}_k \\
     \end{array}
   \right)\left(
            \begin{array}{cc}
              {\bf U}_k^{T} {\bf Y}_{k} {\bf V}_k & \hat{\bf R}_k^{T} \\
              {\bf R}_k & {\bf 0} \\
            \end{array}
          \right)\left(
                      \begin{array}{c}
                        {\bf V}_k^T \\
                        \hat{\bf Q}_k^T \\
                      \end{array}
                    \right)\\
:=&~\left(
     \begin{array}{cc}
       {\bf U}_k & {\bf Q}_k \\
     \end{array}
   \right) {\bf M}_k \left(
                      \begin{array}{c}
                        {\bf V}_k^T \\
                        \hat{\bf Q}_k^T \\
                      \end{array}
                    \right).
\end{align*}
Let ${\bf M}_k= {\bf \Psi}_k {\bf \Gamma}_k {\bf \Phi}_k^T$
be the SVD of ${\bf M}_k$ which can be computed using $O(r^3)$ flops since
${\bf M}_k$ is a $2r\times 2r$ matrix.
Note that $\left(
              {\bf U}_k , {\bf Q}_k
          \right) $
  and $\left(
              {\bf V}_k , \hat{\bf Q}_k
          \right)$
  are orthogonal, then the SVD of $P_{T_{\mathcal{M}_{r}({\bf X}_{k})}}({\bf Y}_{k})=
  {\bf \Omega}_k {\bf \Theta}_k {\bf \Upsilon}_k^T$
  can be computed by
 \begin{align*}
 {\bf \Omega}_k =\left(
              {\bf U}_k , {\bf Q}_k
          \right) {\bf \Psi}_k,
          {\bf \Theta}_k = {\bf \Gamma}_k ~ \text{and} ~
 {\bf \Upsilon}_k=\left(
              {\bf V}_k , \hat{\bf Q}_k
          \right) {\bf \Phi}_k.
 \end{align*}
 It follows that  the overall computational cost of $\pi_{1}(P_{T_{\mathcal{M}_{r}({\bf X}_{k})}}
 ({\bf Y}_{k}))$ can be expressed as two matrix-matrix multiplications.
 In addition, the calculation procedure involves
 the QR decomposition of two matrices of sizes $m\times r$ and $n\times r$ matrices,
 and the SVD of a matrix of size $2r\times 2r$.
 The total cost per iteration is of
  $4mnr+O(r^2m+r^2n+r^3)$. In contrast,
  the computation of the best rank-$r$ approximation of a non-structured $m\times n$ matrix
  costs $O(mnr)+mn$ flops where the constant in front of $mnr$ can be very large. In practice,
  the cost per iteration of the proposed TAP method is less than that of original alternating projection method.
  In Section 4, numerical examples will be given to demonstrate the total computational time
  of the proposed TAP method is less than that of the original alternating projection method.

  \section{The Convergence Analysis}\label{sec:conv}

  In this section, we study the convergence of the proposed TAP method in Algorithm 1.
  We note that the convergence result of the original alternating projection method has been
  established in \cite{andersson2013alternating}. The key concept is the angle of a point in the
  intersection of two manifolds.
  In our setting, the angle $\alpha({\bf B})$ of ${\bf B} \in \mathcal{M}_{rn}$
where
\begin{align}\label{df1}
\alpha( {\bf B} )=cos^{-1}(\sigma( {\bf B} ))
\end{align}
and
\begin{align*}
\sigma( {\bf B} )=\lim_{\xi\rightarrow 0} \sup_{{\bf B}_{1}\in F^{\xi}_{1}({\bf B}),
{\bf B}_{2}\in F^{\xi}_{2}({\bf B})}
\left\{\frac{\left< {\bf B}_{1}- {\bf B}, {\bf B}_{2}- {\bf B} \right>}
{\| {\bf B}_{1}\hspace{-.3mm}-\hspace{-.3mm} {\bf B} \|_{F}\| {\bf B}_{2}\hspace{-.3mm}-\hspace{-.3mm} {\bf B}\|_{F}}\right\},
\end{align*}
with
\begin{align*}
F_{1}^{\xi}( {\bf B})=
\{ {\bf B}_1 \ | \ {\bf B}_1 \in \mathcal{M}_{r}\backslash {\bf B},&
\| {\bf B}_1- {\bf B} \|_{F}\leq \xi,\\& {\bf B}_{1}- {\bf B} \bot T_{\mathcal{M}_{r}\cap
\mathcal{M}_{n}}( {\bf B})
\},
\end{align*}
\begin{align*}
F_{2}^{\xi}( {\bf B})=
\{
{\bf B}_2 \ |\ {\bf B}_2 \in \mathcal{M}_{n}\backslash {\bf B},&
\| {\bf B}_2- {\bf B} \|_{F}\leq \xi,\\& {\bf B}_{2}- {\bf B} \bot T_{\mathcal{M}_{r}\cap
\mathcal{M}_{n}}({\bf B})
\},
\end{align*}
and
$T_{\mathcal{M}_{r}\cap \mathcal{M}_{n}}( {\bf B})$ is
the tangent space of $\mathcal{M}_{r} \cap \mathcal{M}_{n}$ at point ${\bf B}$.
The angle is calculated based on the two points
belonging $\mathcal{M}_r$ and $\mathcal{M}_n$ respectively. A point ${\bf B}$ in $\mathcal{M}_{rn}$
is nontangential if $\alpha({\bf B})$ has a positive angle, i.e.,
$0 \le \sigma(A) < 1$.

In the following, we list the main convergence results of Algorithm 1 that
has been studied in the literature.

\begin{theorem}\label{thm_convergence}
Let $\mathcal{M}_{r}$, $\mathcal{M}_{n}$ and $\mathcal{M}_{rn}$ be given as \eqref{v1}, \eqref{v2} and \eqref{v3}, the projections onto $\mathcal{M}_{r}$ and $\mathcal{M}_{n}$ be given as \eqref{p1}-\eqref{p2}, respectively.
Suppose that ${\bf P} \in \mathcal{M}_{rn}$ is a non-tangential intersection point, then for any
given $\epsilon>0$ and $1>c>\sigma({\bf P})$, there exist an
$\xi>0$ such that for any ${\bf A}\in Ball({\bf P},\xi)$
(the ball neighborhood of ${\bf P}$ with radius $\xi$ contains the given
nonnegative matrix ${\bf A}$), the sequence ${\bf X}_k$ generated by Algorithm 1
converges to a point ${\bf X}_\infty \in {\cal M}_{rn}$, and satisfy
 \begin{enumerate}[(1)]
\item $\| {\bf X}_\infty - \pi({\bf A}) \|_{F} \leq \epsilon \| {\bf A} - \pi({\bf A}) \|_{F}$,
\item $\| {\bf X}_\infty - {\bf X}_k \|_{F} \leq {\rm const} \cdot c^k \| {\bf A} - \pi({\bf A}) \|_{F}$,
  \end{enumerate}
where $\pi({\bf A})$ is defined in (\ref{pj1}).
\end{theorem}

Here we first present and establish some preliminary results and then give the proof of Theorem 3.1.
These results are used to estimate the approximation when tangent spaces are used in the projections.

\begin{lemma}[Proposition 4.3 and Theorem 4.1 in \cite{andersson2013alternating}]\label{jia2}
Let ${\bf P} \in \mathcal{M}_{r}$ be given and $\pi_{1}, \pi$ be defined as \eqref{p1} and \eqref{pj1}.
For each $0<\epsilon<\frac{3}{5}$, there exist an $s({\epsilon})>0$ and  an $\varepsilon(\epsilon) >0,$
such that for any given ${\bf Z} \in Ball({\bf P},s({\epsilon}))$,
\begin{align} \label{lemma31}
\|\pi_{1}({\bf Z})-P_{T_{\mathcal{M}_{r}}(\pi({\bf Z}))}({\bf Z})\|_{F}<4\sqrt{\epsilon}
\| {\bf Z}-\pi( {\bf Z})\|_{F},
\end{align}
and
\begin{align}\label{an4}
\|\pi(\pi_{1}({\bf Z}))-\pi({\bf Z})\|_{F}< \varepsilon(\epsilon) \| {\bf Z}-\pi({\bf Z})\|_{F}.
\end{align}
\end{lemma}

\begin{lemma}[Proposition 2.4 in \cite{andersson2013alternating}]\label{newproof1}
Let $ {\bf P}\in \mathcal{M}_{r}$ be given. For each $\epsilon >0,$ there exists $s>0$ such that for all
${\bf C}\in Ball({\bf P},s)\cap \mathcal{M}_{r},$ we have:\\
$(i)~\min_{{\bf D}'\in T_{\mathcal{M}_{r}({\bf C})}}\|{\bf D}-{\bf D}'\|_{F}\leq \epsilon \|{\bf D}-{\bf C}\|_{F},~\forall~{\bf D}\in Ball({\bf P},s)\cap \mathcal{M}_{r}.$\\
$(ii)~ \|{\bf D}-\pi_{1}({\bf D})\|_{F}\leq \epsilon \|{\bf D}-{\bf C}\|_{F},~\forall~ {\bf D}\in Ball({\bf P},s)\cap T_{\mathcal{M}_{r}}({\bf C})$.
\end{lemma}

Next we can estimate the distance with respect to the other point in $\mathcal{M}_{r}$
instead of using $\pi({\bf Z})$. The following results which are proved in Appendix \ref{Proof1}
are needed.
\begin{lemma}\label{ne1}
Let ${\bf P} \in \mathcal{M}_{r}$ be given, $P_{T_{\mathcal{M}_{r}}}$ and $\pi_{1}$ be given as  \eqref{pj2} and \eqref{p1}.  For each $0<\epsilon<\frac{3}{5}$, there exist an $s({\epsilon})>0$ and a point ${\bf Q}\in Ball({\bf P},s({\epsilon}))\cap \mathcal{M}_{r}$
such that for any given
${\bf Z} \in Ball({\bf P},s({\epsilon}))$, we have
\begin{align} \label{lemma32}
\|\pi_{1}({\bf Z})-P_{T_{\mathcal{M}_{r}}({\bf Q})}({\bf Z})\|_{F}<4\sqrt{\epsilon}\| {\bf Z} - {\bf Q}
\|_{F}.
\end{align}
\end{lemma}

\begin{lemma}[Theorem 4.5 in \cite{andersson2013alternating}]\label{an5}
Suppose ${\bf P}$ is a nontangential point with $\sigma({\bf P}) < c$.
Then there exists an $s>0$ such that for all ${\bf Z} \in \mathcal{M}_{n}\cap Ball({\bf P},s)$,
we have
\begin{align}\label{eq3}
\|\pi_{1}({\bf Z})-\pi({\bf Z})\|_{F}<c\| {\bf Z}-\pi({\bf Z})\|_{F}.
\end{align}
\end{lemma}
Next we would like to estimate the distance
between $\pi(\pi_{1}(P_{T_{\mathcal{M}_{r}}({\bf Q})}({\bf Z})))$ and $\pi({\bf Z})$
where they are on the manifold $\mathcal{M}_{rn}$.

\begin{lemma}\label{new1}
Let ${\bf P} \in \mathcal{M}_{rn}$ be given.
For each $0<\epsilon<\frac{3}{5}$, there exist $\varepsilon_1({\epsilon})>0,$
$\varepsilon_2({\epsilon}) >0$ and $s_1({\epsilon})>0$
such that for all ${\bf Z} \in Ball({\bf P},s_1({\epsilon}))$,
\begin{align*}
\|\pi(\pi_{1}(P_{T_{\mathcal{M}_{r}}({\bf Q})}({\bf Z})))-\pi({\bf Z})\|_{F}
 \leq &\varepsilon_1({\epsilon}) \| {\bf Z}-\pi({\bf Z})\|_{F}\\
 &~~~~~+ \varepsilon_2({\epsilon}) \|
{\bf Q}-\pi({\bf Z})\|_{F},
\end{align*}
where ${\bf Q} \in \mathcal{M}_{r} \cap Ball({\bf P},s_1({\epsilon}))$.
\end{lemma}

The proof of Lemma \ref{new1} can be found in Appendix \ref{Proof2}.
 According to Lemma \ref{new1},
if ${\bf Z}=\pi_{2}({\bf Q})$,
then
$$\|{\bf Z}-{\bf Q}\|_{F}=\|{\bf Q}-\pi_{2}({\bf Q})\|_{F}\leq \|{\bf Q}-\pi({\bf Z})\|_{F}$$
by noting that  $\pi({\bf Z}) \in \mathcal{M}_{n}$, and
\begin{align*}
&~\|\pi(\pi_{1}(P_{T_{\mathcal{M}_{r}}({\bf Q})}({\bf Z})))-\pi({\bf Z})\|_{F} \\
\leq &~ 8\alpha\sqrt{\epsilon}\| {\bf Z}-{\bf Q}\|_{F}+\varepsilon({\epsilon})\|{\bf Z}-\pi({\bf Z})\|_{F}\\
\leq &~\varepsilon_1({\epsilon}) \| {\bf Z}-\pi({\bf Z})\|_{F}+
\varepsilon_2({\epsilon}) \|
{\bf Q}-\pi({\bf Z})\|_{F},
\end{align*}
where
$\varepsilon_1({\epsilon})=\varepsilon({\epsilon})$ and $\varepsilon_2({\epsilon})
=8\alpha\sqrt{\epsilon}$.

In order to prove the convergence of Algorithm \ref{ag1}, we need to estimate the distance
between $\pi_{1}(P_{T_{\mathcal{M}_{r}}({\bf Q})}({\bf Z}))$ and $\pi({\bf Z})$. The proof can be found in Appendix \ref{Proof3}.

\begin{lemma}\label{new2}
Suppose ${\bf P}$ is a nontangential point in ${\cal M}_{rn}$ with $\sigma({\bf P}) < c$,
and ${\bf Q}\in \mathcal{M}_{r}$. Then
there exists an $s>0$ such that
when ${\bf Z}=\pi_{2}({\bf Q})\in \mathcal{M}_{n}\cap Ball({\bf P},s)$  and
$P_{T_{\mathcal{M}_{r}}({\bf Q})}({\bf Z})\in T_{\mathcal{M}_{r}}({\bf Q})
\cap Ball({\bf P},s)$, we have
\begin{align}\label{eq4}
\|\pi_{1}(P_{T_{\mathcal{M}_{r}}({\bf Q})}({\bf Z}))-\pi({\bf Z})\|_{F}<c\|{\bf Z}-\pi({\bf Z})\|_{F}.
\end{align}
\end{lemma}

With the above preliminaries, we give the proof of Theorem \ref{thm_convergence}.

\vspace{2mm}
\noindent
{\it Proof of Theorem 3.1}
Suppose that $\epsilon<1$, and set $\sigma({\bf P})<c<1$ and
\begin{align*}
\varepsilon =\frac{1-c}{2(3-c)}\epsilon,~~\varepsilon_{2}(\epsilon)=\frac{1-c}{2+2\alpha}\epsilon,
\end{align*}
where $\alpha$ is a constant given as in \eqref{continuous}.
It follows Lemma \ref{an5}-\ref{new2} that there exist some possibly distinct radii that  guarantee \eqref{eq3}-\eqref{eq4} are satisfied. Let $s$
 denote the minimum of these possibly  radii and pick
 $r<\frac{s(1-\epsilon)}{4(2+\epsilon)}$, so that $\pi(Ball({\bf P},r))\subseteq Ball({\bf P},\frac{s}{4})$.
Then $\|\pi({\bf A})- {\bf P}\|_{F}<\frac{s}{4}$ follows from the latter condition.
Denote $l=\| {\bf A}-\pi({\bf A})\|_{F}$ and note that
$$
l=\| {\bf A}-{\bf P}+{\bf P}-\pi({\bf A})\|_{F}\leq \| {\bf A}- {\bf P} \|_{F}+
\| {\bf P}-\pi({\bf A})\|_{F}\leq r+\frac{s}{4}.
$$
As $\pi({\bf A})\in \mathcal{M}_{rn}$ and note that ${\bf X}_{1}=\pi_{1}({\bf A})$,
we have
\begin{align*}
\|{\bf X}_{1}-{\bf A}\|_{F}=\|\pi_{1}({\bf A})-{\bf A}\|_{F}\leq \|\pi({\bf A})-{\bf A}\|_{F}=l
\end{align*}
and
\begin{align*}
\|{\bf X}_{1}-\pi({\bf X}_{1})\|_{F}\leq&~ \| {\bf X}_{1}-\pi({\bf A})\|_{F}\\
\leq &~
\| {\bf X}_{1}- {\bf A}\|_{F}+\| {\bf A}-\pi({\bf A})\|_{F}\leq 2l.
\end{align*}
In order to prove $\{ {\bf X}_{k}\}$ derived by Algorithm \ref{ag1} is convergent, we need to prove
$\{ {\bf X}_{k}\}$ is a Cauchy sequence.
By Lemma \ref{new2}, there exist an $c_{1}$ such that
\begin{align}\label{jia31}
\| {\bf X}_{2k+1}-\pi( {\bf X}_{2k+1})\|_{F}
\leq &~\|{\bf X}_{2k+1}-\pi( {\bf X}_{2k})\|_{F}\nonumber\\
\leq &~ c_{1} \| {\bf X}_{2k}-\pi( {\bf X}_{2k})\|_{F}.
\end{align}
In addition, by Lemma \ref{an5}, there exist an $c_{2}$ such that
 \begin{align}\label{jia32}
\| {\bf X}_{2k}-\pi( {\bf X}_{2k})\|_{F}\leq &~\| {\bf X}_{2k}-\pi( {\bf X}_{2k-1})\|_{F}\nonumber\\
\leq &~c_{2}\| {\bf X}_{2k-1}-\pi( {\bf X}_{2k-1})\|_{F}.
\end{align}
Set $c=\max\{c_{1},c_{2}\}$, combine \eqref{jia31} and \eqref{jia32} together gives
 \begin{align}\label{jia3}
\| {\bf X}_{k}-\pi( {\bf X}_{k})\|_{F}\leq c \| {\bf X}_{k-1}-\pi( {\bf X}_{k-1})\|_{F}.
\end{align}
Then  $\{ {\bf X}_{k}\}$ is a Cauchy sequence if and only if
\begin{align}\label{m1}
\{ {\bf X}_{k}\}_{k=1}^{\infty}\subseteq Ball({\bf P},s)
\end{align}
 is satisfied. The remaining task is to show \eqref{m1} is satisfied by induction.
 For $k=1$,
\begin{align*}
\| {\bf X}_{1}- {\bf P}\|_{F}\leq &~\| {\bf X}_{1}- {\bf A} \|_{F}+\| {\bf A}- {\bf P} \|_{F}\leq l+
\frac{r}{2}\\
\leq&~ 2r+\frac{s}{4}\leq \frac{s(1-\epsilon)}{2(2+\epsilon)}+\frac{s}{4}<s.
\end{align*}
Assume that \eqref{m1} is satisfied when $n=k$, then it follows from \eqref{jia3} that
\begin{align}\label{jia5}
\| {\bf X}_{k}-\pi( {\bf X}_{k})\|_{F}\leq c^{k}\| {\bf X}_{1}-\pi( {\bf X}_{1})\|_{F}\leq 2lc^{k}.
\end{align}
For an arbitrary $k$ and $i=1$ or $2$, we have
\begin{align}
 &~\| {\bf X}_{k-2}-\pi( {\bf X}_{k-1})\|_{F}\nonumber \\
 =&~\| {\bf X}_{k-2}-\pi(\pi_{i}( {\bf X}_{k-2}))\|_{F}\nonumber \\
 =&~\| {\bf X}_{k-2}-\pi( {\bf X}_{k-2})+\pi( {\bf X}_{k-2})-\pi(\pi_{i}( {\bf X}_{k-2}))\|_{F}\nonumber\\
 \leq &~\| {\bf X}_{k-2}-\pi({\bf X}_{k-2})\|_{F}+\|\pi({\bf X}_{k-2})-\pi(\pi_{i}({\bf X}_{k-2}))\|_{F}\nonumber\\
 \leq &~ \|{\bf X}_{k-2}-\pi({\bf X}_{k-2})\|_{F}+\alpha\|{\bf X}_{k-2}-\pi_{i}(X_{k-2})\|_{F}\nonumber\\
 \leq &~(1+\alpha)\|{\bf X}_{k-2}-\pi({\bf X}_{k-2})\|_{F}.\nonumber
\end{align}
The second part of the second inequality follows by the continuous of $\pi,$ the third inequality follows by
$$\|{\bf X}_{k-2}-\pi_{i}({\bf X}_{k-2})\|_{F}\leq\|{\bf X}_{k-2}-\pi({\bf X}_{k-2})\|_{F}, ~~i=1,2.$$
In addition, when $k$ is  even, by lemma \ref{jia2}, we have
\begin{align}\label{new10}
\|\pi( {\bf X}_{k})-\pi( {\bf X}_{k-1})\|_{F}<\varepsilon(\epsilon) \| {\bf X}_{k-1}-\pi({\bf X}_{k-1})\|_{F}.
\end{align}
When $k$ is odd, applying Lemma \ref{new1} gives
\begin{align}
&~\|\pi( {\bf X}_{k})-\pi( {\bf X}_{k-1})\|_{F}\nonumber\\
<&~\varepsilon_{1}(\epsilon) \| {\bf X}_{k-1}-\pi( {\bf X}_{k-1})\|_{F}+\varepsilon_{2}(\epsilon) \| {\bf X}_{k-2}-\pi(
{\bf X}_{k-1})\|_{F}\nonumber\\
<&~\varepsilon_{1}(\epsilon) \| {\bf X}_{k-1}-\pi( {\bf X}_{k-1})\|_{F}\nonumber\\
&~+\varepsilon_{2}(\epsilon)(1+\alpha) \|
{\bf X}_{k-2}-\pi( {\bf X}_{k-2})\|_{F}\nonumber\\
\leq&~ 2\varepsilon_{1}(\epsilon)c^{k-1}l+2\varepsilon_{2}(\epsilon)(1+\alpha)c^{k-2}l \nonumber\\
=&~(\varepsilon_{1}(\epsilon)c+\varepsilon_{2}(\epsilon)(1+\alpha))2c^{k-2}l.\nonumber
\end{align}
Set $\varepsilon=\max\{\varepsilon(\epsilon),\varepsilon_{1}(\epsilon)\},$ then for an arbitrary $k$, we have
\begin{align}\label{new4}
\|\pi( {\bf X}_{k})-\pi( {\bf X}_{k-1})\|_{F}\leq (\varepsilon c+\varepsilon_{2}(\epsilon)(1+\alpha))2c^{k-2}l.
\end{align}
By combining \eqref{new4} and using Lemma \ref{jia2}, we obtain
\begin{align}
& \|\pi( {\bf X}_{k})-\pi( {\bf A})\|_{F} \nonumber \\
\leq &~ \|\pi( {\bf A})-\pi( {\bf X}_{1})\|_{F}+\|
\pi( {\bf X}_{2})-\pi( {\bf X}_{1})\|_{F}\nonumber\\
&~+\|\sum_{j=3}^{k}\pi( {\bf X}_{j})-\pi( {\bf X}_{j-1})\|_{F} \nonumber\\
\leq &~\varepsilon l+2\varepsilon l+\sum_{j=3}^{k}(\varepsilon_{1}(\epsilon)c+\varepsilon_{2}(\epsilon)(1+\alpha))2c^{j-2}l \nonumber\\
\leq &~ 3\varepsilon l+\frac{2(\varepsilon_{1}(\epsilon)c+\varepsilon_{2}(\epsilon)(1+\alpha))}{1-c}l \nonumber\\
=&~\frac{3\varepsilon(1-c)+2\varepsilon c+(1+\alpha)\varepsilon_{2}(\epsilon)}{1-c}l
\leq\epsilon l.\label{jia4}
\end{align}
Thus,
\begin{align*}
\|{\bf P}- {\bf X}_{k}\|_{F}&\leq \| {\bf P}-\pi( {\bf A})\|_{F}+\|\pi({\bf A})-\pi({\bf X}_{k})\|_{F}\\
&+\|\pi({\bf X}_{k})-{\bf X}_{k}\|_{F}\leq s/4+  \epsilon l  +2l<s,
\end{align*}
which shows that \eqref{m1} is satisfied.

It follows from \eqref{new4} that the sequence $(\pi( {\bf X}_{k}))_{k=1}^{\infty}$ is a Cauchy sequence which converges to a point ${\bf Z}_{\infty}$.
Note that \eqref{jia5} is satisfied, the sequence $( {\bf X}_{k})_{k=1}^{\infty}$  also converges. In addition, ${\bf Z}_{\infty}=\pi( {\bf Z}_{\infty})$ can be derive by noting that
the projection is local continuous.
Moreover, by  taking the limit of \eqref{jia4} we can get $(i).$ For $(ii).$ Note that
\begin{align*}
\|\pi( {\bf X}_{k})- {\bf X}_{\infty}\|_{F}\leq&  \sum_{j=k+1}^{\infty}\|\pi( {\bf X}_{j})-\pi( {\bf X}_{j-1})\|_{F}\\
\leq&  \frac{2l\varepsilon c^{k}}{1-c}+\frac{2(1+\alpha)l\varepsilon_{2}(\epsilon) c^{k-1}}{1-c},
\end{align*}
and combine with \eqref{jia5}, we can get
\begin{align*}
 \| {\bf X}_{k}- {\bf X}_{\infty}\|_{F} \nonumber
 \leq &~\| {\bf X}_{k}-\pi( {\bf X}_{k})\|_{F}+\|\pi( {\bf X}_{k})- {\bf X}_{\infty}\|_{F}\\
 \leq &~\left(2 l +\frac{2 l\varepsilon }{1-c}+\frac{2(1+\alpha)l\varepsilon_{2}(\epsilon)}{1-c}\right) c^{k}\\
 = &~\beta c^{k} l,
\end{align*}
with a constant $\beta$ as desired.

\section{Experimental Results}\label{sec:experiment}

The main aim of this section is to
demonstrate that (i)
the computational time requried by the proposed TAP method is faster than
that by the original alternating projection (AP) method with about the same
approximation; (ii)
the performance of the proposed TAP method is better than that of nonnegative matrix factorization methods in terms of computational time and accuracy
for examples in data clustering, pattern recognition and hyperspectral data
analysis.

The experiments in Subsection \ref{sec4.1} are performed under Windows 7 and MATLAB R2018a
running on a desktop (Intel Core i7, @ 3.40GHz, 8.00G RAM) and experiments in Subsections \ref{sec4.2}-\ref{sec4.5} are performed under Windows 10 and MATLAB R2020a
running on a desktop (AMD Ryzen 9 3950, @ 3.49GHz, 64.00G RAM).

\subsection{The First Experiment}\label{sec4.1}
\begin{table*}[!t]
\renewcommand\arraystretch{1.0}\setlength{\tabcolsep}{6pt}
\centering
\caption{The relative approximation error  and computation time  on the synthetic data sets.  The \textbf{best} values  are respectively highlighted by bolder fonts.}\label{table1a}
\begin{tabular}{lcccccc}
\toprule
&     \multicolumn{6}{c}{200-by-200 matrix}\\
 \cmidrule(r){2-7}
 &\multicolumn{3}{c}{Relative approximation error }& \multicolumn{3}{c}{Computation time}\\
  \cmidrule(r){2-7}
Method         & $r=10$            & $r=20$      & $r=40$ & $r=10$            & $r=20$               & $r=40$ \\
\midrule
 TAP   &  {\bf 0.4576}          &  {\bf 0.4161}               & {\bf 0.3247} & {\bf  0.42}          &  {\bf 0.48}               & {\bf 0.38}  \\
\midrule
 AP  &  {\bf 0.4576}          &  {\bf 0.4161}               & {\bf 0.3247} &  0.66          &  0.66               & 0.42\\
\midrule
 A-MU:mean  &  0.4592              &  0.4249      &     0.3733   &  8.32              & 9.54                  & 15.34\\
A-MU:range & [0.4591, 0.4593] & [0.4246, 0.4251] & [0.3729, 0.3737] & [8.00, 8.81]     &  [9.41, 9.61]         &  [14.72, 15.75] \\
\midrule
A-HALS:mean    &      0.4591          &    0.4246                    &  0.3717 &     1.09            &    1.95                   &   4.01\\
A-HALS:range   &     [0.4590, 0.4593] &     [0.4244, 0.4247]         &  [0.3714, 0.3719]  &    [0.98, 1.22]    &   [1.86, 2.05]         &  [3.86, 4.13]\\
\midrule
A-PG:mean    &  0.4591               & 0.4244                     & 0.3717&  14.77             & 16.24                      &  21.52  \\
A-PG:range   & [0.4590, 0.4592]       &  [0.4243, 0.4246]         &  [0.3715, 0.3719]  &  [14.50,15.03]     &  [15.81, 16.55]        & [21.02, 21.77]\\

\bottomrule
 \multirow{3}{*}{Method} &     \multicolumn{6}{c}{400-by-400 matrix}\\
 \cmidrule(r){2-7}
 &\multicolumn{3}{c}{Relative approximation error }& \multicolumn{3}{c}{Computation time}\\
  \cmidrule(r){2-7}
           & $r=20$                & $r=40$               & $r=80$   & $r=20$                & $r=40$               & $r=80$  \\
\midrule
 TAP    & {\bf 0.4573}                   & {\bf 0.4161}                     & {\bf 0.3421}  & {\bf 1.55}                 & {\bf 1.32}                     &  {\bf 1.10}  \\
\midrule
 AP   & {\bf 0.4573}                   & {\bf 0.4161}                     & {\bf 0.3421}    & 2.95                  &2.47                      & 1.68   \\
\midrule
 A-MU:mean   &  0.4606              & 0.4301                           &  0.3857                   &  37.80              & 38.72                          & 46.41\\
A-MU:range   &  [0.4605, 0.4607]     & [0.4300, 0.4302]                 & [0.3856, 0.3860]         &  [36.67, 39.03]   &  [38.21, 39.18]              & [45.87, 48.28]  \\
\midrule
A-HALS:mean     &0.4604                    &0.4295                   &0.3836    & 3.10                   &7.40                   & 19.67 \\
A-HALS:range   &[0.4603, 0.4605]          &[0.4294, 0.4296]         &[0.3833, 0.3838]   & [3.03, 3.25]         &[7.12, 7.60]         & [19.04, 20.61]\\
\hline
A-PG:mean     &0.4604                &0.4297                         &0.3850   & 51.68              &60.80                        &61.95 \\
A-PG:range   &[0.4604, 0.4605]      &[0.4296, 0.4298]               &[0.3847, 0.3853]   & [51.04, 52.26]   & [60.62, 61.01]            & [61.34, 62.64] \\
\toprule
 \multirow{3}{*}{Method} &     \multicolumn{6}{c}{800-by-800 matrix}\\
 \cmidrule(r){2-7}
 &\multicolumn{3}{c}{Relative approximation error }& \multicolumn{3}{c}{Computation time}\\
 \cmidrule(r){2-7}
           & $r=40$             & $r=80$                &   $r=160$ & $r=40$             & $r=80$                &   $r=160$ \\
\midrule
 TAP    &{\bf 0.4550}                    &{\bf 0.4144}                      &{\bf 0.3412}  &{\bf 7.14}                    &{\bf 4.84}                      &{\bf 4.80}    \\
\midrule
 AP    &{\bf 0.4550}                    &{\bf 0.4144}                      &{\bf 0.3412}   &15.84                   &9.55                      &7.11 \\
\midrule
 A-MU:mean   &0.4608                 &0.4350                       &0.3984 & 60.29               &60.96                      &61.31      \\
A-MU:range &[0.4607, 0.4609]       &[0.4349, 0.4351]             &[0.3982, 0.3986]   & [60.03, 60.65]     &[60.61, 61.58]           &[60.72, 61.94]     \\
\midrule
A-HALS:mean    &0.4605                       &0.4336                              &0.3984  &18.54                      &47.70                                 &61.30   \\
A-HALS:range    &[0.4604,0.4605]              &[0.4335, 0.4336]                     &[0.3982, 0.3986]  &[17.76, 19.48]           &[43.33, 52.75]                      & [60.71, 61.91] \\
\midrule
A-PG:mean    &0.4606                    &0.4343                                 & 0.4007  & 60.38                  & 61.26                               &  61.76 \\
A-PG:range   &[0.4606, 0.4607]          &[0.4342, 0.4344]                       & [0.4005, 0.4012]  &[60.12, 60.79]         &[60.78, 61.81]                     & [61.29 62.48]\\
\bottomrule

\end{tabular}
\end{table*}
In the first experiment, we randomly generated $n$-by-$n$ nonnegative matrices ${\bf A}$
where their matrix entries follow a uniform distribution in between 0 and 1.
We employed the proposed TAP method and the original alternating projection (AP) method \cite{sm2019}
to test the relative approximation error $\| {\bf A} - {\bf X}_{c} \|_F / \| {\bf A} \|_F$,
where ${\bf X}_c$ are the computed rank $r$ solutions by different methods.
For comparison, we also list the results by nonnegative matrix factorization
algorithms: A-MU \cite{gillis2012accelerated}, A-HALS \cite{gillis2012accelerated} and A-PG \cite{lin2007projected}.

Tables \ref{table1a}  shows the relative approximation error of the computed solutions from the proposed TAP method and the other testing methods
for synthetic data sets of sizes 200-by-200, 400-by-400 and 800-by-800.
Note that there is no guarantee that
other testing NMF algorithms can determine the underlying nonnegative low rank factorization.
In the tables, it is clear that the testing NMF algorithms cannot obtain the underlying low rank factorization.
One of the reason may be that NMF algorithms can be sensitive to initial guesses.
In the tables, we illustrate this phenomena by displaying
the mean relative approximation error and the range containing both the minimum and the maximum relative approximation errors
by using ten initial guesses randomly generated.
We find in the table that the relative approximation errors computed by the TAP method is the same as those
by the AP method. It implies that the proposed TAP method can achieve the same accuracy of classical
alternating projection. According to the tables, the relative approximation errors  by
both TAP and AP methods are always smaller than the minimum relative approximation errors by the testing NMF algorithms.
In addition, we report the computational time (seconds calculated by MATLAB) in the tables. We see that the
computational time required by the proposed TAP method is less than that required by AP method.

\subsection{The Second Experiment}\label{sec4.2}

\subsubsection{Face Data}

In this subsection, we consider two frequently-used face data sets, i.e., the ORL face
date set\footnote{http://www.uk.research.att.com/facedatabase.html} and the Yale B face data set \footnote{http://vision.ucsd.edu/$\sim$leekc/ExtYaleDatabase/ExtYaleB.html}.
The ORL face data set contains images from 40 individuals, each providing 10 different images with the size $112\times92$ .
In the Yale B face data set, we take a subset which consists of 38 people and 64 facial images with different illuminations for each individual.
Each testing image is reshaped to a vector, and all the image vectors are combined together to form a nonnegative matrix.
Here we perform NMF algorithms and TAP algorithm to obtain low rank approximations with a predefined rank $r$.
There are several NMF algorithms to be compared, namely
multiplicative updates (MU) \cite{lee1999learning,lee2001algorithms}, accelerated MU (A-MU) \cite{gillis2012accelerated},
hierarchical alternating least squares (HALS) algorithm \cite{cichocki2007hierarchical},
accelerated HALS (A-HALS) \cite{gillis2012accelerated}, projected gradient (PG) Method \cite{lin2007projected},
and accelerated PG (A-PG)\cite{lin2007projected}.

Firstly, we compare the low rank approximation results by different methods with respect to different predefined ranks $r$. We report the relative approximation errors in Table \ref{Tab-Face-Rec}. For ORL data set, we set $r$ to be 10 and 40 because face data
contains 40 individuals and each individual has 10 different images. Similarly, $r$ is set to be 38 and 64 for Yale B data set.
In the numerical results, we compare the relative approximation error:
$\| {\bf X}_c- {\bf A} \|_F/\| {\bf A}\|_F$.
For the TAP and AP methods
the nonnegative low rank approximation is directly computed, while for the NMF methods, we multiply the factor matrices.
We can see from the table that the relative approximation errors by TAP and AP
methods are lower than those by NMF methods.

\begin{figure}[!t]
\centering
\includegraphics[width=0.75\linewidth]{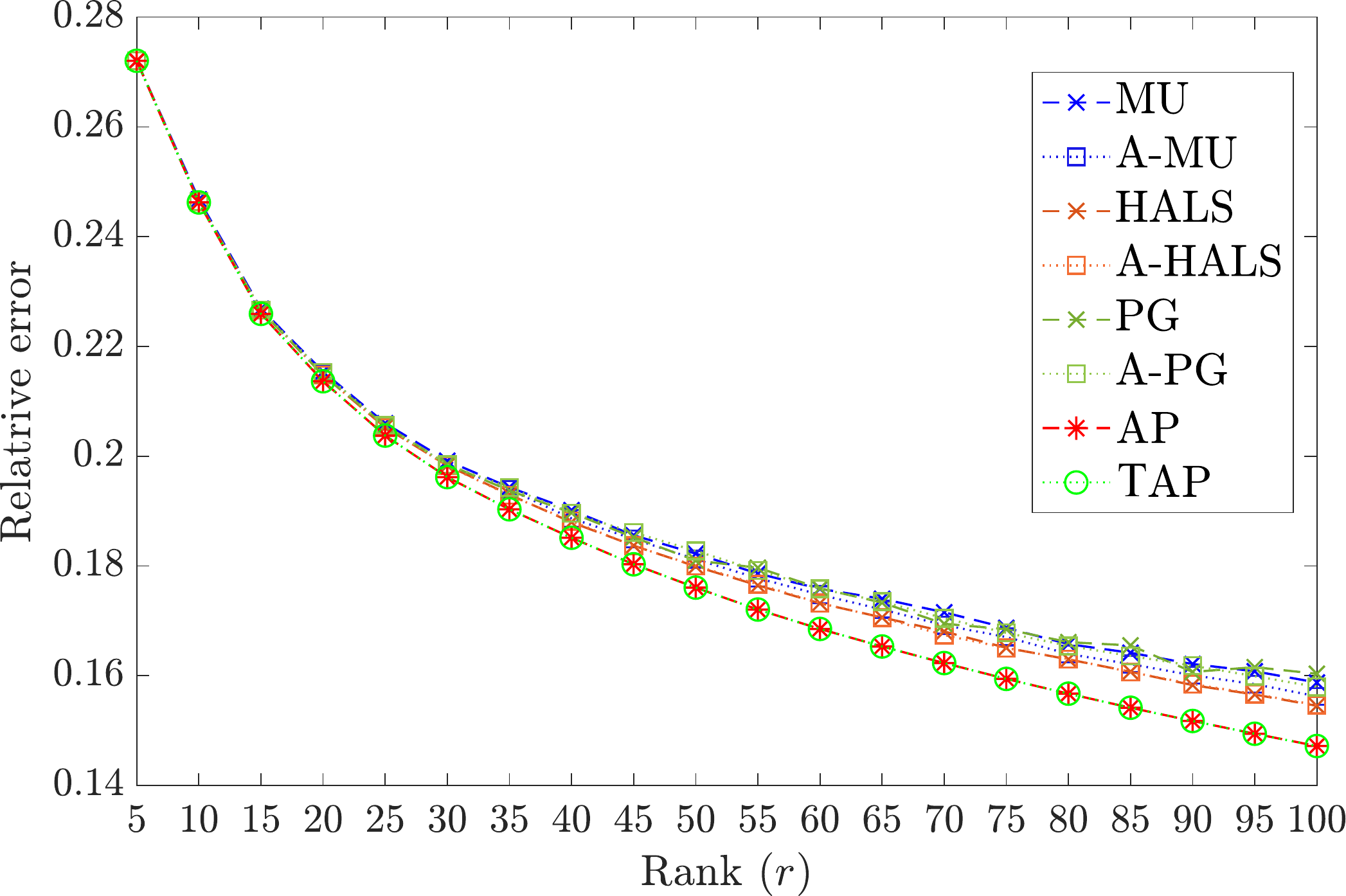}\\\includegraphics[width=0.75\linewidth]{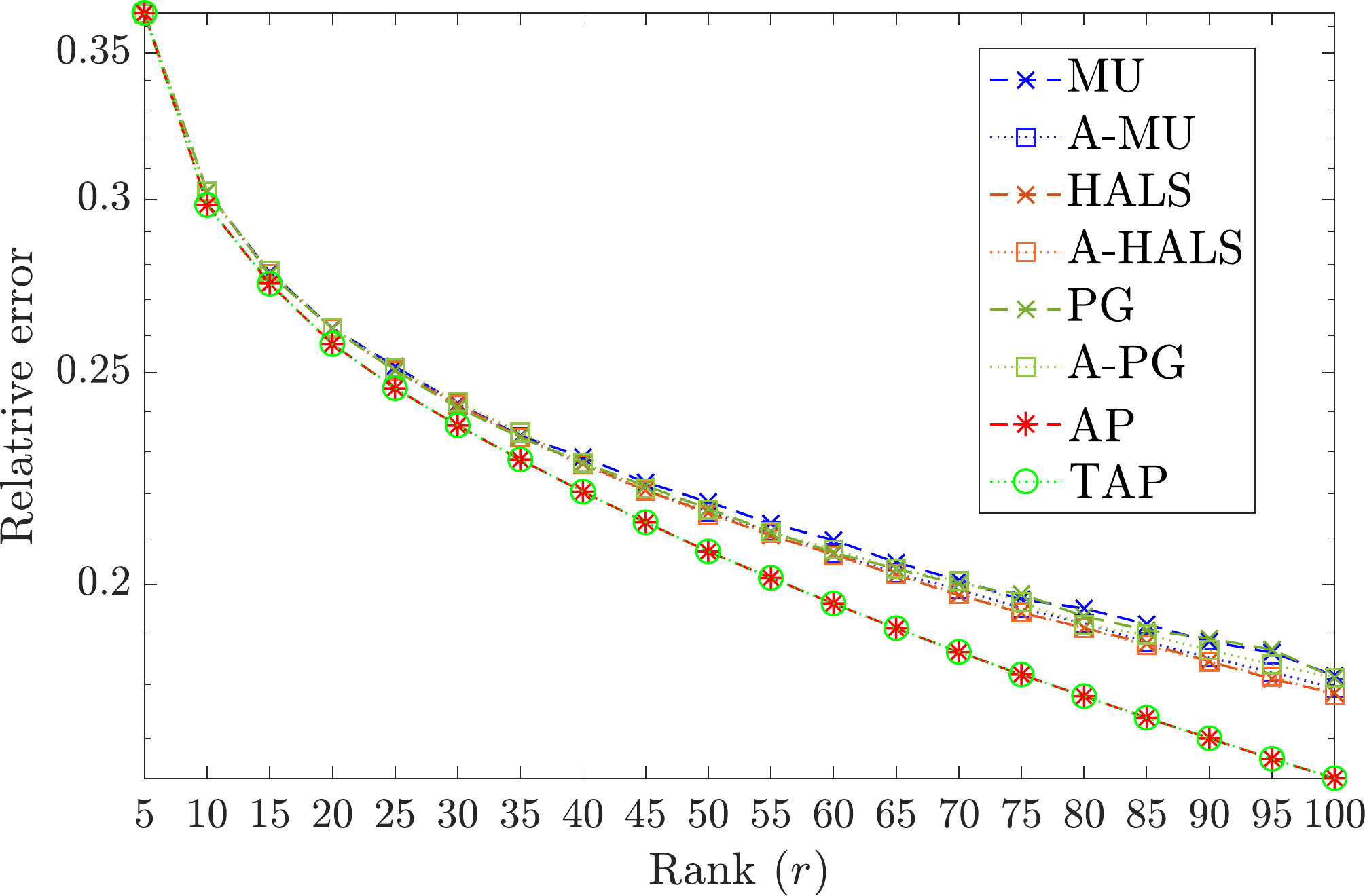}
\caption{Relative approximation errors on the ORL data set (Top) and the Yale B data set (Bottom), with respect to the different ranks $r$. }
\label{fig-face-rec-error}
\end{figure}

The relative approximation errors on these two face data sets with respect to different ranks $r$ are plotted in Figure \ref{fig-face-rec-error}. We can see that as $r$ increases, the gap of relative approximation errors between TAP (or AP) method and NMF methods becomes larger.
 The total computational time required by the proposed TAP method (2.84 seconds) is less than
that (17.44 seconds) required by the AP method. The proposed TAP method
is more efficient than the AP method.

\begin{table}[!t]
\renewcommand\arraystretch{1.0}\setlength{\tabcolsep}{2pt}
\centering\scriptsize
\caption{The relative approximation error on the Yale-B data set and the ORL data set. The \textbf{best} values and the \underline{second best} values are respectively highlighted by bolder fonts and underlines.}
\begin{tabular}{l c c c c c c c c c c c c c}
\toprule
Dataset & Rank&  MU & A-MU & HALS &A-HALS & PG &A-PG & AP & TAP\\\midrule
 \multirow{2}{*}{Yale B}
& $r = 38$ & 0.186 & 0.182 &\underline{0.181} & \underline{0.181} & 0.187 & 0.184 & \bf0.166 & \bf0.166 \\
& $r = 64$ & 0.160 & 0.157 & \underline{0.152} & \underline{0.152} & 0.159 & 0.159 & \bf0.133 & \bf0.133 \\
\midrule
 \multirow{2}{*}{ORL}
& $r = 10$ & 0.206 & 0.206 & \underline{0.205} & \underline{0.205} & 0.206 & 0.206 & \bf0.204 & \bf0.204 \\
& $r = 40$ & 0.159 & 0.156 & \underline{0.155} & \underline{0.155} & 0.160 & 0.158 & \bf0.147 & \bf0.147 \\
\bottomrule
\end{tabular}%
\label{Tab-Face-Rec}
\end{table}

\begin{figure}[!t]
\centering
\includegraphics[width=0.75\linewidth]{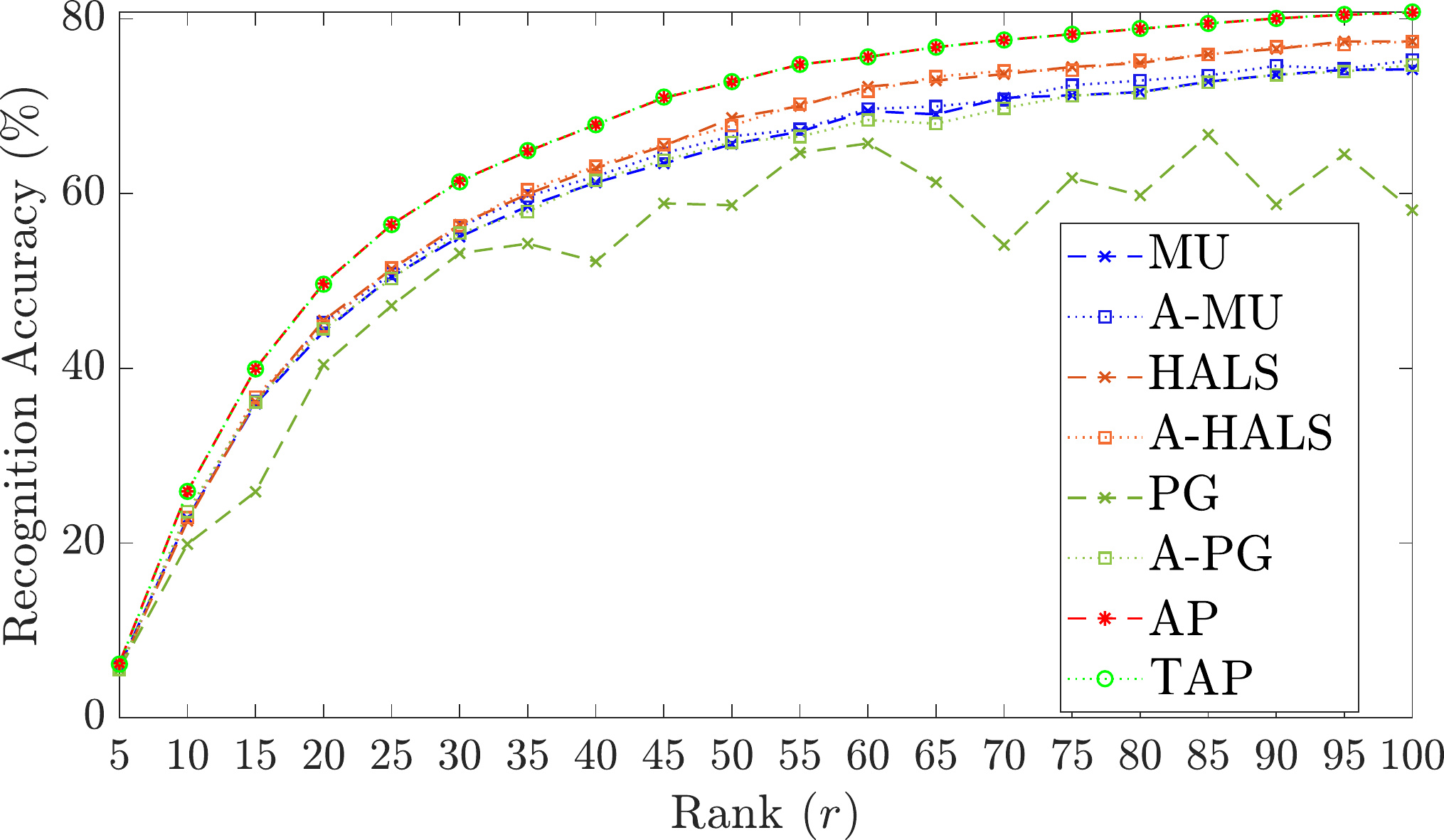}
\caption{The recognition accuracy (\%) on Yale-B dataset with respect to rank $r$.}
\label{fig-face}
\end{figure}
\begin{figure}[!t]
\centering
\begin{tabular}{cccccc}
\multicolumn{2}{c}{MU} & \multicolumn{2}{c}{HALS} & \multicolumn{2}{c}{PG}\\
\multicolumn{2}{c}{\includegraphics[width=0.28\linewidth]{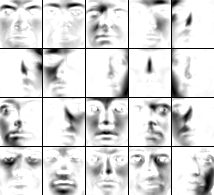}}&
\multicolumn{2}{c}{\includegraphics[width=0.28\linewidth]{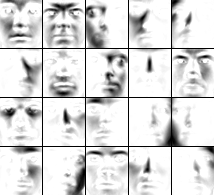}}&
\multicolumn{2}{c}{\includegraphics[width=0.28\linewidth]{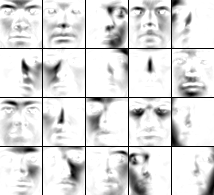}}\vspace{2mm}\\

\multicolumn{2}{c}{A-MU} &\multicolumn{2}{c}{A-HALS} &\multicolumn{2}{c}{A-PG}\\
\multicolumn{2}{c}{\includegraphics[width=0.28\linewidth]{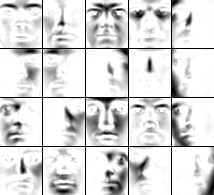}}&
\multicolumn{2}{c}{\includegraphics[width=0.28\linewidth]{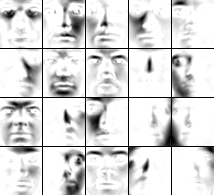}}&
\multicolumn{2}{c}{\includegraphics[width=0.28\linewidth]{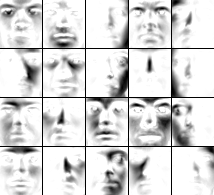}}\vspace{2mm}\\

&\multicolumn{2}{c}{\hspace{1cm}AP}& \multicolumn{3}{c}{TAP}\\
&\multicolumn{2}{c}{\hspace{1cm}\includegraphics[width=0.28\linewidth]{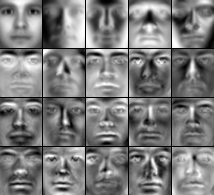}}&
\multicolumn{3}{c}{\includegraphics[width=0.28\linewidth]{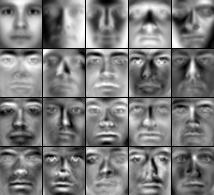}}\\
\end{tabular}
\vspace{-2mm}
\caption{The first 20 singular vectors of the results by TAP (or AP)
method and the columns of left factor matrices resulted by NMF methods when the rank $r = 20$. These vectors are reshaped to the size of facial images and their values are adaptively normalized.}
\label{fig-face2}
\end{figure}
Next, we test the face recognition performance with respect to TAP approximations and NMF approximations.
We use the $k$-fold cross-validation strategy. For each data set, the data is split into $k$ ($k=10$ for the ORL data set and $k = 64$ for the Yale B data set) groups and each group contains one facial image of each individual. For instance, the ORL data set is split into $k=10$ groups and each group contains 40 facial images. Then, we circularly take one group as a test data set and the remaining groups as a training data set until all the groups have been selected as the test data.
Given the original training data ${\bf A}_\text{train}$ with the size $m\times n$, where $n$ indicates the pixels of each face image and $m$ is the amount of training samples, we first perform NMF and TAP (or AP) algorithms to obtain non-negative low rank approximations ${\bf A}_\text{train}
\approx {\bf B}_\text{NMFtran} {\bf C}_\text{NMFtrain}$ and ${\bf A}_\text{train} \approx {\bf U}_\text{TAPtrain} {\bf \Sigma}_\text{TAPtrain} {\bf V}_\text{TAPtrain}$
respectively with rank $r$.
The new representations of ${\bf A}_\text{train}$ are given by
${\bf U}_\text{NMFtrain}^T {\bf A}_\text{train}$ and ${\bf U}_\text{TAPtrain}^T
{\bf A}_\text{train}$
respectively by the NMF methods and the TAP (or AP) method.
The nearest neighbor (NN) classifier is adopted by recognized the testing data based on the distance between
their representations and the projected training data.

\begin{table*}[!t]
\renewcommand\arraystretch{1.0}\setlength{\tabcolsep}{4pt}
\centering
\caption{The recognition accuracy on the Yale-B dataset and ORL dataset. The \textbf{best} values and the \underline{second best} values are respectively highlighted by bolder fonts and underlines.}
\begin{tabular}{l c c c c c c c c c c c c c}
\toprule
Dataset & Parameter&  MU & A-MU & HALS &A-HALS & PG &A-PG & AP & TAP \\\midrule
 \multirow{2}{*}{Yale B}
& $r = 38$ & 61.061\% & 61.143\% & 61.637\% & 62.253\% & 58.306\% & 60.074\% & \bf66.776\% & \underline{67.681}\% \\
& $r = 64$ & 69.942\% & 70.477\% & 72.821\% & 72.821\% & 65.502\% & 68.586\% & \underline{76.563}\% & \bf76.809\% \\
\midrule
 \multirow{2}{*}{ORL}
& $r = 10$ & 95.750\% & 96.250\% & 96.250\% & 96.250\% & \underline{96.500}\% & \underline{96.500}\% & {\bf96.750}\% & {\bf96.750}\% \\
& $r = 40$ & \underline{98.250}\% & 98.000\% & \underline{98.250}\% & {\bf98.500}\% & 79.250\% & \underline{98.250}\% & {\bf98.500}\% & {\bf98.500}\% \\

\bottomrule
\end{tabular}%
\label{Tab-facespec}
\end{table*}

\begin{table*}[!t]
\renewcommand\arraystretch{1.2}\setlength{\tabcolsep}{4pt}
\centering\scriptsize
\caption{The accuracy and NMI values of the document clustering results on the TDT2 data set.}
\begin{tabular}{c c c c c c c c c c c c c c}
\toprule
Metric&MU& A-MU& HALS&A-HALS&PG&A-PG& AP & TAP \\\midrule
Accuracy& 52.800\% & 50.724\% & 54.322\% & 53.108\% & 54.205\% & 51.661\% & \underline{61.294}\% & \bf 61.326\% \\
NMI & 0.674   & 0.651   & 0.663   & 0.643   & \underline{0.681}   & 0.661   & \bf{0.728}   & \bf{0.728}   \\
\bottomrule
\end{tabular}%
\label{Tab-face}
\end{table*}

The face recognition results are exhibited in Table \ref{Tab-facespec}. From this table, we can see that the accuracies
based on TAP approximations are higher than those based on NMF approximations. To further investigate how the rank $r$ affects the recognition results, we plot the recognition accuracy on Yale B data set with respect to $r$ in Figure \ref{fig-face}. It can be found that the recognition accuracy based on TAP and AP approximations is always better than those based on NMF approximations.
Meanwhile, to see the features learned by different methods, we exhibit the column vectors of ${\bf B}_\text{NMFtrain}$ and singular vectors of ${\bf U}_\text{TAPtrain}$
in Figure \ref{fig-face2}.
These vectors are reshaped to the same size as facial images and their values are
normalized to [0,255] for the display purpose. We see that the nonnegative low rank matrix approximation methods do not give
the part-based representations, but provides different important facial representations in the recognition.

\subsubsection{Document Data}

In this subsection, we use the NIST Topic Detection and Tracking (TDT2) corpus as the document data.
The TDT2 corpus consists of data collected during the first half of 1998 and taken from 6 sources, including 2 newswires (APW,
NYT), 2 radio programs (VOA, PRI) and 2 television programs (CNN, ABC).
It consists of 11201 on-topic documents which are classified into 96 semantic categories.
In this experiment, the documents appearing in two or more categories were removed, and only the largest 30 categories were kept, thus
leaving us with 9394 documents in total. Then, each document is represented by the weighted term-frequency vector \cite{xu2003document}, and all the documents are gathered as a matrix
${\bf A}_\text{doc}$ of size $9394\times36771$.
By using the procedure given in \cite{xu2003document}, we compute the projected results ${\bf U}_\text{TAP}^T {\bf A}_\text{TAP} =
{\bf \Sigma}_\text{TAP} {\bf V}_\text{TAP}^T$, and then use
$k$-means clustering method and Kuhn-Munkres algorithm to find the best mapping which maps each cluster label to the equivalent label from
the document corpus. For NMF methods, we scale each column of ${\bf B}_\text{NMF}$ such that their $\ell_2$ norms are equal to 1, and the corresponding
scaled ${\bf C}_\text{NMF}$ is used for clustering and label assignment.
To quantitatively evaluate the clustering performance of each method, we selected two metrics, i.e., the accuracy and the normalized mutual information (NMI) (we refer to \cite{cai2005document} for detailed discussion). According to Table (\ref{Tab-face}), it is clear
that nonnegative low rank matrix approximation can provide more effective latent features
(${\bf U}_\text{TAP}^T {\bf A}_\text{TAP} =
{\bf \Sigma}_\text{TAP} {\bf V}_\text{TAP}^T$) for document clustering task.
Note that the computational time required by the proposed TAP method (309.22 seconds) is less than
that (3417.33 seconds) required by the AP method. Again the results demonstrate
that the proposed TAP method is more efficient than the AP method.

\subsection{Separable Nonnegative Matrices}

In this subsection, we compare the performance of the nonnegative low rank matrix approximation method and separable NMF algorithms.
Here we generate two kinds of synthetic separable nonnegative matrices.
\begin{itemize}
  \item (Separable) The first case ${\bf A} = {\bf B} {\bf C}+ {\bf N}$ is
generated the same as \cite{gillis2013fast}, in which ${\bf B}
\in \mathbb{R}^{200\times 20}$ is {\it uniform distributed} and ${\bf C} =
[{\bf I}_{20}, {\bf H}']\in\mathbb{R}^{20\times210}$ with ${\bf H}'$ containing all possible combinations of two non-zero entries equal to 0.5 at different positions.
The columns of ${\bf B} {\bf H}'$ are all the middle points of the columns of ${\bf B}$. Meanwhile, the $i$-th column of ${\bf N}$, denoted as $n_i$, obeys $n_i = \sigma(m_i-\bar{w})$ for $21\leq i\leq 210$, where $\sigma>0$ is the noise level,
$m_i$ is the $i$-th column of ${\bf B}$, and $\bar{w}$ denotes the average of columns of
${\bf B}$.
This means that we move the columns of ${\bf A}$ toward the outside of the convex hull of the columns of ${\bf B}$.
\item (Generalized separable) The second case is generated almost the same as the first case
but simultaneously considering the separability of rows, known as generalized separable NMF \cite{pan2019generalized}. For this case,
the size of ${\bf A}$ is set as $78\times55$ with column-rank $10$ and row-rank $12$, being the same as \cite{pan2019generalized}.
\end{itemize}

\begin{figure}[!t]
\centering
\includegraphics[width=0.75\linewidth]{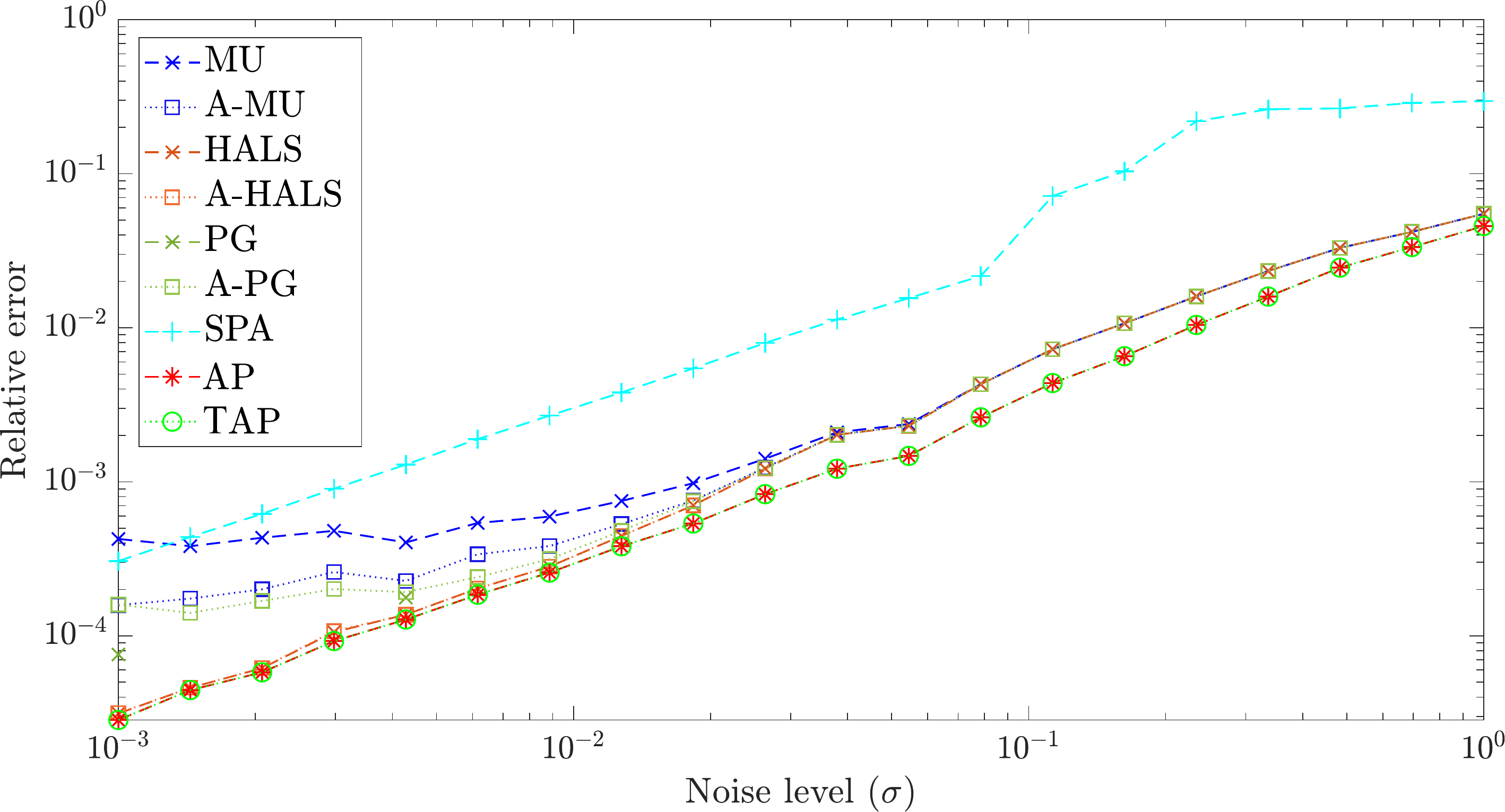}
\caption{Average relative approximation error on separable matrices (Case 1), with respect to the different values of $\sigma$. }
\label{fig-snmf-rec-error}
\end{figure}

\begin{figure}[!t]
\centering
\includegraphics[width=0.75\linewidth]{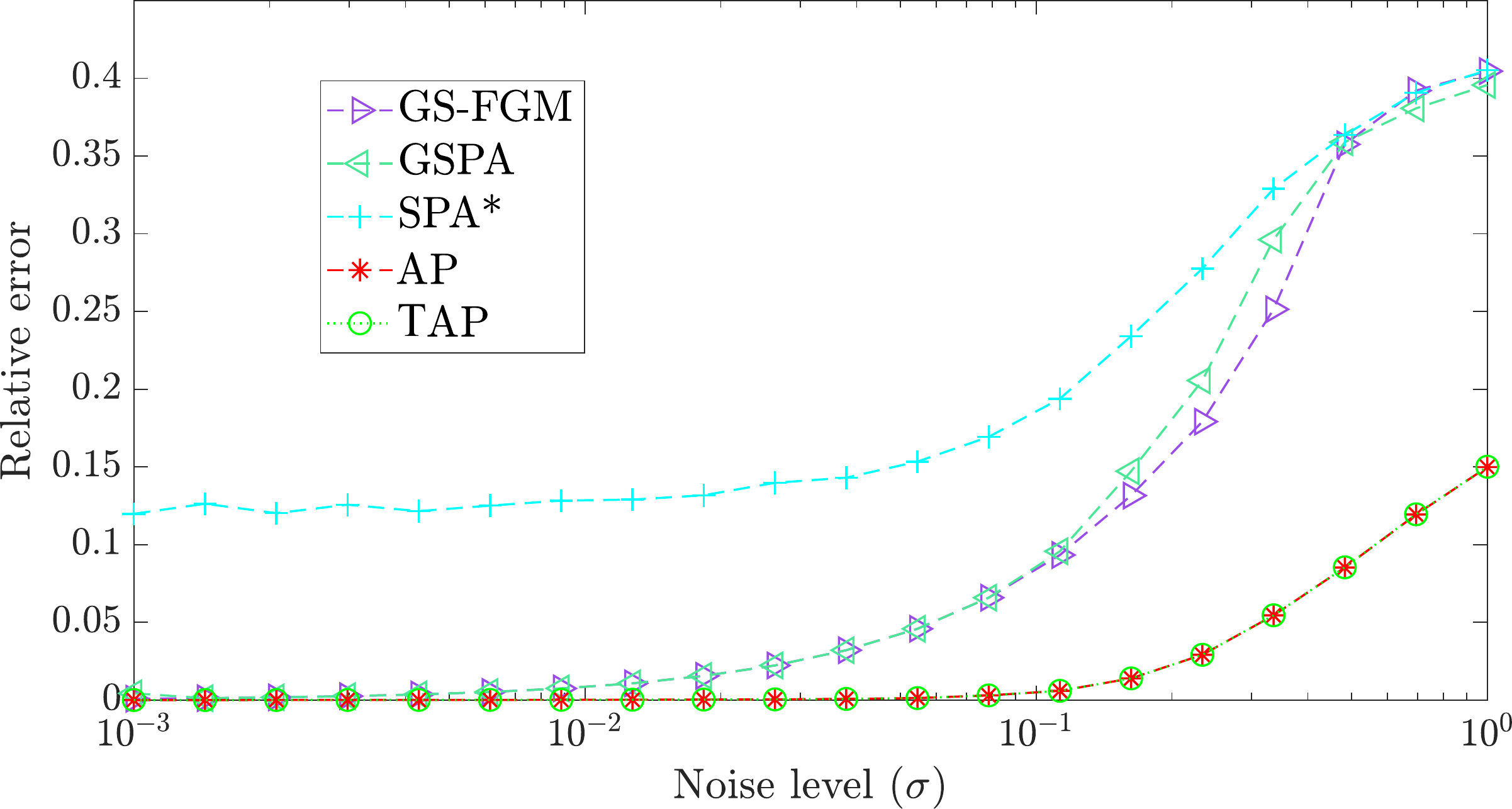}
\caption{Average relative approximation error on generalized separable matrices (Case 2), with respect to the different values of $\sigma$. }
\label{fig-gsnmf-rec-error}
\end{figure}

Firstly, we test the approximation ability of TAP and AP methods, NMF methods, and the successive projection algorithm (SPA) \cite{gillis2013fast,araujo2001successive} for separable NMF for synthetic separable data.
For the generalized separable case, we compare the TAP (or AP) method with SPA, the generalized SPA (GSPA) \cite{pan2019generalized}, and the generalized separable NMF with a fast gradient method (GS-FGM) \cite{pan2019generalized}.
Note that when we apply SPA on the generalized separable matrix, we run it firstly to identify the important columns and with the transpose of the input to identify the important rows. This variant is referred to SPA*.
The noise level $\sigma$ is logarithmic spaced in the interval $[10^{-3}, 1]$. For each noise level, we independently generate 25 matrices for both separable and generalized separable cases, respectively. We report the averaged approximation error in Figures \ref{fig-snmf-rec-error} and \ref{fig-gsnmf-rec-error}. It can be found that TAP and AP methods can achieve the lowest average errors in the testing examples.

\begin{figure}[!t]
\centering
\includegraphics[width=0.45\linewidth]{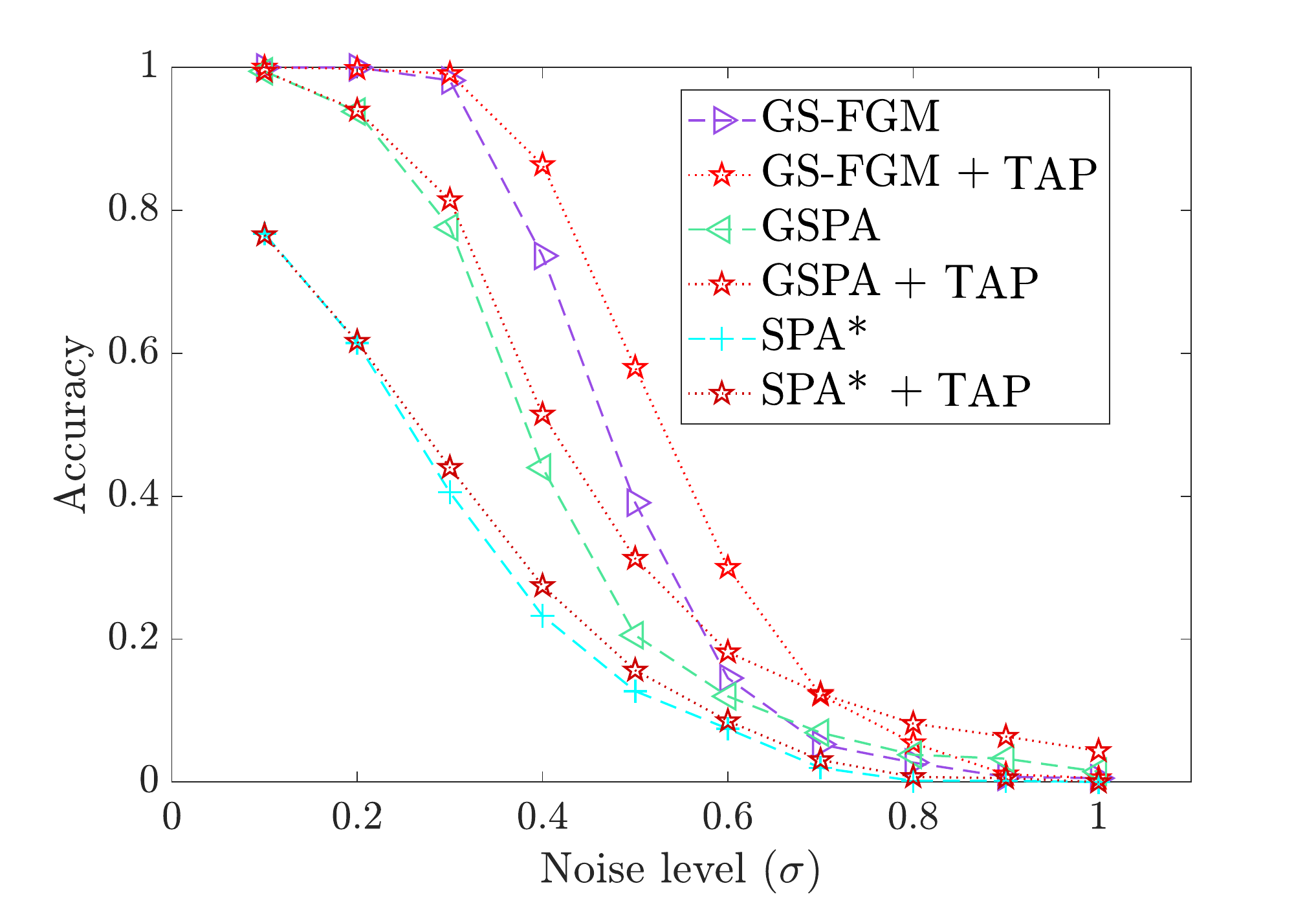}
\includegraphics[width=0.45\linewidth]{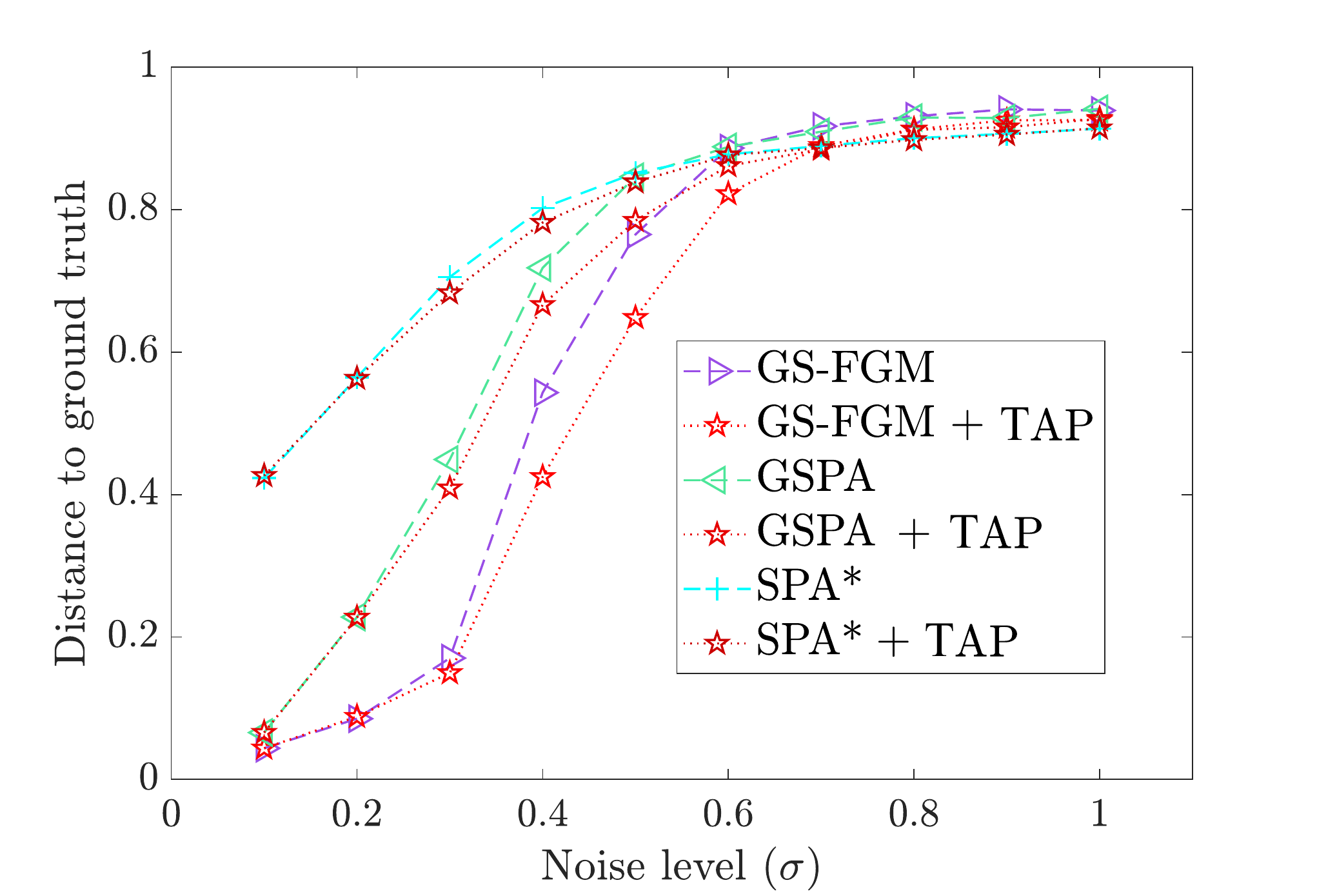}
\caption{Average accuracy (left) and distance to ground truth (right) for the different algorithms on generalized separable matrices (Case 2), with respect to the different $\sigma$s. }
\label{fig-gsnmf-ac-dgt}
\end{figure}

The approximation errors of TAP and AP methods are much lower than separable and generalized separable NMF methods when the noise level is high.
Note that the average
computational time required by the proposed TAP method (0.0064 seconds) is less than
that (0.0165 seconds) required by the AP method.
\begin{figure*}[!t]
\renewcommand\arraystretch{1.0}\setlength{\tabcolsep}{3pt}
\centering\scriptsize
\begin{tabular}{c c c c c c c}
&3 clusters	&3 clusters	&3 clusters	&5 clusters&	4 clusters	&3 clusters\\
\rotatebox[origin=l]{90}{\hspace{2.5mm}Original data}&
\includegraphics[width=0.14\linewidth]{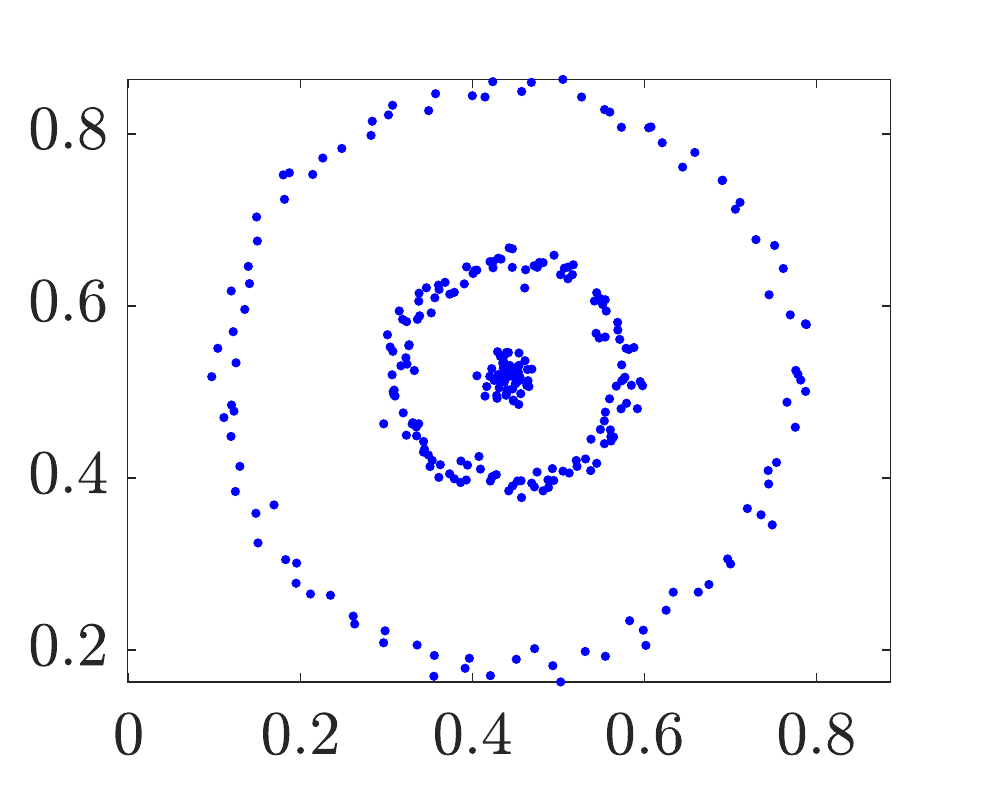}&
\includegraphics[width=0.14\linewidth]{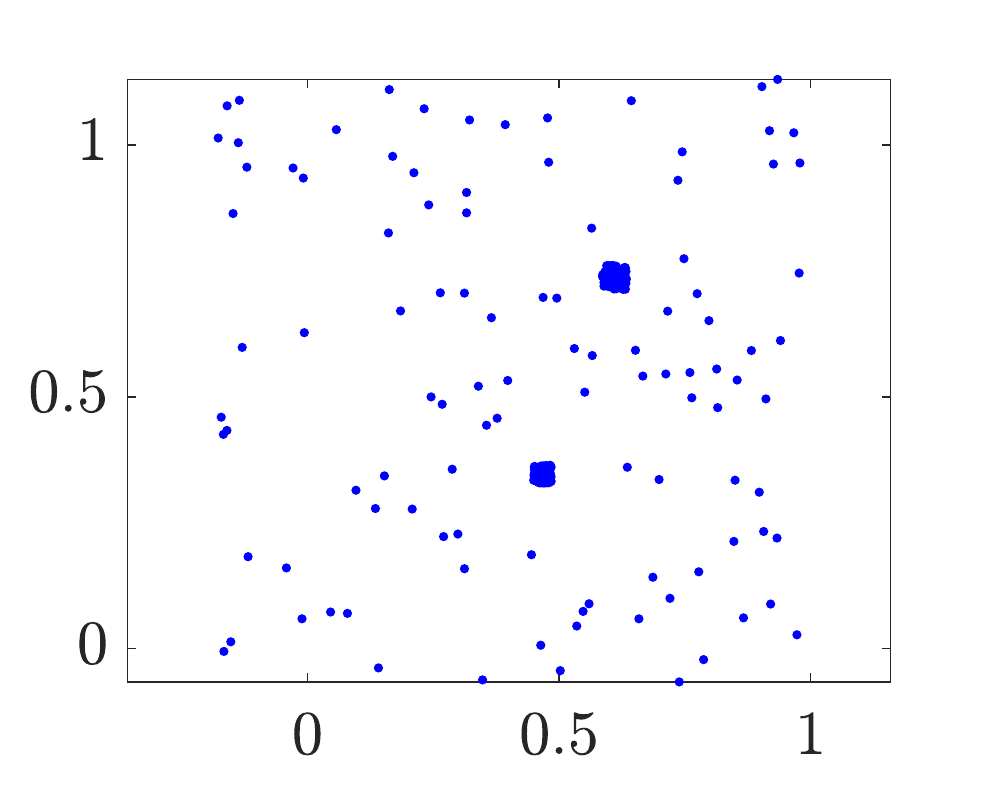}&
\includegraphics[width=0.14\linewidth]{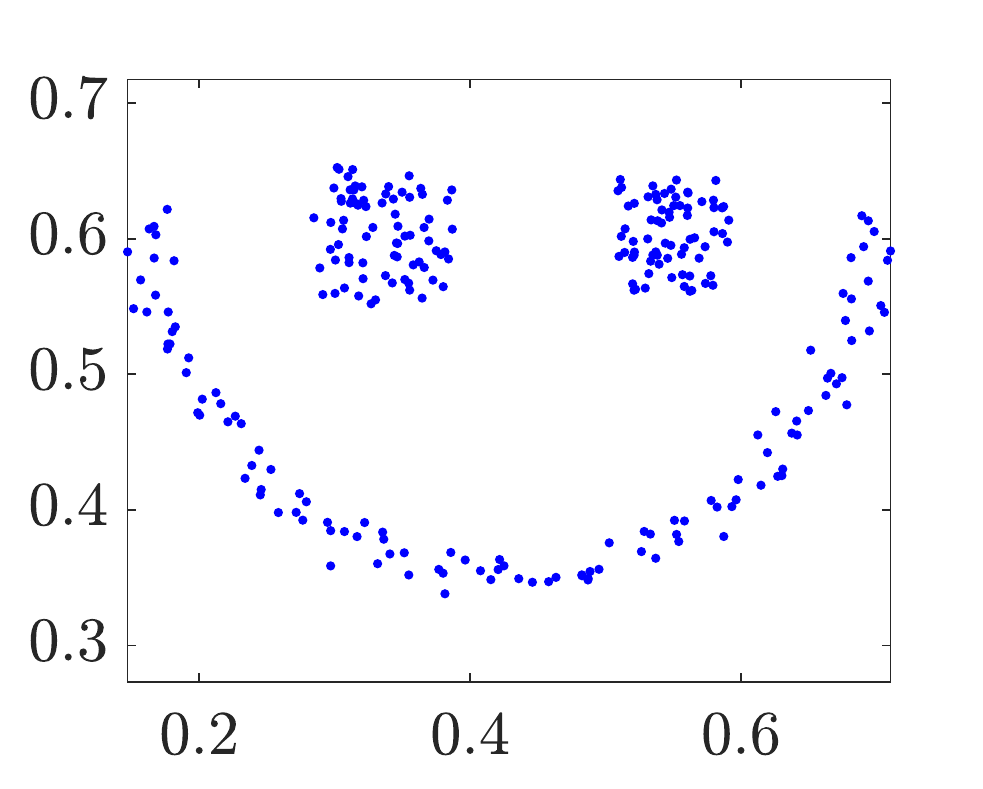}&
\includegraphics[width=0.14\linewidth]{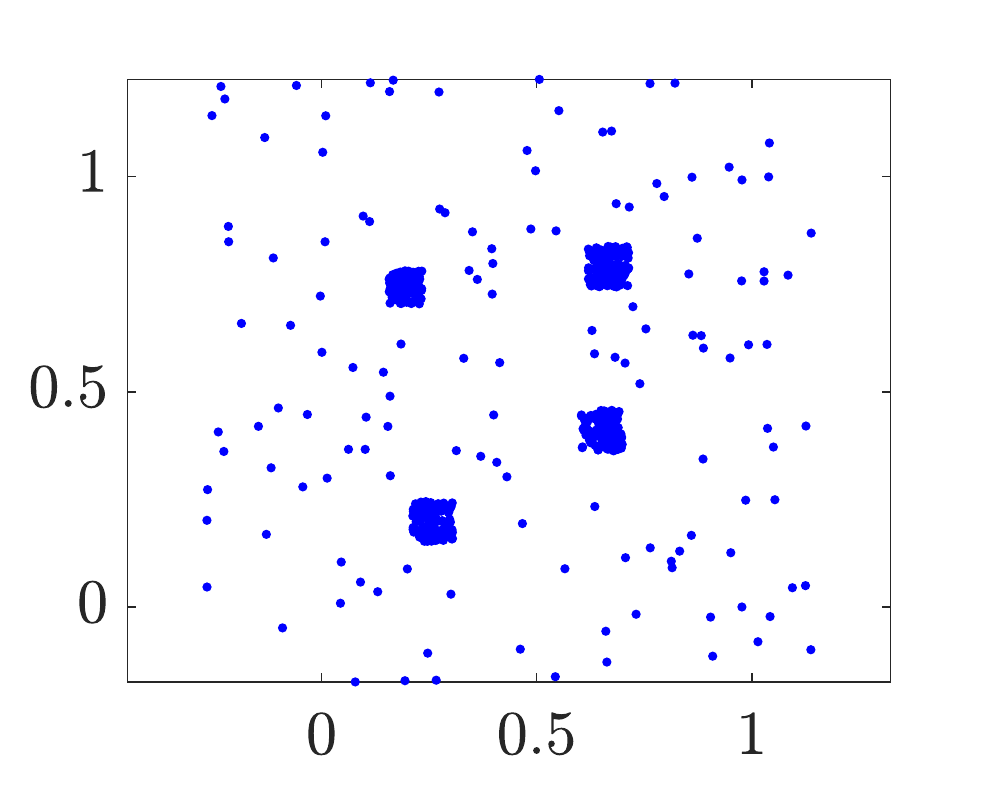}&
\includegraphics[width=0.14\linewidth]{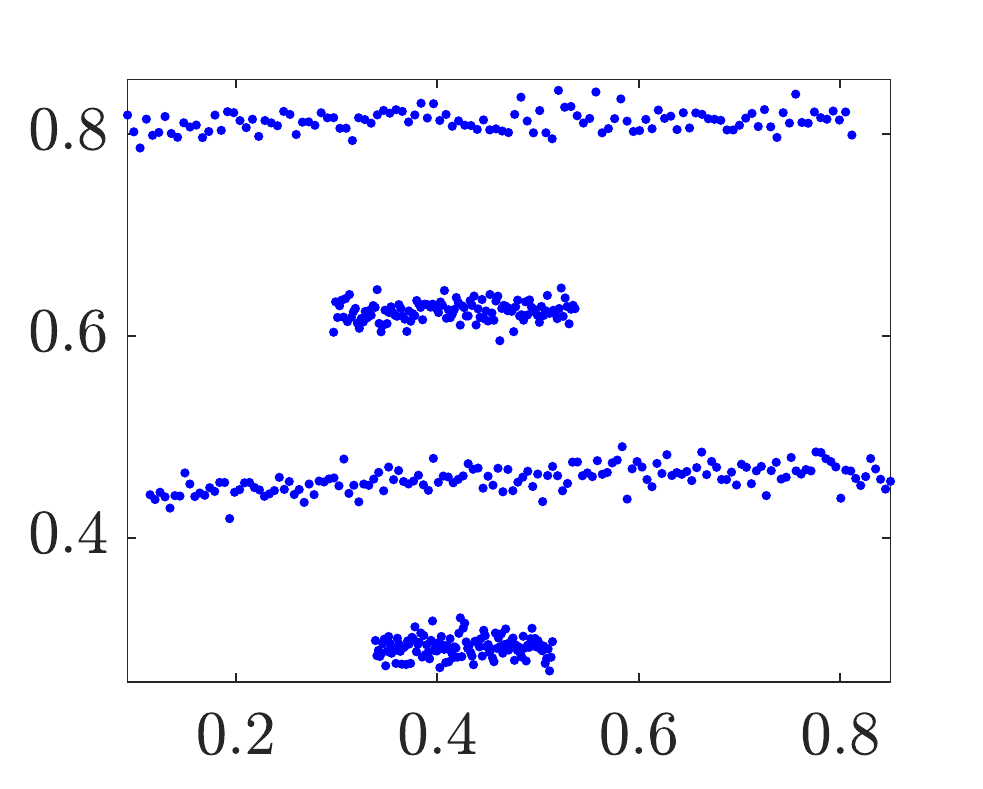}&
\includegraphics[width=0.14\linewidth]{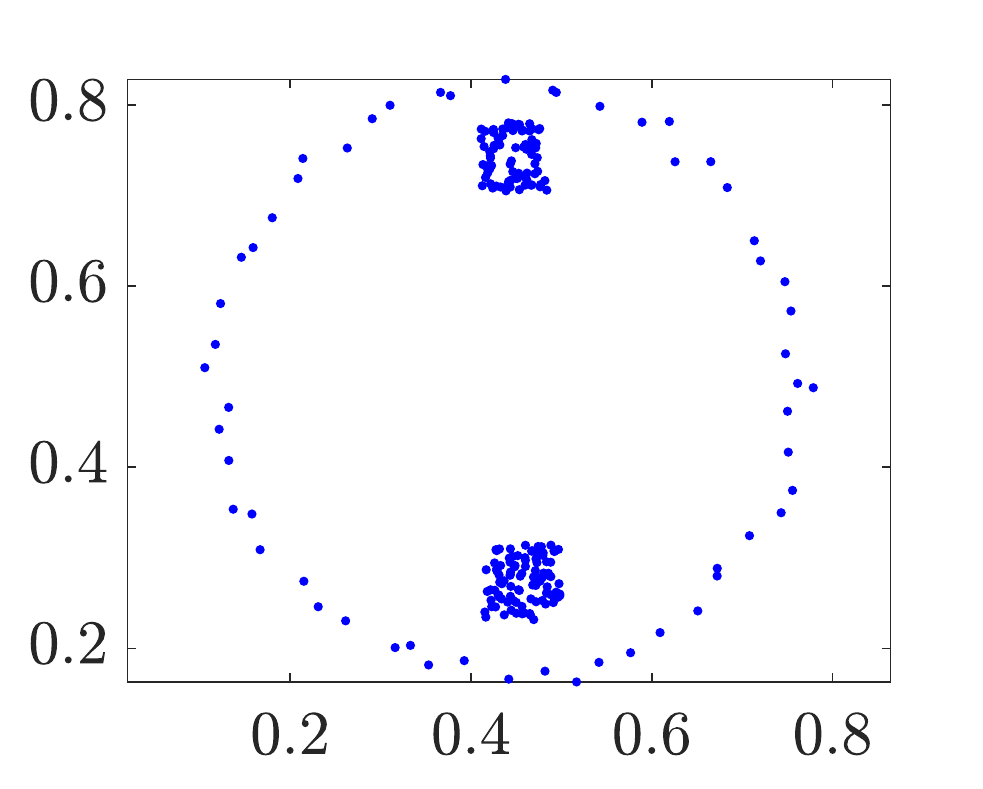}\\

\rotatebox[origin=l]{90}{\hspace{2.5mm}CD-symNMF}&
\includegraphics[width=0.14\linewidth]{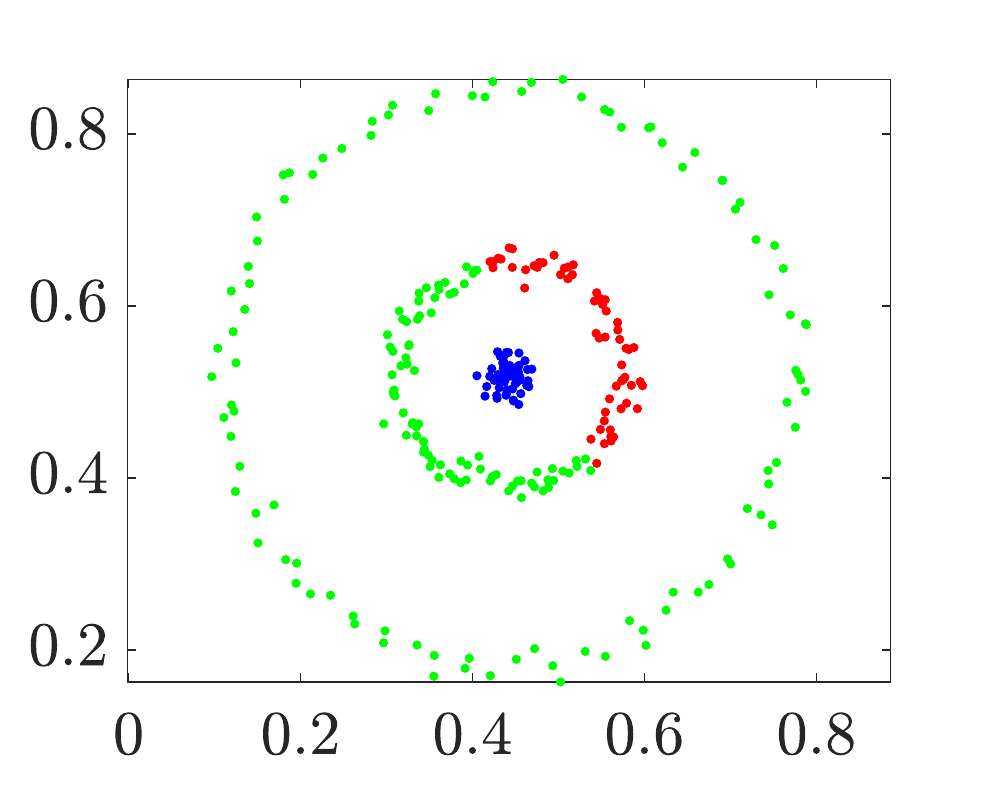}&
\includegraphics[width=0.14\linewidth]{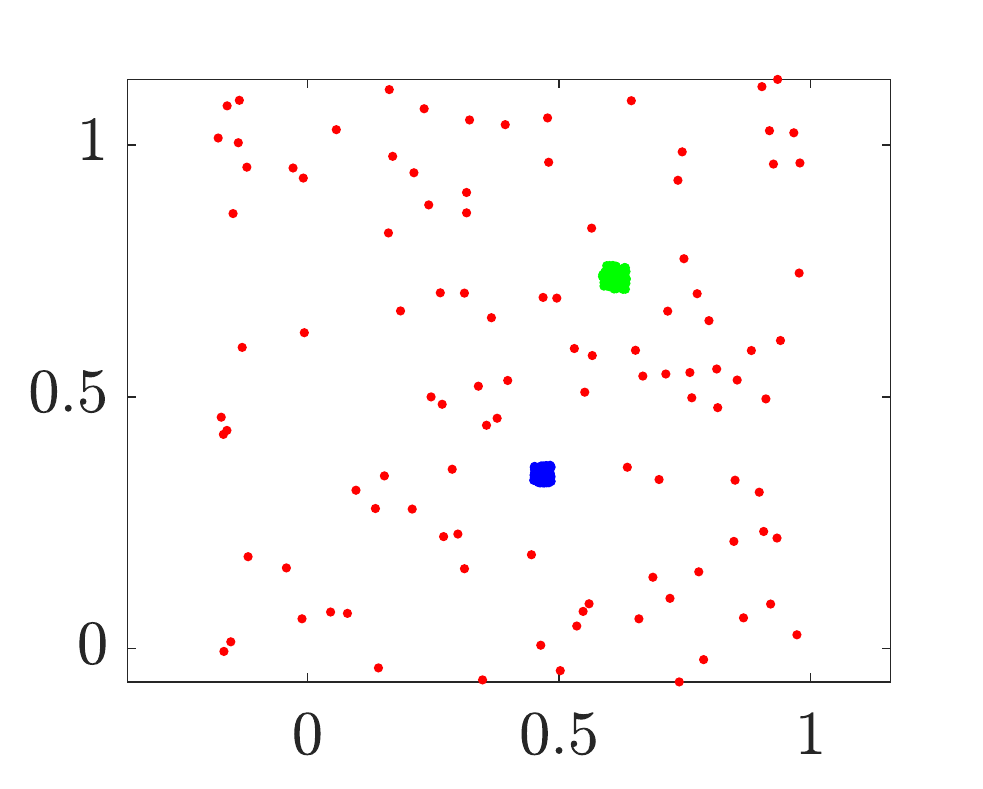}&
\includegraphics[width=0.14\linewidth]{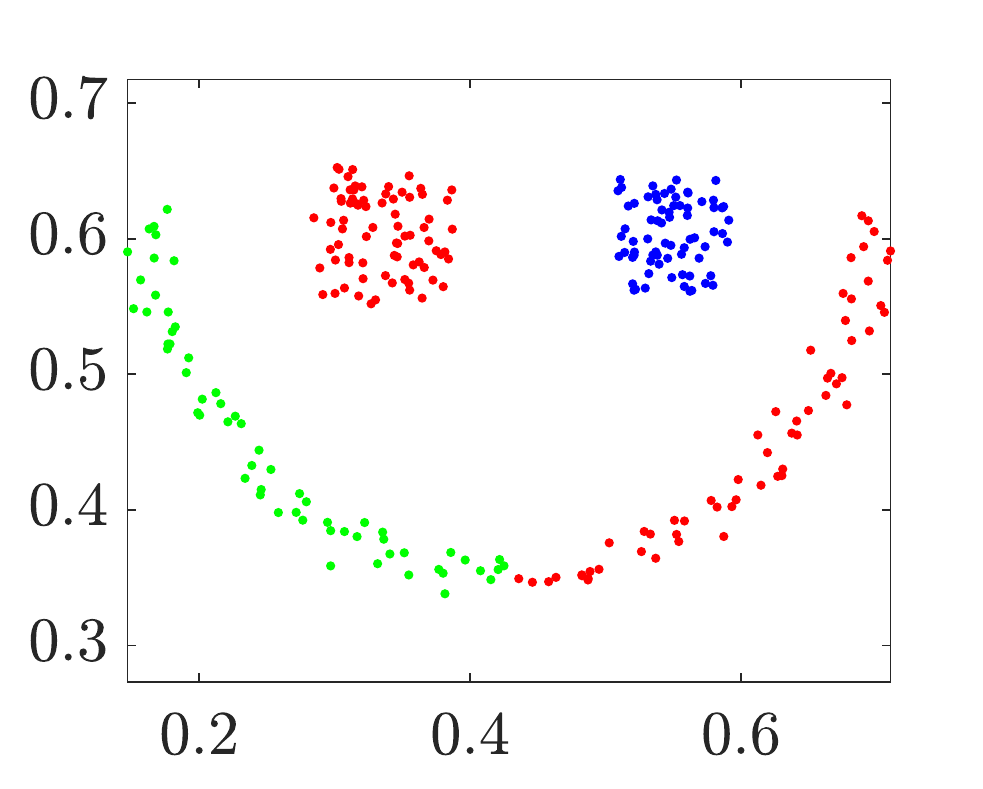}&
\includegraphics[width=0.14\linewidth]{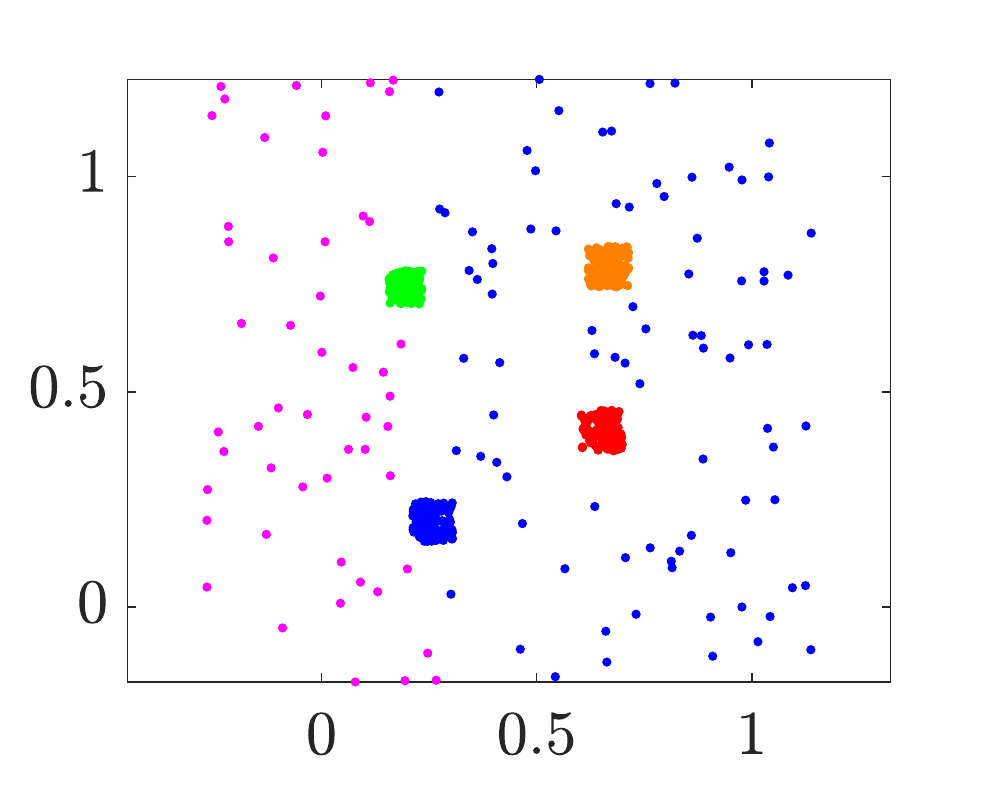}&
\includegraphics[width=0.14\linewidth]{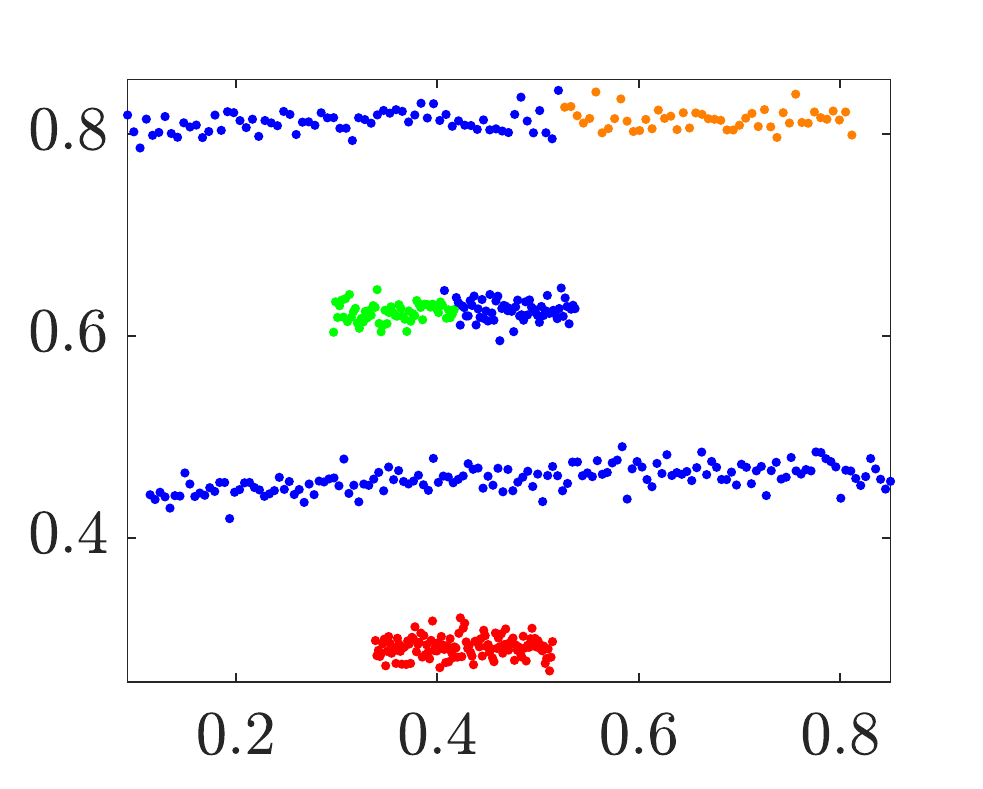}&
\includegraphics[width=0.14\linewidth]{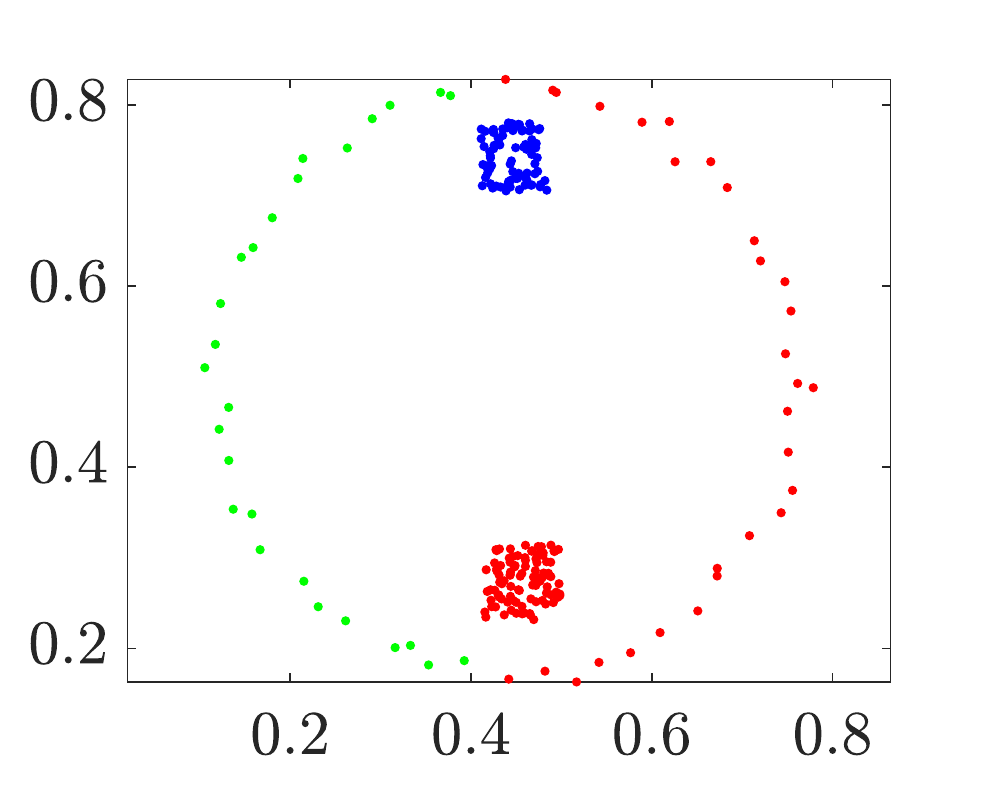}\\

\rotatebox[origin=l]{90}{Newton-symNMF}&
\includegraphics[width=0.14\linewidth]{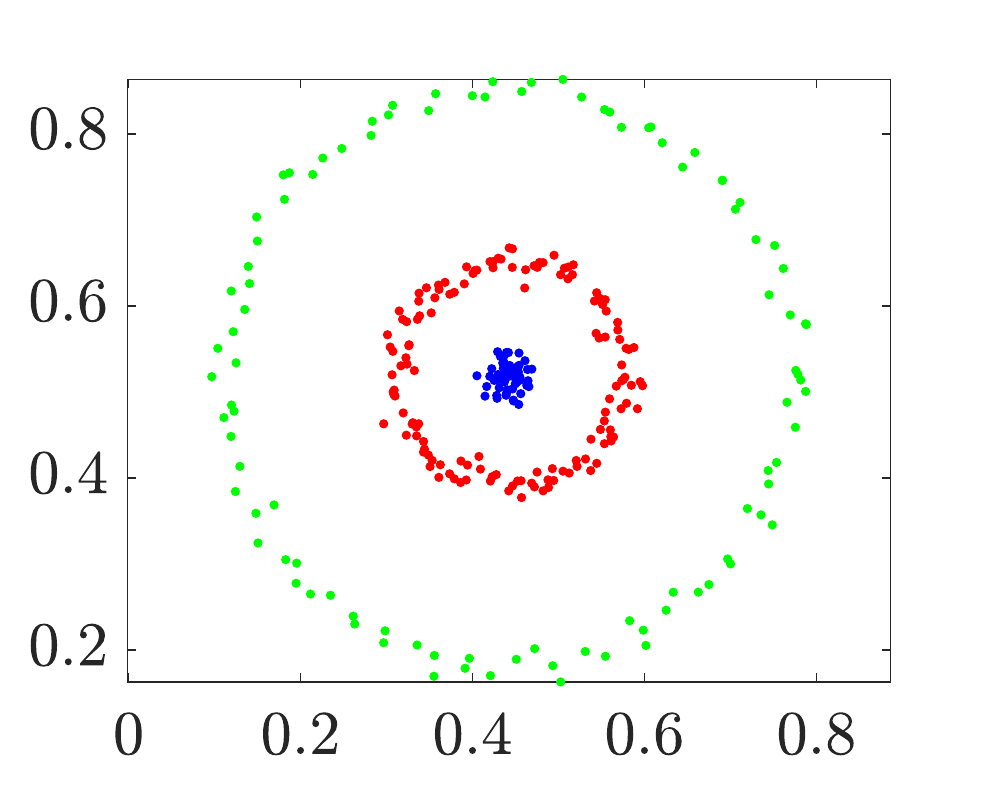}&
\includegraphics[width=0.14\linewidth]{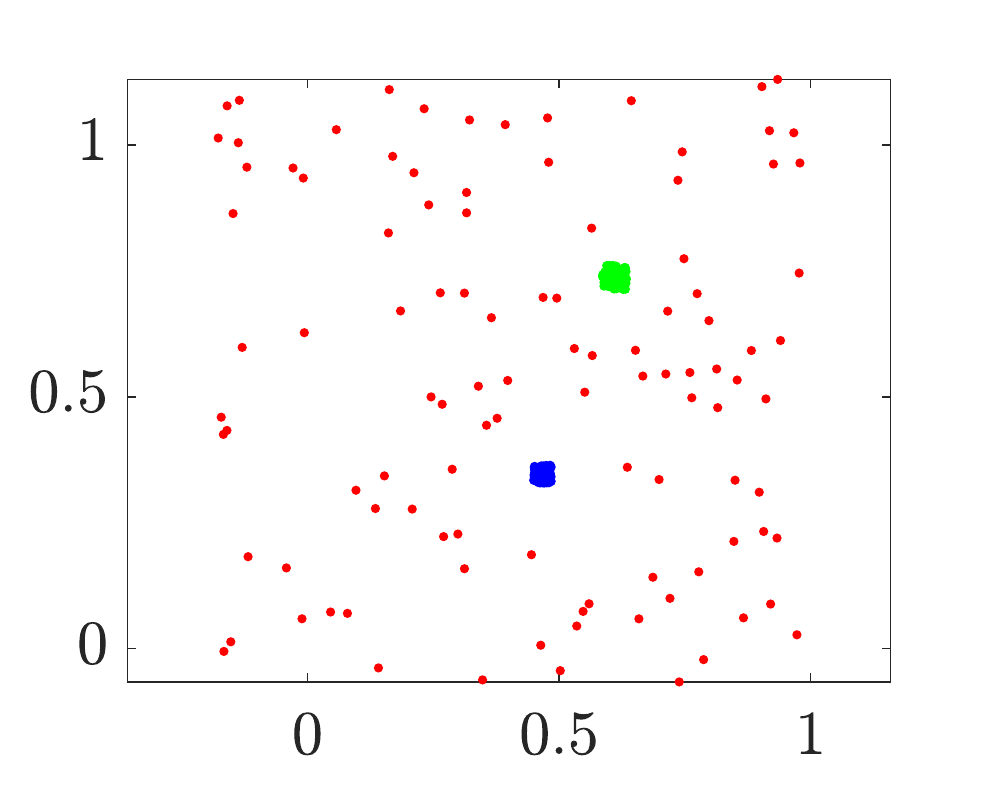}&
\includegraphics[width=0.14\linewidth]{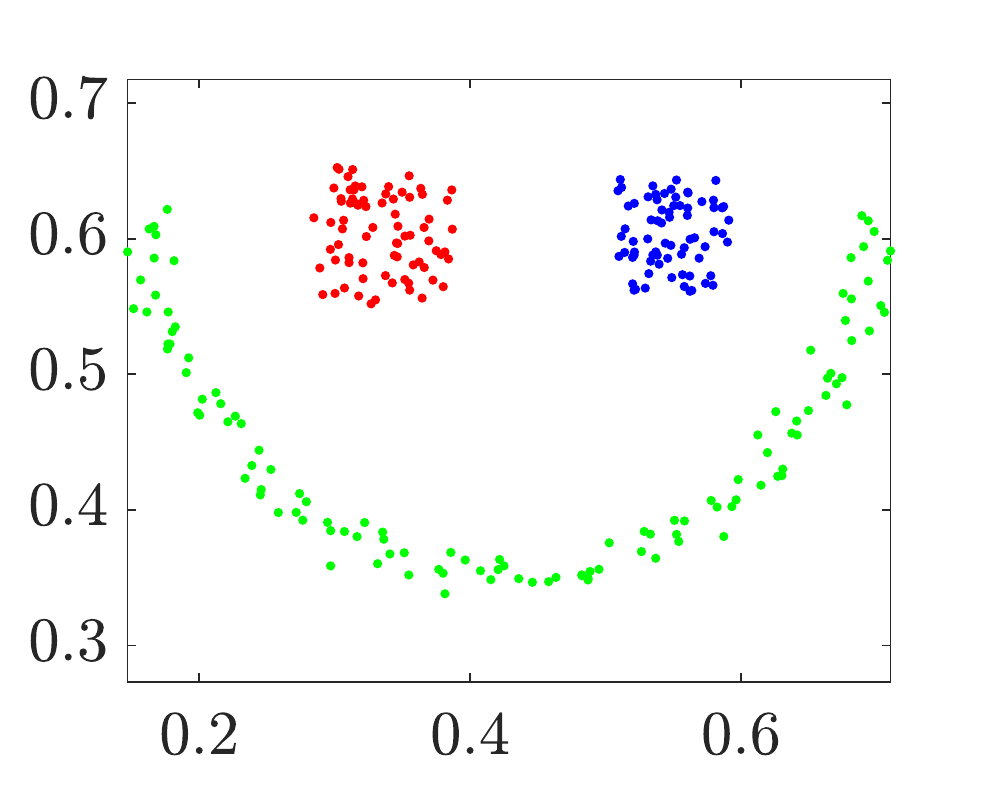}&
\includegraphics[width=0.14\linewidth]{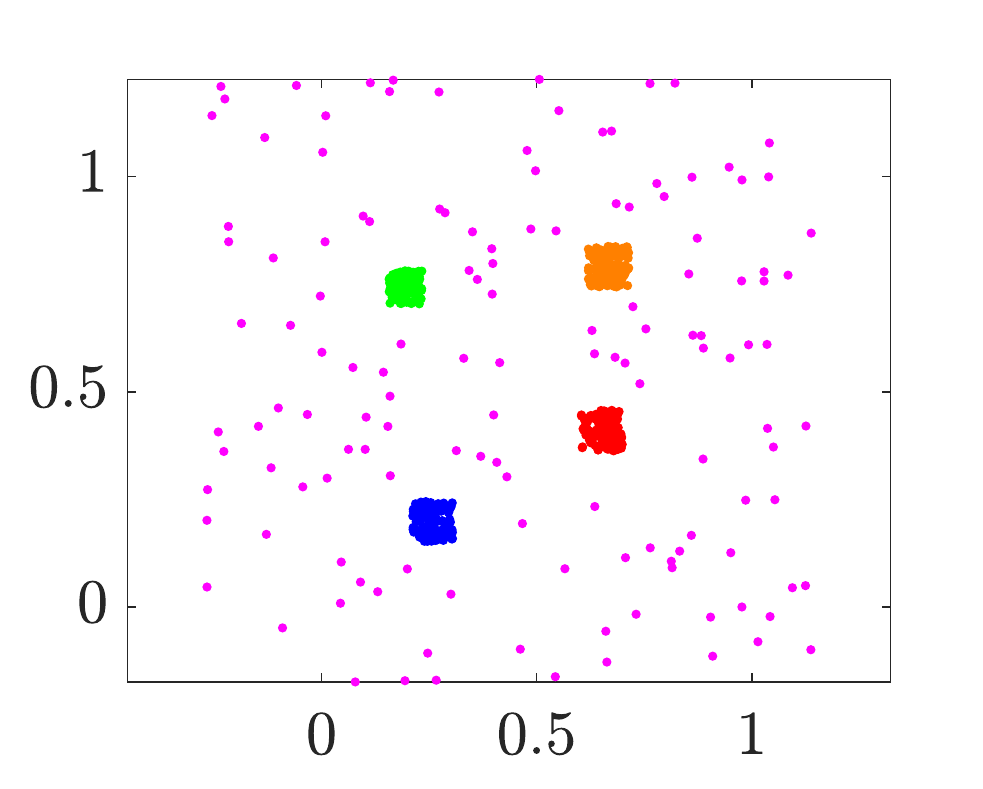}&
\includegraphics[width=0.14\linewidth]{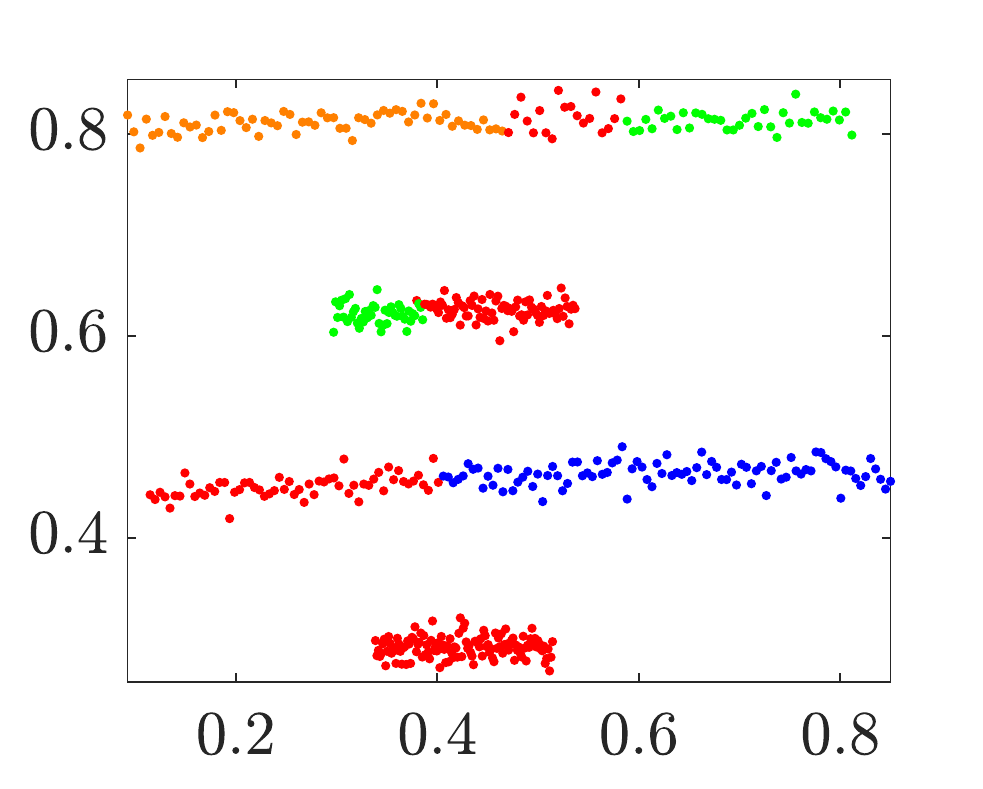}&
\includegraphics[width=0.14\linewidth]{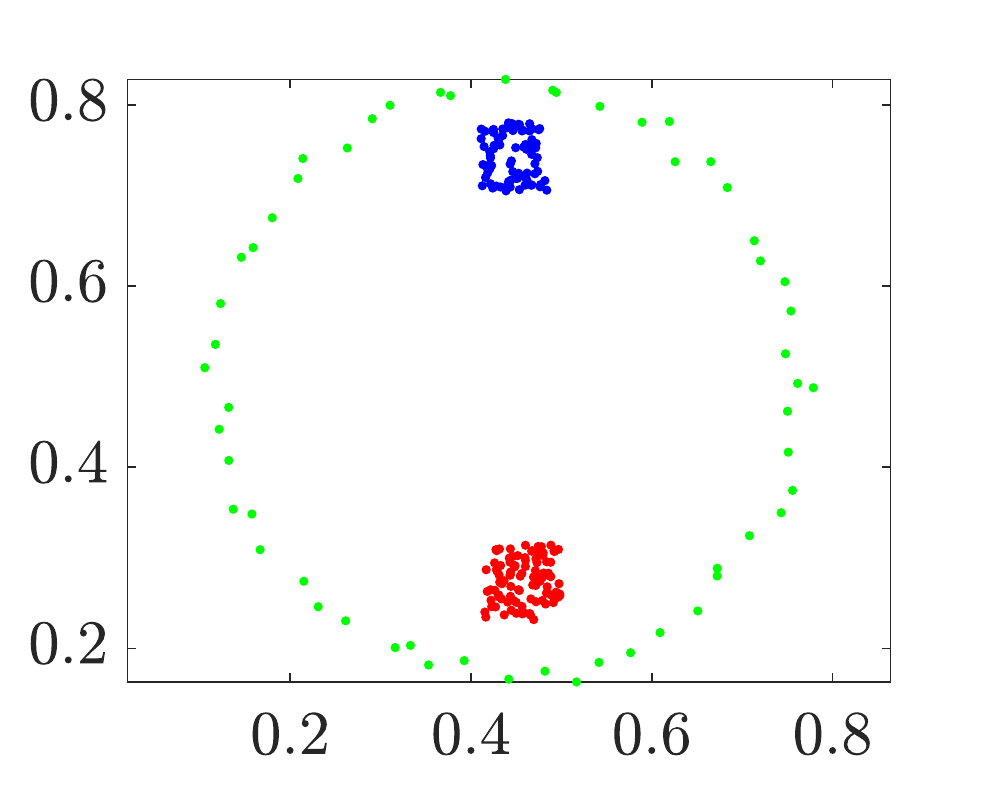}\\

\rotatebox[origin=l]{90}{\hspace{1mm}ALS-symNMF}&
\includegraphics[width=0.14\linewidth]{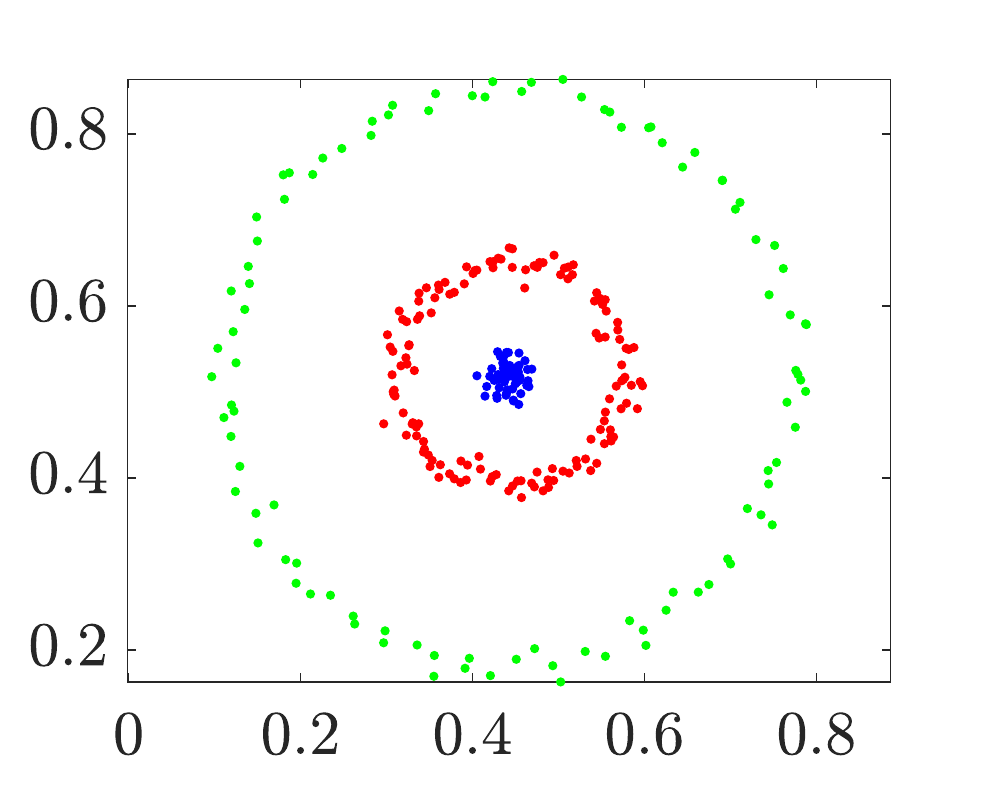}&
\includegraphics[width=0.14\linewidth]{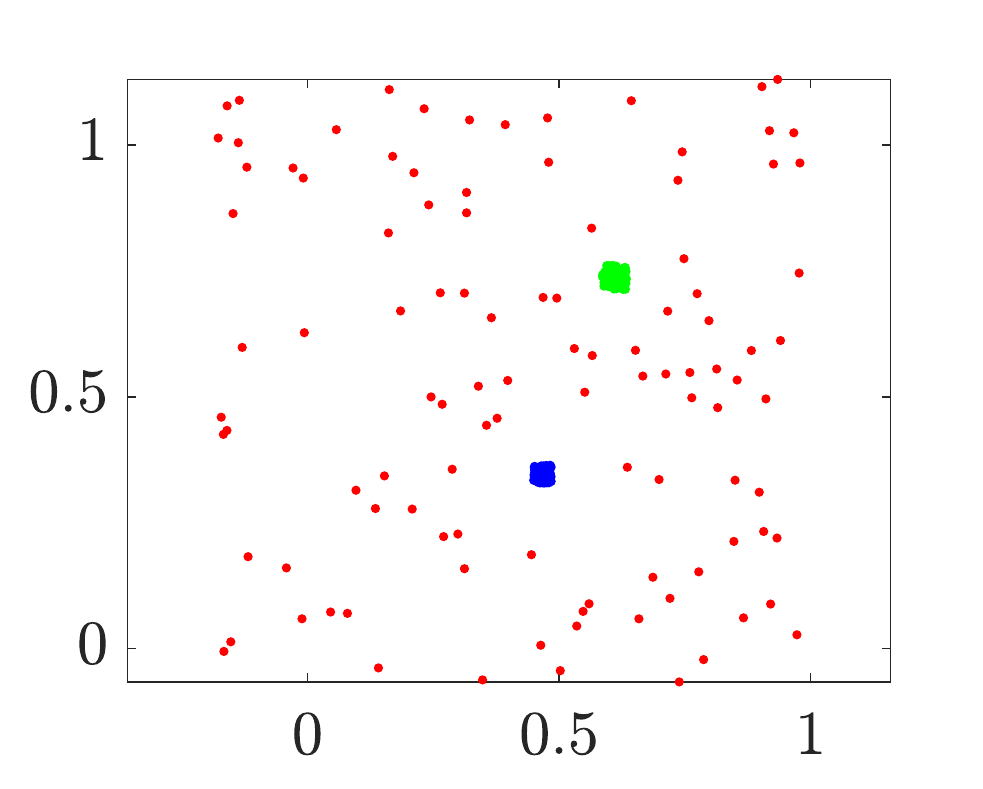}&
\includegraphics[width=0.14\linewidth]{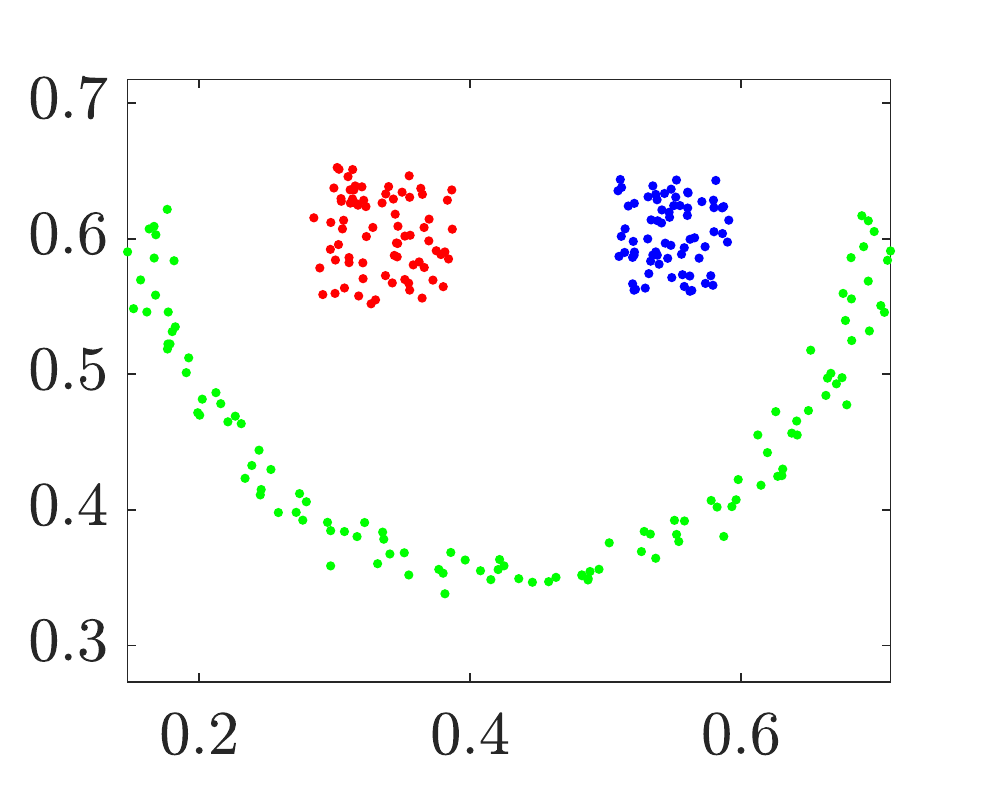}&
\includegraphics[width=0.14\linewidth]{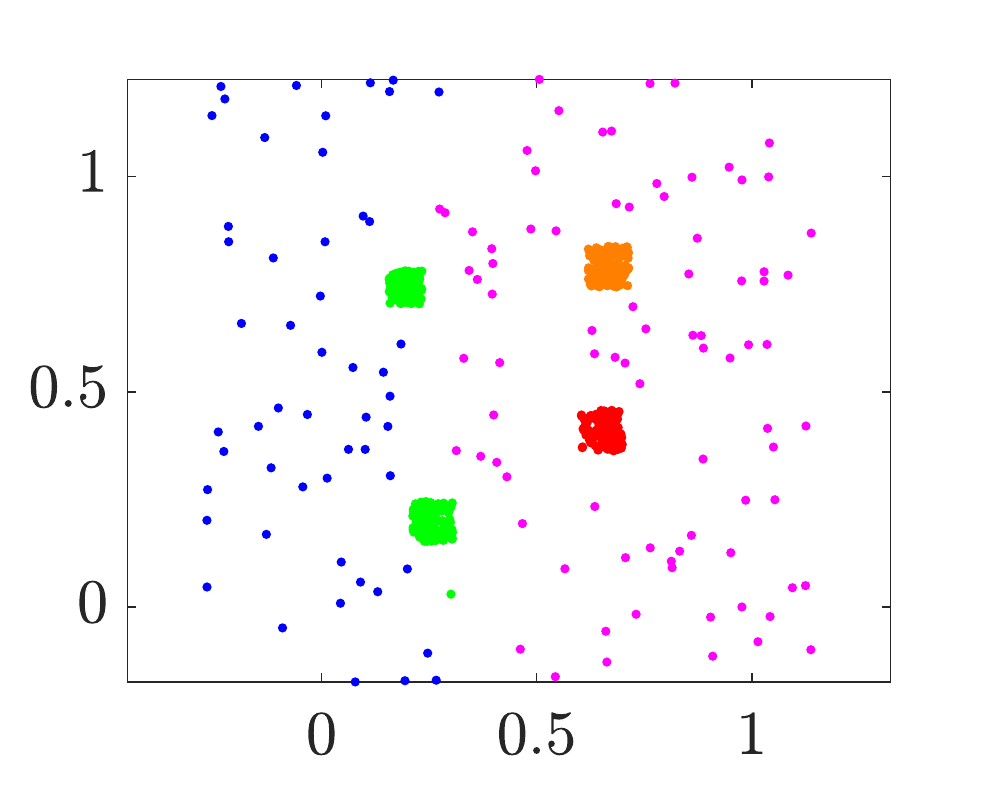}&
\includegraphics[width=0.14\linewidth]{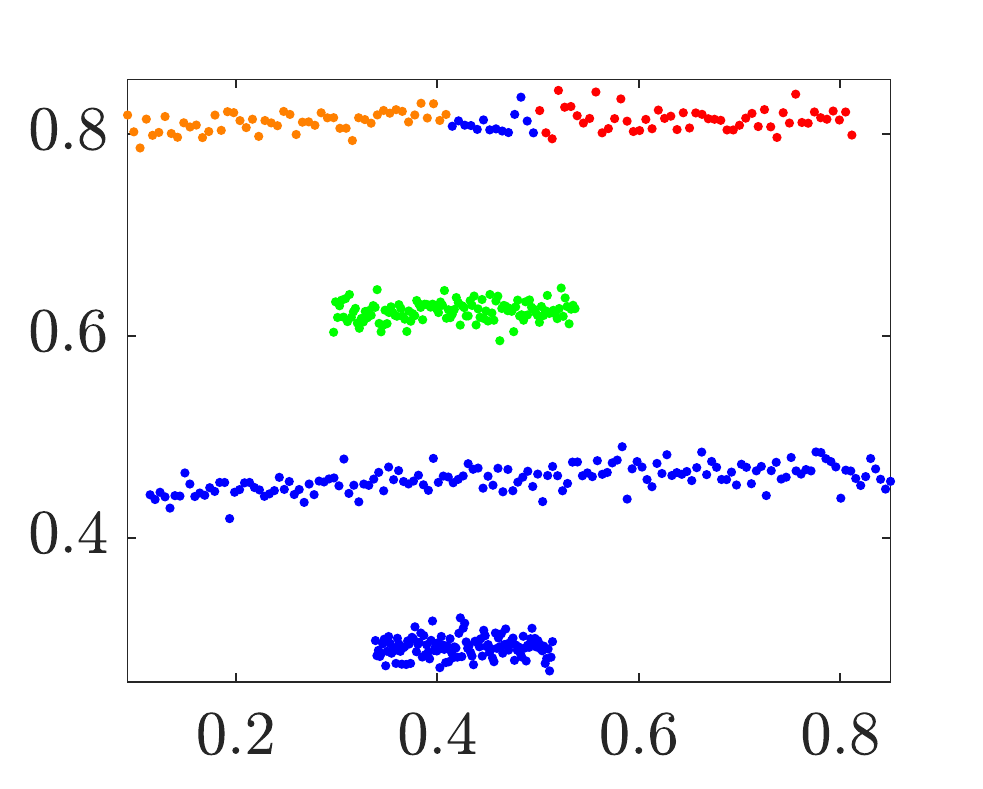}&
\includegraphics[width=0.14\linewidth]{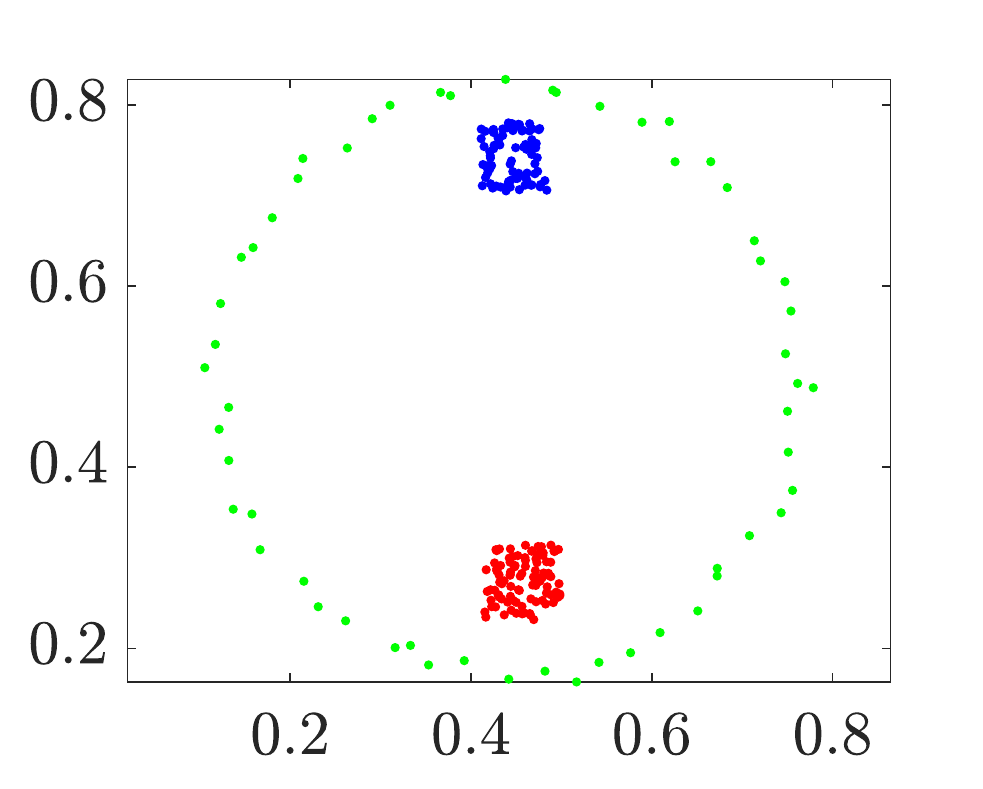}\\

\rotatebox[origin=l]{90}{\hspace{8mm}AP}&
\includegraphics[width=0.14\linewidth]{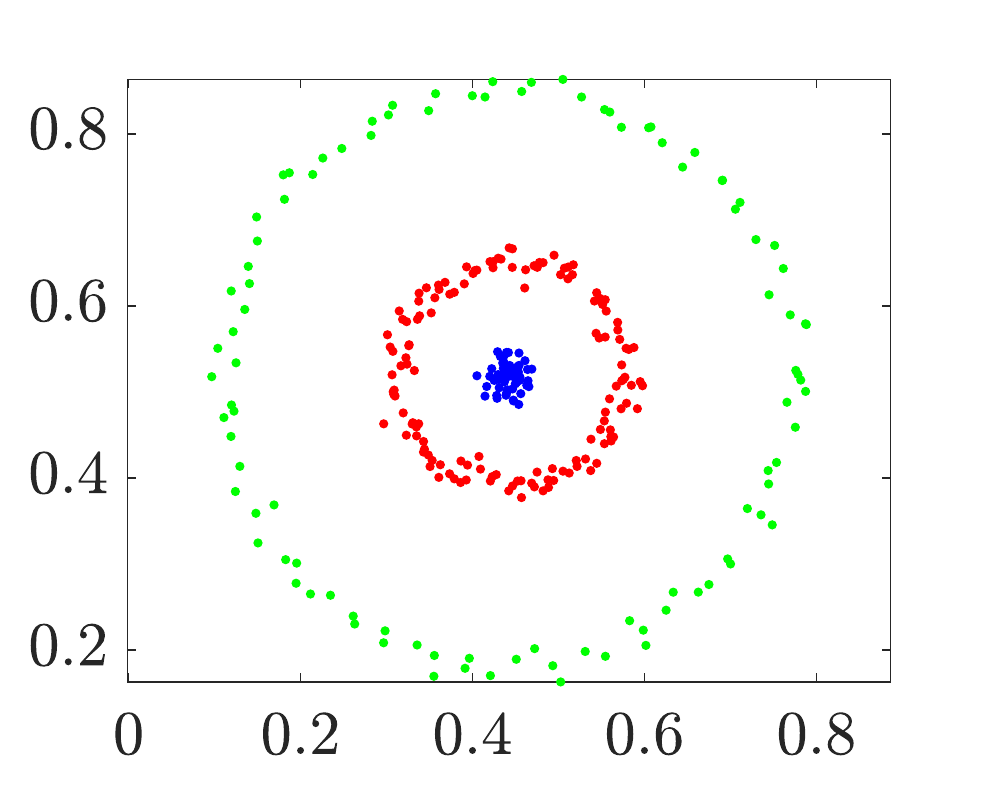}&
\includegraphics[width=0.14\linewidth]{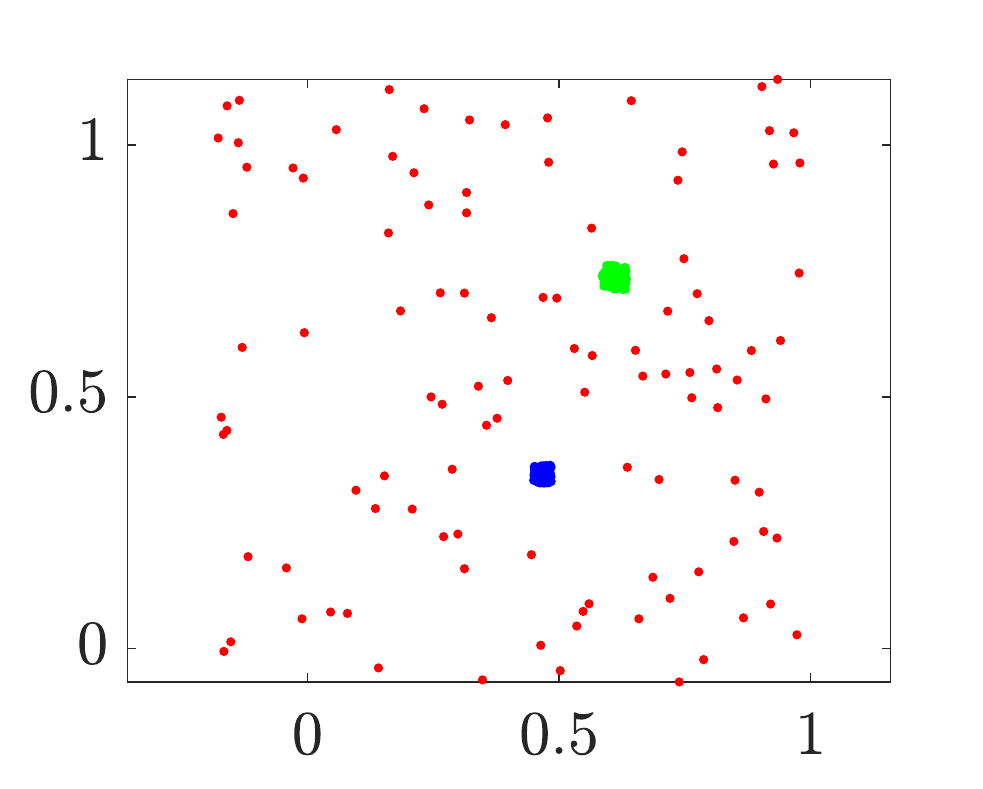}&
\includegraphics[width=0.14\linewidth]{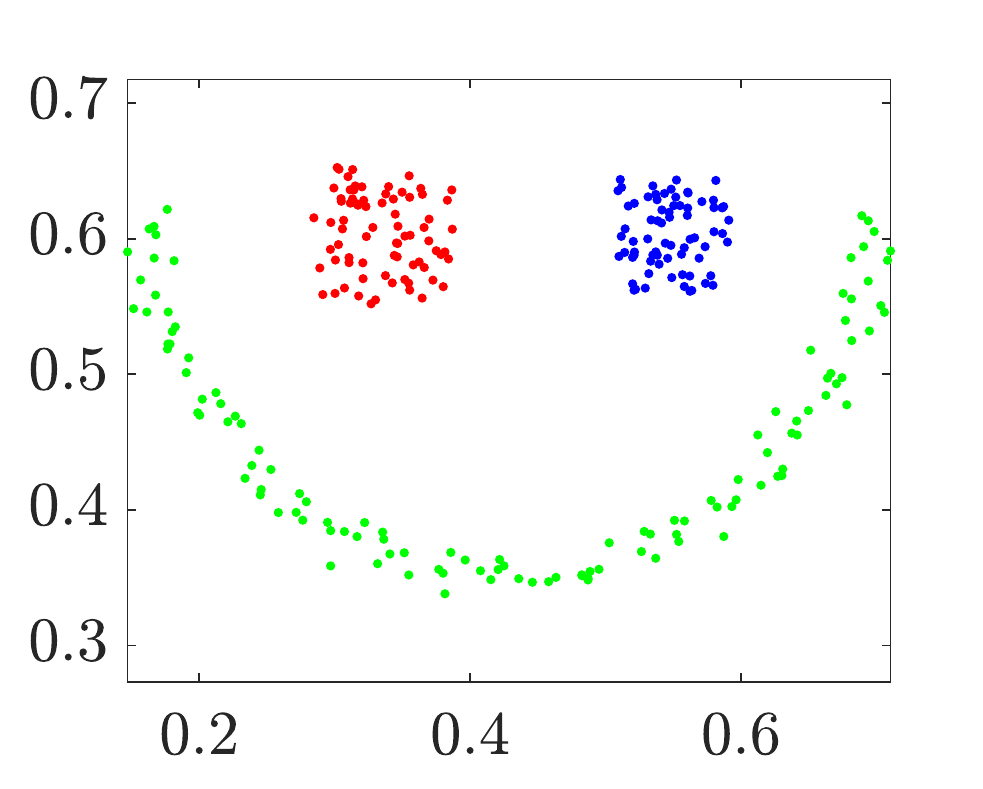}&
\includegraphics[width=0.14\linewidth]{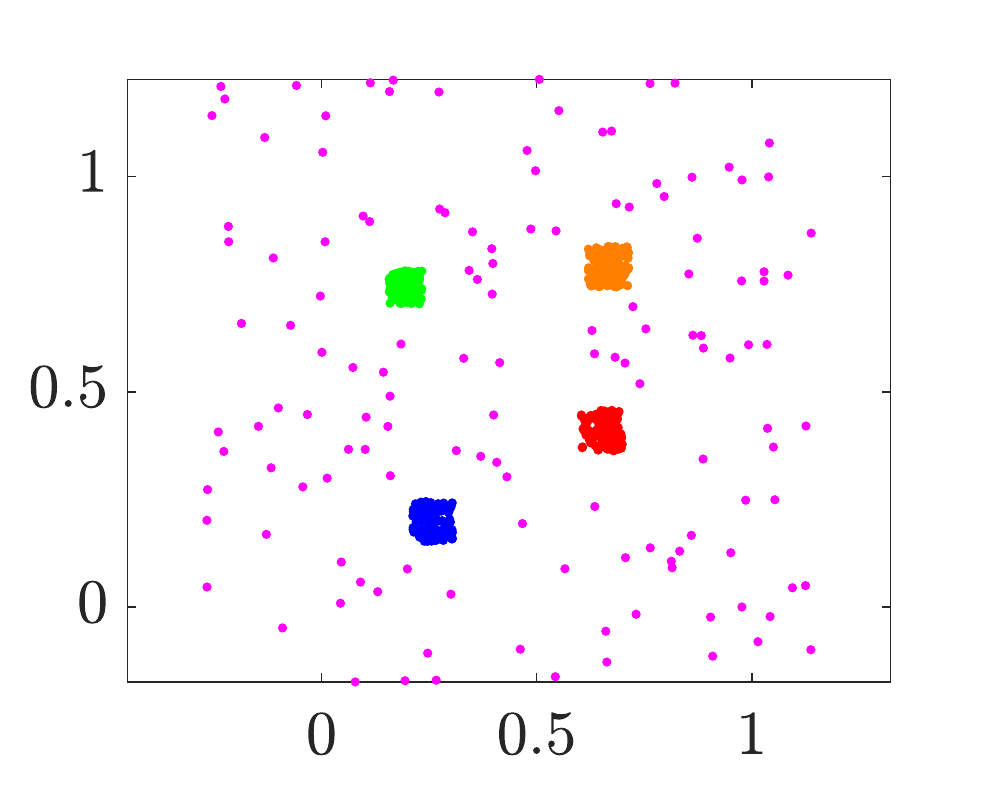}&
\includegraphics[width=0.14\linewidth]{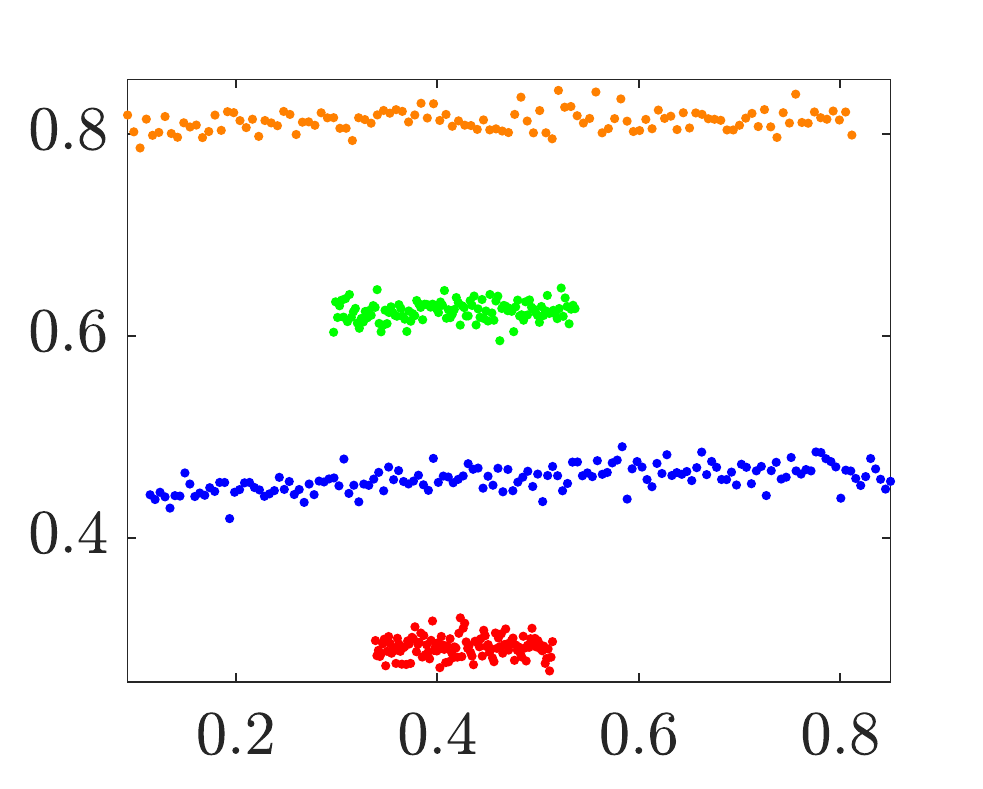}&
\includegraphics[width=0.14\linewidth]{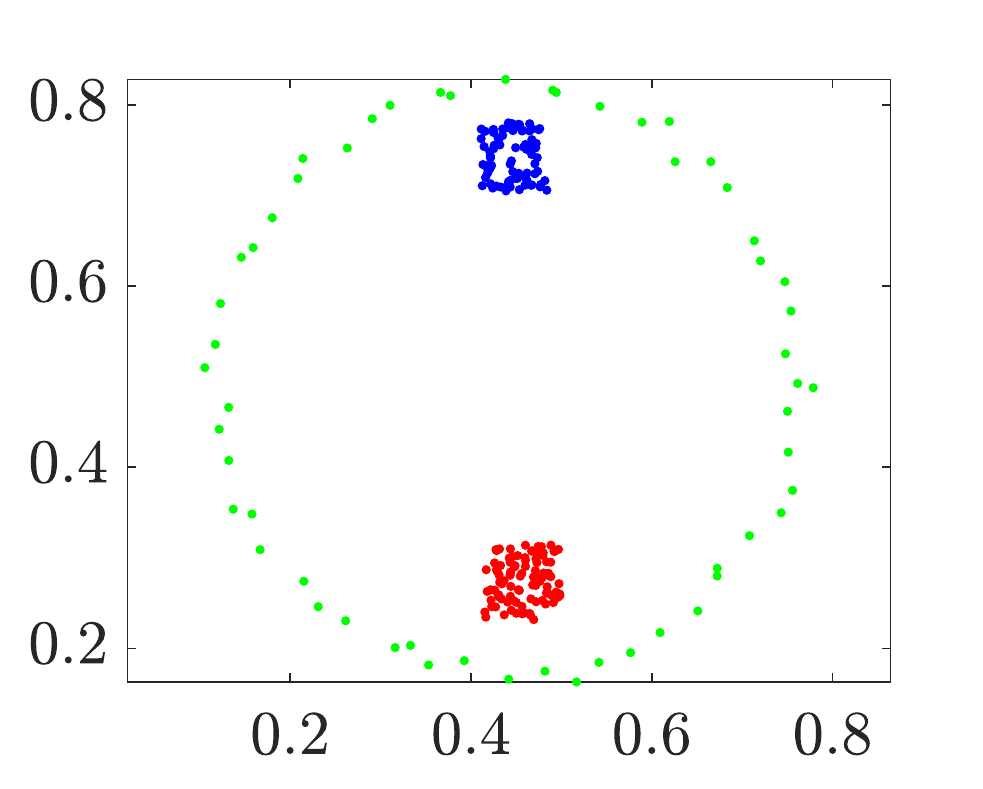}\\

\rotatebox[origin=l]{90}{\hspace{7mm}TAP}&
\includegraphics[width=0.14\linewidth]{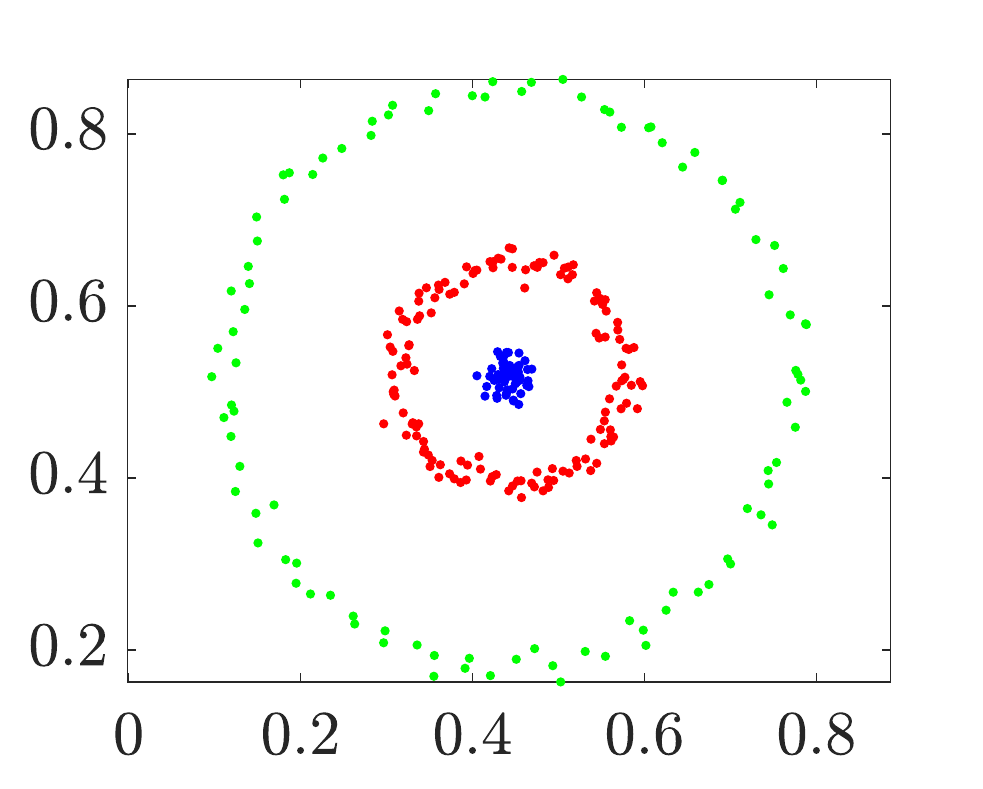}&
\includegraphics[width=0.14\linewidth]{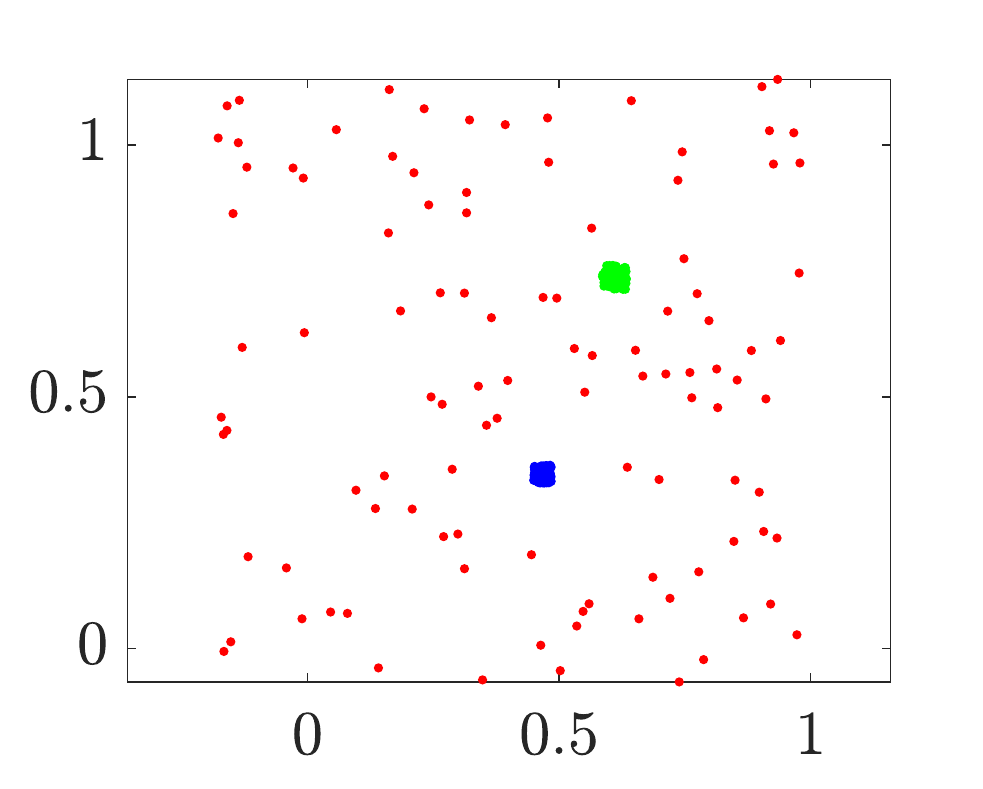}&
\includegraphics[width=0.14\linewidth]{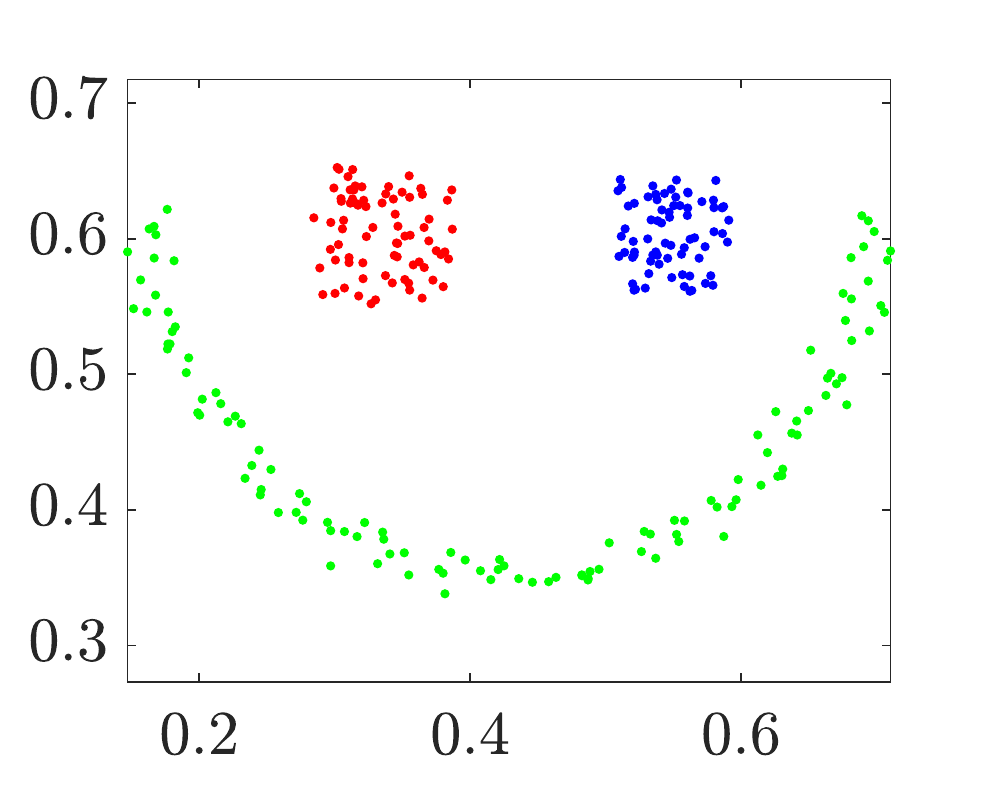}&
\includegraphics[width=0.14\linewidth]{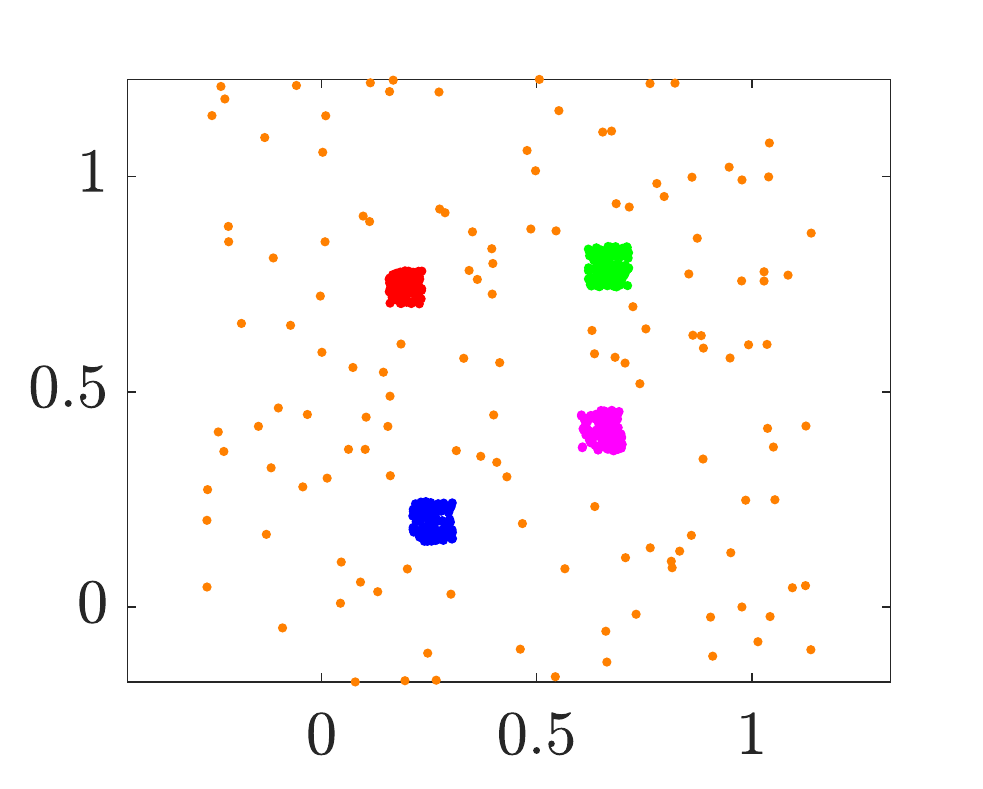}&
\includegraphics[width=0.14\linewidth]{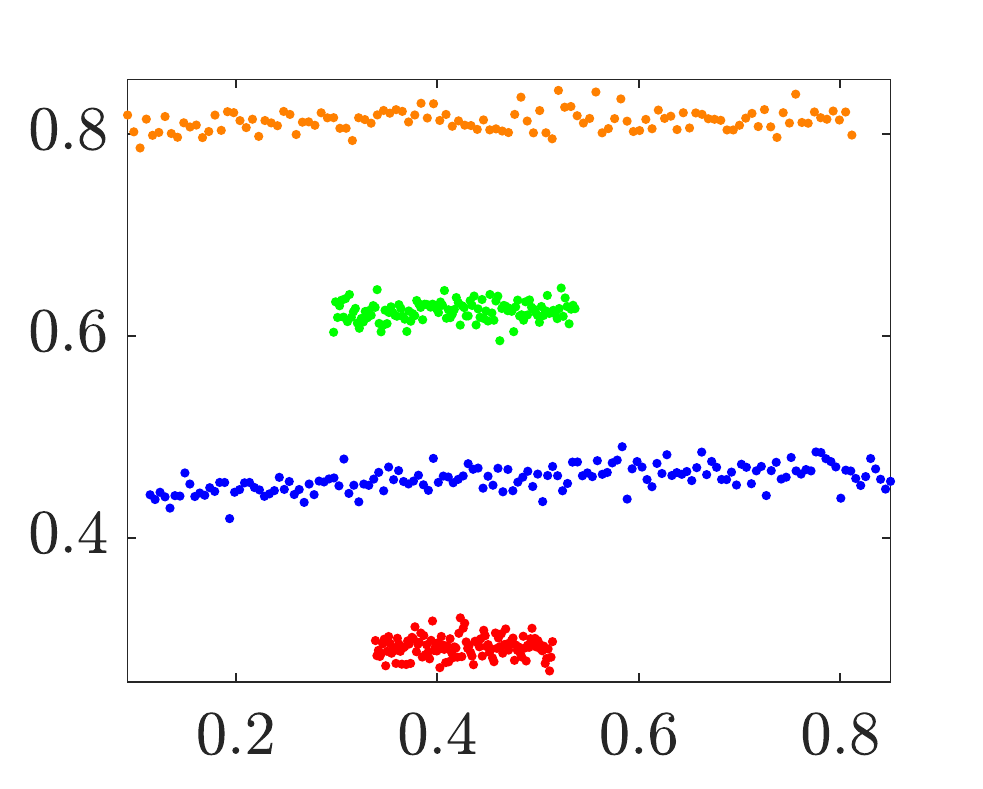}&
\includegraphics[width=0.14\linewidth]{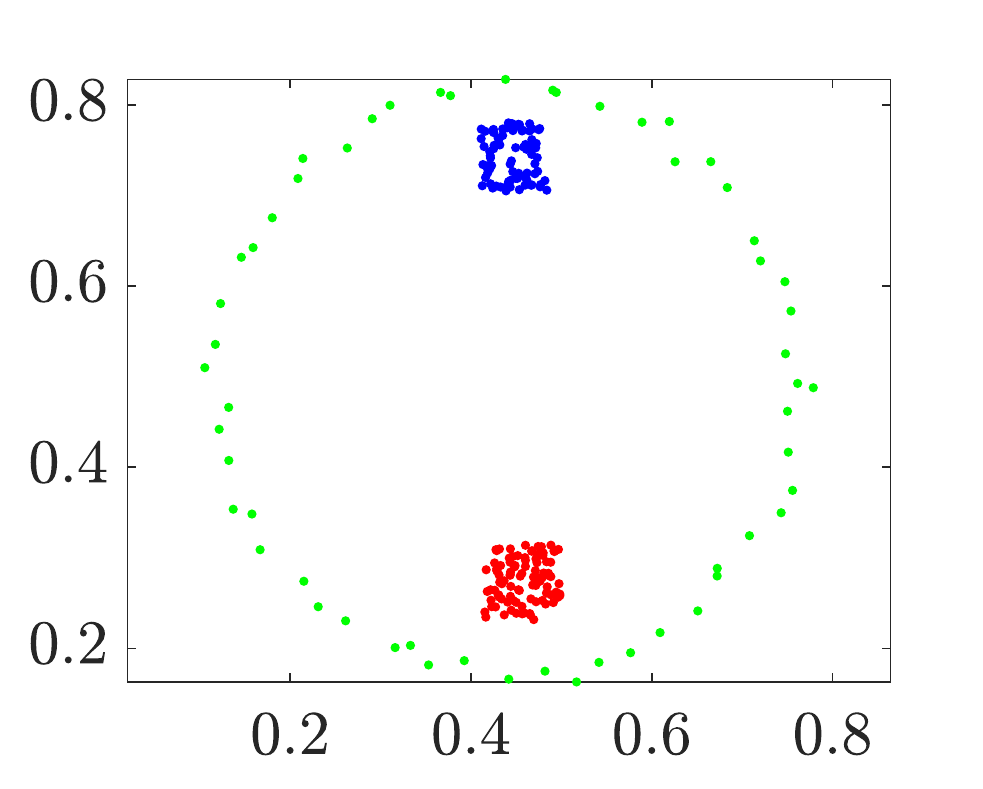}\\
\end{tabular}
\caption{The graph clustering results by the TAP (or AP) method and symmetric NMF methods on 6 cases of synthetic graph data. Different color represents different clusters. }
\label{fig-snmf-gc}
\end{figure*}
It is interesting whether a better nonnegative low rank matrix approximation could contribute to a better separable (or generalized separable) NMF result.
To further investigate whether nonnegative low rank matrix approximation could help separable and generalized separable NMF methods,
we conduct the experiments with inputting the nonnegative low rank approximation to separable and generalized separable NMF methods.
We adopt the accuracy and the distance to ground truth defined in Eqs. (16) and (17) of \cite{pan2019generalized} as the quantitative metrics.
The accuracy reports the proportion of correctly identified row and column indices while the distance to ground truth reports the relative errors between the identified important rows (columns) to the ground truth important rows (columns).
We present the computational results in Figure \ref{fig-gsnmf-ac-dgt}. When the noise level is between 0.1 and 1, the nonnegative low rank matrix
approximation by our TAP method
obviously enhances the accuracy and decrease the distance between the identified rows (columns) to the ground truth.

\subsection{Symmetric Nonnegative Matrices for Graph Clustering}

In this subsection, we test TAP and AP
methods on the symmetric matrices. It can readily be found that the output of TAP and AP algorithms would be symmetric if the input matrix is symmetric since that the projection onto the nonnegative matrix manifold or the low rank matrix manifold would never affect the symmetry. Here symmetric NMF methods are the coordinate descent algorithm (denoted as ``CD-symNMF '') \cite{vandaele2015coordinate}, the Newton-like algorithm (denoted as ``Newton-symNMF'') \cite{kuang2012symmetric}, and the alternating least squares algorithm (denoted as ``ALS-symNMF'') \cite{kuang2012symmetric}.

We perform experiments by using symmetric NMF methods, TAP and AP
methods on the synthetic graph data, which is reproduced from \cite{zelnik2005self} with six different cases.
The data points in 2-dimensional space are displayed in the first row of Figure \ref{fig-snmf-gc}.
Each case contains clear cluster structures. By following the procedures in \cite{kuang2012symmetric,zelnik2005self},
a similarity matrix ${\bf A} \in\mathbb{R}^{n\times n}$, where $n$ represents the number of data points,
is constructed to characterize the similarity between each pair of data points.
Each data point is assumed to be only connected to its nearest nine neighbors.
Given a specific pair of the $i$-th and $j$-th data points $x_i$ and $x_j$, we firstly construct the distance matrix $\mathbf D\in\mathbb{R}^{n\times n}$ with $ D_{ij} = D_{ji} = \|x_i-x_j\|_2^2$. Then, the similarity matrix is given as
\begin{equation}
A_{ij}=\left\{
\begin{aligned}
0,\ &\text{if}\  i=j,\\
e^{(\frac{-D_{ij}}{\sigma_i\sigma_j})},\  &\text{if}\  i\neq j,
\end{aligned}
\right.
\end{equation}
where $\sigma_i$ is the Euclidean distance between the $i$-th data point $x_i$ and its 9-th neighbor.
Then, we perform NMF, TAP and AP methods for ${\bf A}$.

The clustering results of the symmetric NMF methods and
nonnegative low rank matrix approximation are obtained by using $k$-means method on
${\bf B}$ and ${\bf U}$ respectively.
The clustering results are shown in Figure \ref{fig-snmf-gc}.
CD-symNMF method fails in most examples except the example in the second column.
Both Newton-symNMF and ALS-symNMF methods fail in the example in the fifth column.
However, TAP and AP methods perform well for all the examples.
The average computational time required by the proposed TAP method (0.0321 seconds) is less than
that (0.1035 seconds) required by the AP method. The proposed TAP is faster than
the AP method.

\subsection{Orthogonal Decomposable Non-negative Matrices}\label{sec4.5}

In this subsection, we test TAP and AP methods and orthogonal NMF (ONMF) methods \cite{choi2008algorithms,ding2006orthogonal} on the approximation of the synthetic data and the unmixing of hyperspectral images.
The orthogonal NMF method is a multiplicative updating algorithm proposed by Ding et al. \cite{ding2006orthogonal}. We refer to
Ding-Ti-Peng-Park (DTPP)-ONMF.  A multiplicative updating algorithm utilizing the true gradient in Stiefel manifold
is proposed in \cite{choi2008algorithms}. We refer to SM-ONMF.

\begin{figure}[!t]
\centering
\includegraphics[width=0.99\linewidth]{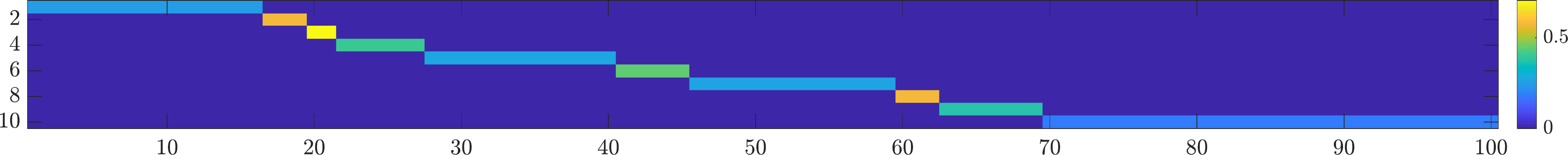}
\caption{An illustration of the generated ${\bf B}^T$.}
\label{fig-orth-B}
\end{figure}

We construct an orthogonal nonnegative matrix ${\bf B}\in\mathbb{R}^{100\times10}$, whose transpose is shown in Figure \ref{fig-orth-B}. Then a matrix ${\bf C}\in\mathbb{R}^{10\times30}$ is generated with entries uniformly distributed in $[0,1]$. Then, we obtain an orthogonal decomposable
matrix ${\bf A} = {\bf B} {\bf C}\in \mathbb{R}^{100\times30}$. Next, a noise matrix based on MATLAB command $\sigma\times{\tt rand}(100, 30)$ is added to ${\bf A}$. We set $\sigma = 0,0.02,0.04,\cdots,0.1$. The relative approximation errors of the results by different methods are shown in Table \ref{Tab-orth-syn}.
We can see that the approximation errors of TAP and AP methods are the lowest among the testing examples.

As a real-world application of ONMF,
hyperspectral image unmixing aims at factoring the observed hyperspectral image in matrix format into an endmember matrix and an abundance matrix. The abundance matrix is indeed the classification of the pixels to different clusters, with each corresponding to a material (endmember).
In this part, we use a sub-image of the Samson data set \cite{zhu2014spectral}, consisting of $95\times95=9025$ spatial pixels and 156 spectral bands.
We form a matrix $A$ of size $9025\times156$ to represent this sub-image.
Three different materials, i.e., ``Tree'', ``Rock'', and ``Water'', are in this sub-image, and we set the rank $r$ as 3.
The factor matrices ${\bf B}\in\mathbb{R}^{9025 \times3}$ and ${\bf C}\in\mathbb{R}^{3\times156}$ can be obtained by the orthogonal NMF methods.
We use k-means and do hard clustering on ${\bf B}\in\mathbb{R}^{9025 \times3}$ to obtain the abundance matrix, and we can obtain the $i$-th feature image by reshaping its $i$-th column
to a $95\times95$ matrix.
Each row of ${\bf C}$ represents the spectral reflectance of on material (``Tree'', ``Rock'', or ``Water'').
As for TAP and AP methods, we apply singular value decomposition on approximated non-negative low rank matrices to obtain the left singular value matrices which contain the first 3 left singular vectors. Then, we use k-means and do hard clustering on the left singular matrices to cluster three materials and obtain abundance matrices and endmember matrices.

\begin{table}[!t]
\renewcommand\arraystretch{1.0}\setlength{\tabcolsep}{4pt}
\centering
\caption{The relative approximation errors ($\times100$) on the orthogonal symmetric matrix data. The \textbf{best} values and the \underline{second best} values are respectively highlighted by bolder fonts and underlines.}
\begin{tabular}{c c c c c c c c c c c c c c}
\toprule
$\sigma$& 0 & 0.02 & 0.04 & 0.06 & 0.08 & 0.1\\
\midrule
DTPP-ONMF& 0.022 & \underline{2.730} & 5.231 & 7.567 & 9.465 & \underline{11.232} \\
SM-ONMF& \underline{0.016} & 2.741 & \underline{5.169} & \underline{7.533} & \underline{9.424} & 14.180 \\
AP & \bf0.000 & \bf2.364 & \bf4.471 & \bf6.529 & \bf8.215 & \bf9.700 \\
TAP & \bf0.000 & \bf2.364 & \bf4.471 & \bf6.529 & \bf8.215 & \bf9.700 \\
\bottomrule
\end{tabular}%
\label{Tab-orth-syn}
\end{table}
\begin{table}[!t]
\renewcommand\arraystretch{1.0}\setlength{\tabcolsep}{4pt}
\centering\scriptsize
\caption{The quantitative metrics of the unmixing results on the hyperspetral image Samson. The \textbf{best} values and the \underline{second best} values are respectively highlighted by bolder fonts and underlines.}
\begin{tabular}{c c c c c c c c c c c c c c}
\toprule
Metric& DTPP-ONMF&SM-ONMF&AP&TAP\\\midrule
SAD & \underline{0.3490} & 0.4389 & \bf 0.0765 & \bf 0.0765 \\
Similartity & \underline{0.5887} & 0.5640 & \bf{0.9383} & \bf{0.9383}\\
\bottomrule
\end{tabular}%
\label{Tab-hsi-um}
\end{table}

\begin{figure}[!t]
\centering
\includegraphics[width=0.95\linewidth]{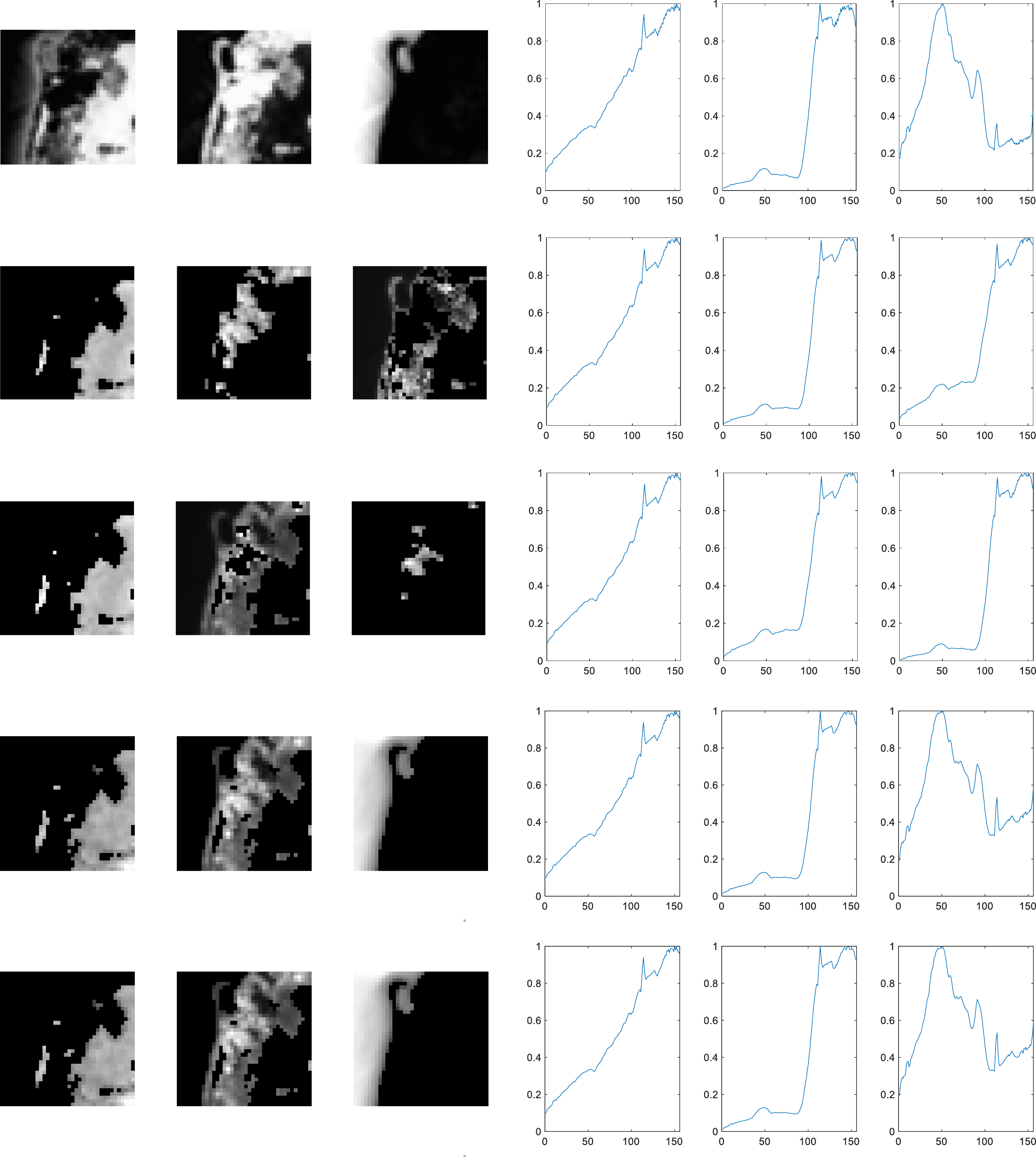}
\caption{Left: Rock, Tree, Water; Right: Reflectance of Rock, Tree, Water. From the top to
bottom: groundtruth, DTPP-ONMF, SM-ONMF, AP, TAP.}
\label{fig-hsi-um}
\end{figure}

To quantitatively evaluate the umixing results, we employ two metrics. The first one is the spectral angle distance (SAD) as follows:
$$
\text{SAD} = \frac{1}{r}\sum_{i=1}^{r}\arccos \left(
\frac{s_i^T \hat{s}_i}{\|s_i\|_2\|\hat{s}_i\|_2}
\right),
$$
where $\{s_i\}_{i=1}^r$  are the estimated spectral reflectance (rows of the endmember matrix) and $\{\hat{s}_i\}_{i=1}^r$ are the groundtruth spectral reflectance.
The second one is the similarity of the abundance feature image \cite{pan2019orthogonal} as follows:
$$
\text{Similarity } = \frac{1}{r}\sum_{i=1}^{r}
\frac{a_i^T \hat{a}_i}{\|a_i\|_2\|\hat{a}_i\|_2},
$$
where $\{a_i\}_{i=1}^r$  are the estimated abundance feature (columns
of the abundance matrix) and $\{\hat{a}_i\}_{i=1}^r$ are the groundtruth ones.
We note that a larger Similarity and a smaller SAD indicate a better unmixing result.
We exhibit the quantitative metrics in Table \ref{Tab-hsi-um}.
We can evidently see that the proposed TAP and AP methods obtain the best metrics.
Meanwhile, we illustrate the estimated spectral reflectance and abundance feature images in Figure \ref{Tab-hsi-um}.
It can be found from the second row that DTPP-ONMF and SM-ONMF perform well for the materials ``Rock'' and ``Tree'' but poor on ``Water''.
TAP and AP methods unmix these three materials well, but
the proposed TAP method (the computational time = 0.1492 seconds)
is faster than
the AP method (the computational time = 0.3738 seconds).

\section{Conclusion}\label{sec:clu}

In this paper, we have proposed a new alternating projection method to compute
nonnegative low rank matrix approximation for nonnegative matrices.
Our main idea is to use the
tangent space of the point in the fixed-rank matrix manifold
to approximate the projection onto the manifold in order to reduce the computational cost.
Numerical examples in data clustering, pattern recognition and hyperspectral data
analysis have shown that the proposed alternating projection method is
better than that of nonnegative matrix factorization methods
in terms of accuracy, and the computational time required by the proposed
alternating projection method is less than that required by
the original alternating projection method.

Moreover, we have shown that
the sequence generated by the alternating projections onto the tangent spaces of the fixed rank matrices manifold and the nonnegative matrix manifold,
converge linearly to a point in the intersection of the two manifolds where the convergent point
is sufficiently close to optimal solutions.
Our theoretical convergence results are new and are not studied in the literauture.
We remark that Andersson and Carlsson
\cite{andersson2013alternating}
assumed that the exact projection onto each manifold and then
obtained the convergence result of the alternating projection method.
Because of our proposed inexact projection onto each manifold, our proof can be
extended to show
the sequence generated by alternating projections on one or two nontangential manifolds
based on tangent spaces, converges linearly to a point in the intersection of
the two manifolds.

As a future research work, it is interesting to study
(i) the convergence results when inexact projections on several manifolds are employed,
and (ii) applications where
the other norms (such as $l_1$ norm) in data fitting instead
of Frobenius norm. It is
necessary to develop the related algorithms for such manifold optimization problems.

\appendices
\section{Proof of Lemma \ref{ne1}, \ref{new1} and \ref{new2}}
\subsection{Proof of Lemma \ref{ne1}}\label{Proof1}
\begin{proof}
For a given $\epsilon>0,$ there exist an $s_{1}(\epsilon)$ such that Lemma \ref{newproof1} applies to $\mathcal{M}_{r} \cap Ball({\bf P},s_{1}(\epsilon))$.
Since $\pi_1(\cdot)$ and $P_{T_{\mathcal{M}_{r}}(\cdot)}(\cdot)$ are continuous,
there exist an $s(\epsilon)<s_{1}(\epsilon)$ such that the image of
$Ball({\bf P},s({\epsilon}))$ under $\pi_{1}$ and $P_{T_{\mathcal{M}_{r}}}$ is
included in $Ball({\bf P},s_{1}({\epsilon}))$.
Now we can choose a point ${\bf Q}\in Ball({\bf P},s({\epsilon}))\cap \mathcal{M}_{r}$.
For any given ${\bf Z}\in Ball({\bf P},s({\epsilon}))$, there
are two cases: ${\bf Z} \in {\cal M}_r$ and ${\bf Z} \notin {\cal M}_r$.
When ${\bf Z} \in {\cal M}_r$, by using Lemma \ref{newproof1} (i) with
${\cal C} = {\bf Q}$ and ${\bf D} = {\bf Z} = \pi_1({\bf Z})$, \eqref{lemma32} follows.

Next we consider ${\bf Z} \notin {\cal M}_r$.
In this case, we set ${\bf C} = {\bf Q}$ and ${\bf D} =
P_{T_{\mathcal{M}_{r}}({\bf Q})}({\bf Z})$. As ${\bf Z}
\in Ball({\bf P},s({\epsilon}))$ and
$P_{T_{\mathcal{M}_{r}}({\bf Q})}({\bf Z}) \in Ball({\bf P},s_{1}({\epsilon}))$.
By using Lemma \ref{newproof1} (ii), we have
\begin{align*}
\| P_{T_{\mathcal{M}_{r}}({\bf Q})}({\bf Z})-\pi_{1}( P_{T_{\mathcal{M}_{r}}({\bf Q})}({\bf Z}))\|_{F}\leq \epsilon\|P_{T_{\mathcal{M}_{r}}({\bf Q})}({\bf Z})-{\bf Q}\|_{F}.
\end{align*}
It implies that the set  $$\mathcal{M}_{r}\cap Ball (P_{T_{\mathcal{M}_{r}}({\bf Q})}({\bf Z}), \epsilon \|P_{T_{\mathcal{M}_{r}}({\bf Q})}({\bf Z})-{\bf Q}\|_{F}) $$ is not void and is included in $Ball ({\bf Z}, \|{\bf Z}-{\bf Z}_{1}\|_{F}+\epsilon\|{\bf Z}_{1}-{\bf Q}\|_{F})$ with ${\bf Z}_{1}=P_{T_{\mathcal{M}_{r}}({\bf Q})}({\bf Z})$.
Note that $\pi_{1}({\bf Z})$ is on the manifold $\mathcal{M}_{r}$ and is included in $Ball({\bf P},s_{1}({\epsilon}))$. By using Lemma \ref{newproof1} (i)
(set ${\bf C}={\bf Q}$ and ${\bf D}=\pi_{1}({\bf Z})$), we have
\begin{align}\label{new11}
\|\pi_{1}({\bf Z})-P_{T_{\mathcal{M}_{r}}({\bf Q})}(\pi_{1}({\bf Z}))\|_{F}\leq \epsilon\|\pi_{1}({\bf Z})-{\bf Q}\|_{F}.
\end{align}
Here the three points $\pi_{1}({\bf Z}),$ $P_{T_{\mathcal{M}_{r}}({\bf Q})}(\pi_{1}({\bf Z}))$ and ${\bf Q}$ form a right triangle, we have
\begin{align*}
&~\|\pi_{1}({\bf Z})-P_{T_{\mathcal{M}_{r}}({\bf Q})}(\pi_{1}({\bf Z}))\|^2_{F}+\|P_{T_{\mathcal{M}_{r}}({\bf Q})}(\pi_{1}({\bf Z}))-{\bf Q}\|^2_{F}\\
=&~\|\pi_{1}({\bf Z})-{\bf Q}\|^2_{F}.
\end{align*}
By using this equality in the calculation of (\ref{new11}), we obtain
\begin{align}\label{newp1}
&~\|\pi_{1}({\bf Z})-P_{T_{\mathcal{M}_{r}}({\bf Q})}(\pi_{1}({\bf Z}))\|_{F}\nonumber\\
=&~\sqrt{\|\pi_{1}({\bf Z})-{\bf Q}\|^2_{F}-\|P_{T_{\mathcal{M}_{r}}({\bf Q})}(\pi_{1}({\bf Z}))-{\bf Q}\|^2_{F}}\nonumber\\
\leq &~\varphi(\epsilon)\|P_{T_{\mathcal{M}_{r}}({\bf Q})}(\pi_{1}({\bf Z}))-{\bf Q}\|_{F},
\end{align}
with $\varphi(\epsilon)=\frac{\epsilon}{\sqrt{1-\epsilon^2}}.$
As ${\bf Z}_{1}=P_{T_{\mathcal{M}_{r}}({\bf Q})}({\bf Z})$, we know that
$({\bf Z}-{\bf Z}_{1}) \perp T_{\mathcal{M}_{r}}({\bf Q}).$ By using
$P_{T_{\mathcal{M}_{r}}({\bf Q})}(\pi_{1}({\bf Z}))\in T_{\mathcal{M}_{r}}({\bf Q})$,
we find that
$P_{T_{\mathcal{M}_{r}}({\bf Q})}(\pi_{1}({\bf Z}))$,
${\bf Z}_{1}$ and ${\bf Z}$ form a right-angled triangle
which is included in $Ball ({\bf Z}, \|{\bf Z}-{\bf Z}_{1}\|_{F}+\epsilon\|{\bf Z}_{1}-{\bf Q}\|_{F})$.
It implies that
\begin{align*}
&~\|P_{T_{\mathcal{M}_{r}}({\bf Q})}(\pi_{1}({\bf Z}))-{\bf Z}_{1}\|_{F}\\
\leq&~ \|{\bf Z}-P_{T_{\mathcal{M}_{r}}({\bf Q})}(\pi_{1}({\bf Z}))\|_{F}
\leq \|{\bf Z}-{\bf Z}_{1}\|_{F}+\epsilon\|{\bf Z}_{1}-{\bf Q}\|_{F},
\end{align*}
and
\begin{align}
&~\|P_{T_{\mathcal{M}_{r}}({\bf Q})}(\pi_{1}({\bf Z}))-{\bf Q}\|_{F}\nonumber\\
=&~\|P_{T_{\mathcal{M}_{r}}({\bf Q})}(\pi_{1}({\bf Z}))-{\bf Z}_{1}+{\bf Z}_{1}-{\bf Q}\|_{F}\nonumber\\
\leq&~ \|P_{T_{\mathcal{M}_{r}}({\bf Q})}(\pi_{1}({\bf Z}))-{\bf Z}_{1}\|_{F}+\|{\bf Z}_{1}-{\bf Q}\|_{F}\nonumber\\
<&~\|{\bf Z}-{\bf Z}_{1}\|_{F}+\epsilon\|{\bf Z}_{1}-{\bf Q}\|_{F}+\|{\bf Z}_{1}-{\bf Q}\|_{F}.\label{newp2}
\end{align}
By combing \eqref{newp1} and \eqref{newp2} with $0 < \epsilon < \frac{3}{5}$, we derive
\begin{align*}
&~\|\pi_{1}({\bf Z})-P_{T_{\mathcal{M}_{r}}({\bf Q})}(\pi_{1}({\bf Z}))\|_{F}\\
< &~ \varphi(\epsilon)((1+\epsilon)\|{\bf Z}_{1}-{\bf Q}\|_{F}+
\|{\bf Z}-{\bf Z}_{1}\|_{F}) \\
 < &~ 2\epsilon (\|{\bf Z}_{1}-{\bf Q}\|_{F}+\|{\bf Z}-{\bf Z}_{1}\|_{F}).
\end{align*}
Now we set ${\bf Z}_{2}$ as the reflection point of
of $\pi_{1}({\bf Z})$ with repsect to the
tangent space $T_{\mathcal{M}_{r}}({\bf Q})$.
It is clear that
$$
\|\pi_{1}({\bf Z})-P_{T_{\mathcal{M}_{r}}({\bf Q})}(\pi_{1}({\bf Z}))\|_{F}
=\|P_{T_{\mathcal{M}_{r}}({\bf Q})}(\pi_{1}({\bf Z}))-{\bf Z}_{2}\|_{F}.
$$
Along the direction of ${\bf Z}-{\bf Z}_1$, we can find a point
${\bf Z}_{3}$ such that
$$
\|{\bf Z}_{1}-{\bf Z}_{3}\|_{F}=
\|\pi_{1}({\bf Z})-P_{T_{\mathcal{M}_{r}}({\bf Q})}(\pi_{1}({\bf Z}))\|_{F}.
$$
Thus we can estimate the distance between
$P_{T_{\mathcal{M}_{r}}({\bf Q})}({\bf Z})$ and $P_{T_{\mathcal{M}_{r}}({\bf Q})}(\pi_{1}({\bf Z}))$ as follows:
\begin{align*}
&~\|P_{T_{\mathcal{M}_{r}}({\bf Q})}({\bf Z})-P_{T_{\mathcal{M}_{r}}({\bf Q})}(\pi_{1}({\bf Z}))\|^2_{F}\\
=&~\|{\bf Z}-{\bf Z}_{2}\|^2_{F}-\|{\bf Z}-{\bf Z}_{3}\|^2_{F}\\
\leq&~(\|{\bf Z}-{\bf Z}_{1}\|_{F}+\epsilon\|{\bf Z}_{1}-{\bf Q}\|_{F})^{2}\\
&~-(\|{\bf Z}-{\bf Z}_{1}\|_{F}-2\epsilon (\|{\bf Z}_{1}-{\bf Q}\|_{F}+\|{\bf Z}-{\bf Z}_{1}\|_{F}))^2\\
=&~(\|{\bf Z}-{\bf Z}_{1}\|_{F}+\epsilon\|{\bf Z}_{1}-{\bf Q}\|_{F})^{2}\\
&~-((1-2\epsilon)\|{\bf Z}-{\bf Z}_{1}\|_{F}-2\epsilon \|{\bf Z}_{1}-{\bf Q}\|_{F})^2.
\end{align*}
With the above inequalities,
for any given ${\bf Z}\in Ball({\bf P},s(\epsilon))$, we have
\begin{align*}
&~\|\pi_{1}({\bf Z})-P_{T_{\mathcal{M}_{r}}({\bf Q})}({\bf Z})\|^2_{F}\\
=&~\|P_{T_{\mathcal{M}_{r}}({\bf Q})}({\bf Z})-P_{T_{\mathcal{M}_{r}}({\bf Q})}(\pi_{1}({\bf Z}))\|_{F}^{2}\\
&~+\|\pi_{1}({\bf Z})-P_{T_{\mathcal{M}_{r}}({\bf Q})}(\pi_{1}({\bf Z}))\|_{F}^{2}\\
\leq &~(\|{\bf Z}-{\bf Z}_{1}\|_{F}+\epsilon\|{\bf Z}_{1}-{\bf Q}\|_{F})^{2}
-((1-2\epsilon)\|{\bf Z}-{\bf Z}_{1}\|_{F}\\
&~-2\epsilon \|{\bf Z}_{1}-{\bf Q}\|_{F})^2+4\epsilon^2 (\|{\bf Z}_{1}\|_{F}+\|{\bf Z}_{2}\|_{F})^2\\
=&~2\epsilon \|{\bf Z}-{\bf Z}_{1}\|_{F}\|{\bf Z}_{1}-{\bf Q}\|_{F}+\epsilon^{2}\|{\bf Z}_{1}-{\bf Q}\|^2_{F}\\
&~+4\epsilon(\|{\bf Z}-{\bf Z}_{1}\|^2_{F}+\|{\bf Z}_{1}-{\bf Q}\|_{F}\|{\bf Z}-{\bf Z}_{1}\|_{F})\\
<&~16\epsilon\|{\bf Z}-{\bf Q}\|_{F}.
\end{align*}
The second inequality is derived by using
the facts that $\|{\bf Z}_{1}-{\bf Q}\|_{F}<\|{\bf Z}-{\bf Q}\|_{F},$
$\|{\bf Z}_{2}-{\bf Q}\|_{F}<\|{\bf Z}-{\bf Q}\|_{F}$ and $\epsilon^2<\epsilon$.
Hence the result follows.
\end{proof}

\subsection{Proof of Lemma \ref{new1}}\label{Proof2}
\begin{proof} Note that
$\pi_{1}(P_{T_{\mathcal{M}_{r}}({\bf Q})}({\bf Z}))$ and $\pi_{1}({\bf Z})$
are on the manifold $\mathcal{M}_{r}$, and $\pi_{1}(P_{T_{\mathcal{M}_{r}}({\bf Q})}({\bf Z}))$ is the closest point to $P_{T_{\mathcal{M}_{r}}({\bf Q})}({\bf Z})$ on the manifold $\mathcal{M}_{r}$.
Therefore, we have
 \begin{align}\label{new03}
 \|P_{T_{\mathcal{M}_{r}}({\bf Q})}({\bf Z})-\pi_{1}(&P_{T_{\mathcal{M}_{r}}({\bf Q})}({\bf Z}))\|_{F}
 \leq\nonumber\\
  &\|P_{T_{\mathcal{M}_{r}}({\bf Q})}({\bf Z})-\pi_{1}({\bf Z})\|_{F}.
 \end{align}
 We remark that ${\cal M}_{rn}$ is a smooth manifold \cite{sm2019} and
 ${\bf P} \in {\cal M}_{rn}$. Thus
 there exists an $s'$ such that $\pi$ is continuous on $Ball({\bf P}, s')$.
 In other words, we can find a constant $\alpha>0$ such that
\begin{align} \label{continuous}
\|\pi( {\bf X} )-\pi( {\bf X}')\|_{F}\leq \alpha \| {\bf X}-{\bf X}'\|_{F},
 \forall~ {\bf X},{\bf X}'\in Ball({\bf P},s').
\end{align}
Now we choose $s_1(\epsilon)$ to be the minimum of $s'$ and $s({\epsilon})$ in
Lemmas \ref{jia2} and \ref{ne1}.
For all ${\bf Z}\in Ball({\bf P},s_1(\epsilon))$, we have
\begin{align*}
&~\|\pi(\pi_{1}(P_{T_{\mathcal{M}_{r}}({\bf Q})}({\bf Z})))-\pi({\bf Z})\|_{F}\\
= &~\|\pi(\pi_{1}(P_{T_{\mathcal{M}_{r}}({\bf Q})}({\bf Z})))-\pi(\pi_{1}({\bf Z}))+\pi(\pi_{1}({\bf Z}))-\pi({\bf Z})\|_{F} \\
\leq &~\|\pi(\pi_{1}(P_{T_{\mathcal{M}_{r}}({\bf Q})}({\bf Z})))\hspace{-0.4mm}-\hspace{-0.4mm}\pi(\pi_{1}({\bf Z}))\|_{F}\hspace{-0.4mm}+\hspace{-0.4mm}\|\pi(\pi_{1}({\bf Z}))\hspace{-0.4mm}-\hspace{-0.4mm}\pi({\bf Z})\|_{F}\\
 \leq &~ \alpha\|\pi_{1}(P_{T_{\mathcal{M}_{r}}({\bf Q})}({\bf Z}))-\pi_{1}({\bf Z})\|_{F}+
 \varepsilon_1({\epsilon}) \|{\bf Z}-\pi({\bf Z})\|_{F} \\
  \leq &~ \alpha\|\pi_{1}(P_{T_{\mathcal{M}_{r}}({\bf Q})}({\bf Z}))-
 P_{T_{\mathcal{M}_{r}}({\bf Q})}({\bf Z})\|_{F}
  +\varepsilon({\epsilon})\|{\bf Z}-\pi({\bf Z})\|_{F}\\
 &~+\alpha\|P_{T_{\mathcal{M}_{r}}({\bf Q})}({\bf Z})-\pi_{1}({\bf Z})\|_{F} \\
\leq &~ 2\alpha\|P_{T_{\mathcal{M}_{r}}({\bf Q})}({\bf Z})-\pi_{1}({\bf Z})\|_{F}+
 \varepsilon({\epsilon}) \|{\bf Z}-\pi({\bf Z})\|_{F}  \\
\leq &~8\alpha\sqrt{\epsilon}\| {\bf Z}-{\bf Q}\|_{F}+\varepsilon({\epsilon})
  \|{\bf Z}-\pi({\bf Z})\|_{F}\\
\leq &~(\varepsilon({\epsilon})+8\alpha\sqrt{\epsilon})\| {\bf Z}-\pi({\bf Z})\|_{F} +
    8\alpha\sqrt{\epsilon}\| {\bf Q}-\pi({\bf Z}))\|_{F}.
\end{align*}
The second inequality is derived by  (\ref{continuous})  and  (\ref{an4}), the fourth inequality is derived by (\ref{new03}) and the fifth inequality is derived by (\ref{lemma32}).
We choose $\varepsilon_1({\epsilon}) =
\varepsilon({\epsilon})+8\alpha\sqrt{\epsilon}$ and
$\varepsilon_2({\epsilon}) = 8\alpha\sqrt{\epsilon}$. The result follows.
\end{proof}

\subsection{Proof of Lemma \ref{new2}}\label{Proof3}
\begin{proof}
 Without loss of generality, we can  assume $\pi({\bf Z})=0$, thus it is sufficient to prove
\begin{align*}
\|\pi_{1}(P_{T_{\mathcal{M}_{r}}({\bf Q})}({\bf Z}))\|_{F}<c\| {\bf Z}\|_{F} ~~\text{or}~~\frac{\pi_{1}(P_{T_{\mathcal{M}_{r}}({\bf Q})}({\bf Z}))\|_{F}}{\| {\bf Z}\|_{F}}<c
\end{align*}
is satisfied.
By the definition of $\sigma({\bf P})$, we can find a constant $c_{1}$ such that  $\sigma({\bf P})<c_{1}<c$.
Note that $\sigma(\cdot)$ is a local continuous function, it implies that there exist a constant
$s_{1}$ such that $\sigma({\bf S})<c_{1}$ is satisifed for every point ${\bf S}
\in \mathcal{M}_{rn} \cap Ball({\bf P},s_{1})$.
Let $s<s_{1}$. Since $\pi(\cdot)$ is a local continuous function,
$\pi({\bf Z})\in \mathcal{M}_{rn} \cap Ball({\bf P},s)$,
thus we have $\sigma(\pi({\bf Z}))=\sigma(0)<c_{1}$.
Now we set
${\bf D}=\pi_{1}({\bf Z})$, ${\bf D}'=P_{T_{\mathcal{M}_{r}}(\pi({\bf Z}))}({\bf Z})$
and ${\bf E}=\pi_{1}(P_{T_{\mathcal{M}_{r}}({\bf Q})}({\bf Z}))$, then we have
\begin{align*}
\frac{\|\pi_{1}(P_{T_{\mathcal{M}_{r}}({\bf Q})}({\bf Z}))\|_{F}}{\|{\bf Z}\|_{F}}=\frac{\|{\bf E}\|_{F}}{\| {\bf Z} \|_{F} }
=&\left ( \frac{\|{\bf E}\|_{F}}{\|{\bf D}' \|_{F}} \right )
\left ( \frac{\|{\bf D}' \|_{F}}{\| {\bf Z} \|_{F}} \right ).
\end{align*}
The remaining task is to estimate the values of
$\frac{\|{\bf E}\|_{F}}{\|{\bf D}' \|_{F}}$ and $\frac{\|{\bf D}' \|_{F}}{\|{\bf Z}\|_{F}}$.
 Recall that $\mathcal{M}_{n}$ is a linear affine manifold, and ${\bf Z}$ is on $\mathcal{M}_{n}$, it implies that
${\bf Z}=\pi_{2}({\bf Z})=P_{T_{\mathcal{M}_{n}}(\pi({\bf Z}))}({\bf Z})$.
Therefore,
\begin{align*}
\|{\bf D}'\|_{F}=&\|P_{T_{\mathcal{M}_{r}}(\pi({\bf Z}))}({\bf Z})\|_{F}
=\|P_{T_{\mathcal{M}_{r}}(\pi({\bf Z}))}({\bf Z})-\pi({\bf Z})\|_{F}\\
=&\sigma(\pi({\bf Z}))\|P_{T_{\mathcal{M}_{n}}(\pi({\bf Z}))}({\bf Z})\|_{F},
\end{align*}
and thus
$\frac{\|{\bf D}' \|_{F}}{\|{\bf Z}\|_{F}}=\sigma(\pi({\bf Z}))=\sigma(0)<c_1$.

In order to estimate $\frac{\|{\bf E}\|_{F}}{\|{\bf D}' \|_{F}}$, we consider
$\| {\bf E}-{\bf D}' \|_{F}$ which can be bounded by the following inequality:
\begin{align}
 &~ \| {\bf E}-{\bf D}'\|_{F} \nonumber \\
=&~\|{\bf E}-{\bf D}+{\bf D}-{\bf D}'\|_{F}\leq \|{\bf E}-{\bf D}\|_{F}+\|{\bf D}-{\bf D}'\|_{F} \nonumber\\
=&~ \|{\bf E}-P_{T_{\mathcal{M}_{r}}({\bf Q})}({\bf Z})+P_{T_{\mathcal{M}_{r}}({\bf Q})}({\bf Z})-
{\bf D}\|_{F}+\| {\bf D}-{\bf D}'\|_{F}\nonumber\\
\leq &~ \| {\bf E}-P_{T_{\mathcal{M}_{r}}({\bf Q})}({\bf Z})\|_{F}+
\|P_{T_{\mathcal{M}_{r}}({\bf Q})}({\bf Z})-{\bf D}\|_{F}\nonumber\\
&~+\|{\bf D}-{\bf D}'\|_{F}\nonumber\\
=&~\|\pi_{1}(P_{T_{\mathcal{M}_{r}}({\bf Q})}({\bf Z}))-P_{T_{\mathcal{M}_{r}}({\bf Q})}({\bf Z})
\|_{F}\nonumber\\
&~+\|P_{T_{\mathcal{M}_{r}}({\bf Q})}({\bf Z})-\pi_{1}({\bf Z})\|_{F}+\| {\bf D}-{\bf D}'\|_{F}.\label{song1}
\end{align}
By using Lemma \ref{ne1}
and ${\bf Z}= \pi_2({\bf Q})$ is the closest point to ${\bf Q}$ with respect to $\pi_2(\cdot)$,
\begin{align}\label{song2}
\|P_{T_{\mathcal{M}_{r}}({\bf Q})}({\bf Z})-\pi_{1}({\bf Z}) \|_{F}
<&4\sqrt{\epsilon}\| {\bf Z} - {\bf Q} \|_{F}\nonumber\\
\leq& 4\sqrt{\epsilon}\| {\bf Q} -\pi({\bf Z})\|_{F}.
\end{align}
By using the fact that $\pi_{1}(P_{T_{\mathcal{M}_{r}}({\bf Q})}({\bf Z}))$
is the closest point to $P_{T_{\mathcal{M}_{r}}({\bf Q})}({\bf Z})$ with respect
to $\pi_1(\cdot)$ and (\ref{song2}), we have
\begin{align}\label{song3}
&~\|\pi_{1}(P_{T_{\mathcal{M}_{r}}({\bf Q})}({\bf Z}))-P_{T_{\mathcal{M}_{r}}({\bf Q})}({\bf Z})\|_{F}\nonumber\\
\leq&~ \|P_{T_{\mathcal{M}_{r}}({\bf Q})}({\bf Z})-\pi_{1}({\bf Z})\|_{F}
\leq  4\sqrt{\epsilon}
\|{\bf Q}-\pi({\bf Z})\|_{F}.
\end{align}
By applying Lemma \ref{jia2} on $\| {\bf D}- {\bf D}'\|_{F}$, we get
\begin{align}\label{song5}
\| {\bf D}-{\bf D}'\|_{F}=\|P_{T_{\mathcal{M}_{r}}(\pi({\bf Z}))}({\bf Z})-\pi_{1}({\bf Z})\|_{F}
\leq 4\sqrt{\epsilon}\| {\bf Z} \|_{F}.
\end{align}
By putting \eqref{song2}, \eqref{song3} and \eqref{song5} into \eqref{song1}, we
obtain the following estimate
\begin{align}
&~\|{\bf E}-{\bf D}'\|_{F}\nonumber\\
<&~4\sqrt{\epsilon}\| {\bf Q}-\pi({\bf Z})\|_{F}+4\sqrt{\epsilon}\| {\bf Q}-\pi(
{\bf Z})\|_{F}\nonumber\\
&~+4\sqrt{\epsilon}\| {\bf Z}-\pi({\bf Z})\|_{F} \nonumber\\
=&~ 8\sqrt{\epsilon}\| {\bf Q}-\pi({\bf Z})\|_{F}+4\sqrt{\epsilon}\| {\bf Z}-{\bf Q}+{\bf Q}-\pi({\bf Z})
\|_{F} \nonumber\\
\leq&~ 8\sqrt{\epsilon}\| {\bf Q}-\pi({\bf Z})\|_{F}+
4\sqrt{\epsilon}(\| {\bf Z}-{\bf Q}\|_{F}+\| {\bf Q}-\pi({\bf Z})\|_{F}) \nonumber\\
<&~8\sqrt{\epsilon}\| {\bf Q}-\pi({\bf Z})\|_{F}+8\sqrt{\epsilon}\| {\bf Q}-\pi({\bf Z})\|_{F}\nonumber\\
=&~16\sqrt{\epsilon}\| {\bf Q}-\pi({\bf Z})\|_{F}.\label{nsong4}
\end{align}
Now we choose $c_{2}>1$ such that $c_{2}c_{1}<c$, and
also sufficiently small $\epsilon\in (0,\frac{3}{5})$ such that
\begin{align}\label{new6}
16\sqrt{\epsilon} \left ( \frac{c_{2}}{c_{2}-1} \right ) \| {\bf Q} \|_{F}<c \| {\bf Z} \|_{F}
\end{align}
is satisfied.
For the value of $\frac{\| {\bf E}\|_{F}}{\|{\bf D}'\|_{F}}$,
there are two cases to be considered:
$\frac{\|{\bf E}\|_{F}}{\|{\bf D}'\|_{F}}\leq c_{2}$ or
$\frac{\|{\bf E}\|_{F}}{\|{\bf D}'\|_{F}}> c_{2}$.
For the first case, it is easy to check
\begin{align*}
\frac{\| {\bf E}\|_{F}}{\| {\bf Z}\|_{F}}=\frac{\| {\bf E}\|_{F}}{\| {\bf D}'\|_{F}}\frac{\|{\bf D}'\|_{F}}{\| {\bf Z}\|_{F}}<c_{2}c_{1}<c.
\end{align*}
For the second case,
$\frac{\|{\bf E}\|_{F}}{\|{\bf D}'\|_{F}}> c_{2}$.
By using \eqref{nsong4}, we derive
$$
\|{\bf E}\|_{F}-\|{\bf D}'\|_{F}\leq\| {\bf E}-{\bf D}'\|_{F}<16\sqrt{\epsilon}\|{\bf Q}\|_{F}.
$$
or $\|{\bf D}'\|_{F}>\|{\bf E}\|_{F}-16\sqrt{\epsilon}\|{\bf Q}\|_{F}$.
It implies that
$$c_{2}<\frac{\|{\bf E}\|_{F}}{\|{\bf D}'\|_{F}}<\frac{\|{\bf E}\|_{F}}{\|{\bf E}\|_{F}-16\sqrt{\epsilon}\| {\bf Q} \|_{F}} .
$$
By using \eqref{new6},
we have
$$
\frac{\|{\bf E}\|_{F}}{\|{\bf Z}\|_{F}}<\frac{\left ( \frac{c_{2}}{c_{2}-1} \right ) 16\sqrt{\epsilon}\| {\bf Q} \|_{F}}{\|{\bf Z}\|_{F}}<c.
$$
The results follow.

\end{proof}

\bibliographystyle{IEEEtran}
\bibliography{ref1}

\begin{IEEEbiography}
[{\includegraphics[width=1in,height=1.25in,clip,keepaspectratio]{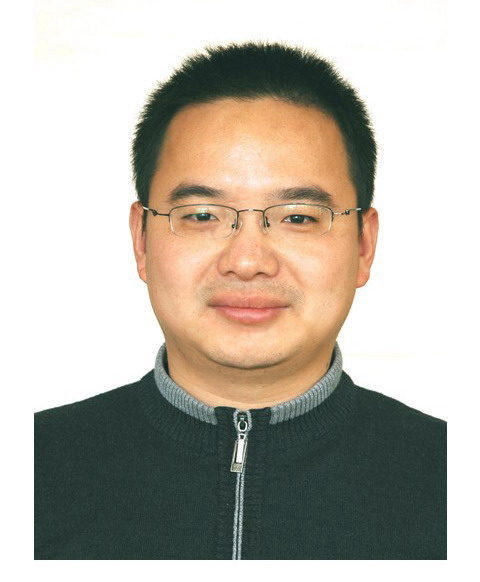}}]{Guang-Jing Song} received the Ph.D. degree in mathematics from  Shanghai University, Shanghai, China, in 2010. He is currently a professor of School of Mathematics and Information Sciences, Weifang University. His research interests include numerical linear algebra, sparse and low-rank modeling, tensor decomposition and multi-dimensional image processing.
\end{IEEEbiography}

\begin{IEEEbiography}[{\includegraphics[width=1in,height=1.25in,clip,keepaspectratio]{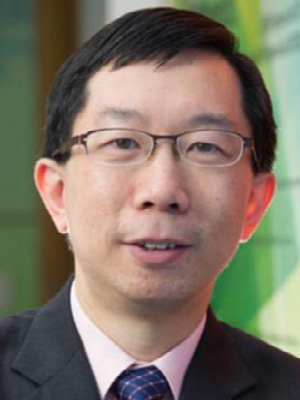}}]{Michael K. Ng} is the Director of Research Division for Mathematical and Statistical Science, and Chair Professor of Department of Mathematics, the University of Hong Kong,  and Chairman of HKU-TCL Joint Research Center for AI. His research areas are data science, scientific computing, and numerical linear algebra.
\end{IEEEbiography}

\begin{IEEEbiography}[{\includegraphics[width=1in,height=1.25in,clip,keepaspectratio]{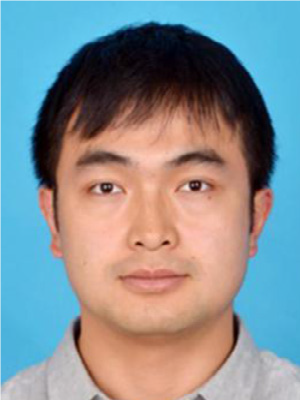}}]{Tai-Xiang Jiang} received the B.S., Ph.D. degrees in mathematics and applied mathematics from the University of Electronic Science and Technology of China (UESTC), Chengdu, China, in 2013. He is currently a lecturer with the School of Economic Information Engineering, Southwestern University of Finance and Economics. His re- search interests include sparse and low-rank model- ing, tensor decomposition and multi-dimensional image processing. https://sites.google.com/view/taixiangjiang/
\end{IEEEbiography}




\end{document}